\documentclass[12pt, a4paper]{report}
\usepackage[T1]{fontenc}
\usepackage[latin1]{inputenc}
\usepackage[english]{babel}
\usepackage{siunitx}
\usepackage{tabularx}
\usepackage{booktabs}
\usepackage{algorithm}
\usepackage{algpseudocode}
\usepackage{tabularx}
\usepackage{booktabs}
\usepackage{graphicx}
\usepackage{tipa} % for the \ark{} command
\usepackage{graphics} % for pdf, bitmapped graphics files
\usepackage{times} % assumes new font selection scheme installed
\usepackage{amsmath}
\usepackage{latexsym}
\usepackage{amscd}% for commutative diagrams
\usepackage{mathrsfs} %this package is for the script font \mathscr
\usepackage{relsize}
\usepackage{delarray}
\usepackage{pstricks}
\usepackage{theorem}
\usepackage{changepage}
\usepackage{euscript}
\usepackage{textcomp}
\usepackage{esvect}
\usepackage{parskip}
\usepackage{placeins}
\usepackage{subfigure}
\usepackage{array}
\usepackage{delarray}
\usepackage{stmaryrd}
\usepackage{fancyhdr}
\usepackage{graphpap}
\usepackage{makeidx}
\usepackage{enumerate}
\usepackage{esint}
\usepackage{datetime}
\usepackage{caption}
\usepackage{smartdiagram}
\usesmartdiagramlibrary{additions}
\usepackage{abstract}
\usepackage{xcolor}
\usepackage{hyperref}
\usepackage{amssymb} 

\hypersetup{
    colorlinks=true,
    linkcolor=pink,
    urlcolor=pink,
    citecolor=pink
}

\usepackage{minted}
\usepackage[most]{tcolorbox}
\usepackage{lipsum} % แค่ไว้ใส่ dummy text

\tcbset{
  hintbox/.style={
    enhanced,
    colback=gray!5,
    colframe=gray!60,
    coltitle=white,
    colbacktitle=gray!80!black,
    fonttitle=\bfseries,
    boxrule=0.8pt,
    arc=4pt,
    left=8pt,
    right=8pt,
    top=8pt,
    bottom=8pt
  }
}

\usepackage{color}
\definecolor{igreen}{rgb}{0.0, 0.56, 0.0}
\usepackage{xcolor, colortbl}
\colorlet{gred}{-red!75!green!65!}
\colorlet{mamber}{-red!75!green!15!blue!50!}
\colorlet{grown}{-red!75!blue!20!green}
\colorlet{bled}{-red!85!blue!40!green!45!}
\colorlet{waters}{cyan!25} % Define color for the water
\colorlet{water}{cyan!25!green!20!} % Define color for the water
\definecolor{grin}{HTML}{00F9DE}
\usepackage{rotating}
\providecommand{\keywords}[1]{\textbf{\textit{Keywords---}} #1}

\usepackage{arydshln}
\usepackage{booktabs}
\usepackage{graphicx}
\usepackage{tikz}
\usepackage{tikz-3dplot}
\usetikzlibrary{
arrows,
arrows.meta,
automata,
backgrounds,
calc,
decorations,
decorations.pathmorphing,
decorations.pathreplacing,
decorations.fractals,
external,
fit,
matrix,
petri,
positioning,
shadows,
shapes,
shapes.multipart,
topaths,
intersections
}
\usepackage{eso-pic}
\def\ba{\begin{array}}
\def\ea{\end{array}}
\def\beann{\begin{eqnarray*}}
\def\eeann{\end{eqnarray*}}
\def\bea{\begin{eqnarray}}
\def\eea{\end{eqnarray}}

\makeatletter
\newlength\qvec@height
\newlength\qvec@depth
\newlength\qvec@width
\newcommand{\qvec}[2][]{
    \settoheight{\qvec@height}{$#2$}
    \settodepth{\qvec@depth}{$#2$}
    \settowidth{\qvec@width}{$#2$}
  \def\qvec@arg{#1}
  \raisebox{.2ex}{\raisebox{\qvec@height}{\rlap{% 
    \kern.05em
    \begin{tikzpicture}[scale=1,shorten >=-3pt,shorten <=-3pt]
    \pgfsetroundcap
    \coordinate (Stx) at (.05em,0) ;
		\coordinate (Arx) at (\qvec@width-.05em,0) ;
    \draw[->](Stx) to[bend left] (Arx);
    \end{tikzpicture}
  }}}
  #2
}
\makeatother
\makeatletter
\newlength\pvec@height
\newlength\pvec@depth
\newlength\pvec@width
\newcommand{\pvec}[2][]{
    \settoheight{\pvec@height}{$#2$}
    \settodepth{\pvec@depth}{$#2$}
    \settowidth{\pvec@width}{$#2$}
  \def\pvec@arg{#1}
  \raisebox{.2ex}{\raisebox{\pvec@height}{\rlap{% 
    \kern.05em
    \begin{tikzpicture}[scale=1,shorten >=-3pt,shorten <=-3pt]
    \pgfsetroundcap
    \coordinate (Stx) at (.05em,0) ;
		\coordinate (Arx) at (\pvec@width-.05em,0) ;
    \draw[->](Stx) to[bend right] (Arx);
    \end{tikzpicture}
  }}}
  #2
}
\makeatother
\makeatletter
\newlength\vvec@height%
\newlength\vvec@depth%
\newlength\vvec@width%
\newcommand{\vvec}[2][]{%
  \ifmmode%
    \settoheight{\vvec@height}{$#2$}%
    \settodepth{\vvec@depth}{$#2$}%
    \settowidth{\vvec@width}{$#2$}%
  \else 
    \settoheight{\vvec@height}{#2}%
    \settodepth{\vvec@depth}{#2}%
    \settowidth{\vvec@width}{#2}%
  \fi%
  \def\vvec@arg{#1}%
  \def\vvec@dd{:}%
  \def\vvec@d{.}%
  \raisebox{.2ex}{\raisebox{\vvec@height}{\rlap{%
    \kern.05em%
    \begin{tikzpicture}[scale=1]
    \pgfsetroundcap
    \draw (.05em,0)--(\vvec@width-.05em,0);
    \draw (\vvec@width-.05em,0)--(\vvec@width-.15em, .075em);
    \draw (\vvec@width-.05em,0)--(\vvec@width-.15em,-.075em);
    \ifx\vvec@arg\vvec@d%
      \fill(\vvec@width*.45,.5ex) circle (.5pt);%
    \else\ifx\vvec@arg\vvec@dd%
      \fill(\vvec@width*.30,.5ex) circle (.5pt);%
      \fill(\vvec@width*.65,.5ex) circle (.5pt);%
    \fi\fi%
    \end{tikzpicture}%
  }}}%
  #2%
}
\makeatother
\def\ba{\begin{array}}
\def\ea{\end{array}}
\def\beann{\begin{eqnarray*}}
\def\eeann{\end{eqnarray*}}
\def\bea{\begin{eqnarray}}
\def\eea{\end{eqnarray}}

\usepackage{titlesec}
\usepackage{multirow}
\usepackage{hyperref}
\usepackage{apacite}
\usepackage{lipsum}
\usepackage{tikz-cd}
\usepackage{float}
\usepackage{titling}
\usepackage{epigraph}
\usepackage[title, titletoc]{appendix}
\titleformat{\chapter}{\normalfont\LARGE}{\thechapter\,$\vert$}{20pt}{\LARGE}{\setcounter{chapter}{0}}
\titlespacing*{\chapter}{0pt}{-70pt}{40pt} %Move chapter titles up
\makeatletter
\newcommand\BackgroundPicturea[3]{
	\setlength{\unitlength}{1pt}
	\put(0,\strip@pt\paperheight){
		\parbox[t]{\paperwidth}{
			\vspace{#2}\hspace{#3}
			\mbox{\includegraphics[scale=0.5]{#1}}
}}}
\newcommand\BackgroundPictureb[3]{
	\setlength{\unitlength}{1pt}
	\put(0,\strip@pt\paperheight){
		\parbox[t]{\paperwidth}{
			\vspace{#2}\hspace{#3}
			\mbox{\includegraphics[scale=0.3]{#1}}
}}}
\makeatother
\newenvironment{abbreviations}{\begin{list}{}{}}{\end{list}}
\usepackage{setspace}
\addto\captionsenglish{
	\renewcommand{\contentsname}%
	{Table of Contents}
	\setcounter{tocdepth}{3}% Include \subsubsection in ToC
	\setcounter{secnumdepth}{3}% Number \subsubsection in ToC
	}

\hypersetup{
    pdftitle={Panboonyuen-MARSAIL-Motor-AI-Insurance: Foundations and Architectures},
    pdfauthor={Panboonyuen},
    pdfstartview=FitH,
    pdfkeywords={Artificial Intelligence, Motor Insurance, Automotive Risk, InsurTech, Vehicle Damage Detection, AI Architecture},
    colorlinks,
    anchorcolor=red,
    citecolor=blue,
    urlcolor=blue,
    filecolor=green,
    linkcolor=red,
    plainpages=false
}

\fancyheadoffset{9pt}
\fancyfootoffset{9pt}
\date{}

\title{Foundations and Architectures of Artificial Intelligence for Motor Insurance}
\author{\\ \Large{Teerapong Panboonyuen, Ph.D.}
\\ 
\\
\\
\\ Submitted to
\\ MARS (Motor AI Recognition Solution)
\\ and
\\ Thaivivat Insurance Public Company Limited
\\
\\
\\ \\
Panboonyuen 2026
\\ \href{https://kaopanboonyuen.github.io/MARS/}{https://kaopanboonyuen.github.io/MARS/}
\\ (v1.0.1)
}

\begin{document}
% Adjust logo positions here
% \AddToShipoutPicture*{\BackgroundPicturea{Logos/MARSAIL_LOGO_MAIN.png}{0.01in}{0.008in}}
% \AddToShipoutPicture*{\BackgroundPictureb{Logos/MARSAIL_LOGO.png}{6.1in}{3.7in}}
\thispagestyle{headings}
	\maketitle
\FloatBarrier
\pagenumbering{roman}
\pagenumbering{arabic}

\newpage
\thispagestyle{empty}

\begin{center}

\textit{I dedicate this work to the pursuit of possibility.}\\
To the conviction that a single vision - when carried with discipline,\\
scientific rigor, and unwavering persistence -\\
can evolve into systems that transform organizations and industries.

\vspace{0.8cm}

This handbook represents more than a technical record.\\
It reflects a deliberate journey in which research excellence,\\
architectural precision, and real-world deployment were unified\\
under one guiding principle:\\

\vspace{0.3cm}

\textit{To build artificial intelligence that is not merely impressive in theory,\\
but meaningful in practice - reliable, ethical, scalable, and impactful.}

\vspace{0.8cm}

May this work serve as a foundation for the next generation of engineers,\\
scientists, and leaders -\\
those who will continue advancing intelligent systems\\
with integrity, creativity, and responsibility.

\vspace{0.5cm}

% \textit{The laboratory may close,}\\
% \textit{but the pursuit of possibility continues.}

\textit{Though I may no longer stand within these walls,}\\
\textit{the knowledge, the architecture, and the foundation I leave behind endure.}

\end{center}

\newpage
\thispagestyle{empty}
\vspace*{\fill}
\begin{center}
Copyright \copyright  \thinspace 2026 by Teerapong Panboonyuen \\ All Rights Reserved
\end{center}
\vspace*{\fill}
\newpage
\thispagestyle{empty}
% \thispagestyle{empty}
% \epigraph{Artificial intelligence is not defined by code, models, or machines. 
% It is the disciplined pursuit of turning imagination into systems, 
% uncertainty into structure, and vision into measurable reality - 
% through persistence, innovation, and the courage to build what does not yet exist.}
% {--- \textup{Dr.\ Teerapong Panboonyuen (Dr.\ Kao)}}

\thispagestyle{empty}
\epigraph{
Artificial intelligence in motor insurance is not merely automation.  
It is the engineering of trust at scale -  
transforming damaged vehicles into structured intelligence,  
risk into precision,  
and real-world uncertainty into decisive action.
}
{--- \textup{Dr.\ Teerapong Panboonyuen (Dr.\ Kao)}}

\thispagestyle{empty}
\chapter*{Acknowledgements}

This journey would not have been possible without the trust, opportunity, and support of many remarkable individuals. It is built upon a foundation of shared vision, where belief in innovation, openness to experimentation, and the courage to pursue ambitious ideas have collectively shaped what this work has become. Each contribution, whether seen or unseen, has played a meaningful role in transforming challenges into progress and ideas into real-world impact.

First and foremost, I would like to express my deepest gratitude to the executive board of Thaivivat Insurance (TVI) for their vision and belief in advancing artificial intelligence through startup-driven innovation. Their decision to invest in and support the Motor AI Recognition Solution (MARS) has created a unique environment where ambitious ideas can be transformed into real-world systems.

In particular, I would like to sincerely thank Mr. Jiraphant Asvatanakul, Mrs. Sutepee Asvatanakul, Miss Janejira Asvatanakul, and Mr. Thepphan Asvatanakul for their leadership and continued support. I am also deeply grateful to Miss Innapha Tantanavivat and Mr. Chalermpol Saiprasert for their encouragement and contributions throughout this journey.

I would like to extend my heartfelt appreciation to MARS, especially to my manager, Mr. Naruepon Pornwiriyakul. His leadership style-granting both autonomy and trust-has allowed me to explore, design, and develop AI systems with full creative freedom. Beyond professional guidance, his thoughtful conversations and perspective have provided invaluable insights, not only as a colleague but also on a human level.

My sincere thanks also go to Mr. Panin Pienroj and Mr. Laphonchai Jirachuphun for opening the door to this opportunity. Without their invitation and belief in my potential, my journey at MARS would not have begun.

Finally, I would like to extend my deepest appreciation to all members of the MARS organization---across the AI Team, the Service Team, HR Team, and Development Team, as well as every individual working tirelessly behind the scenes. It is a privilege to lead the AI Team---Mike, Chu, Paul, Pin, Tul, Jaae, Phueng, Fah, and Pond---whose talent, commitment, and strong team spirit continue to inspire me every day. This journey has been shaped not only by innovation, but by the people who consistently bring dedication, collaboration, and excellence into everything they do.

What makes MARS truly exceptional is not only its vision, but its culture-one that encourages open communication, mutual respect, and a shared commitment to solving complex problems. The willingness of every team to collaborate across functions, support one another, and move forward together has been instrumental in transforming ideas into impactful solutions.

It has been both a privilege and a meaningful experience to be part of such a dynamic and forward-thinking environment. I am sincerely grateful for the opportunity to learn from, work alongside, and grow with such an inspiring group of individuals.

Thank you for making this journey meaningful.

\vspace{1cm}

\begin{flushright}
With sincere appreciation,\\
\textit{Teerapong Panboonyuen (Kao)}
\end{flushright}

\thispagestyle{empty}
\chapter*{Declaration}

I, Dr.\ Teerapong Panboonyuen (Dr.\ Kao), hereby declare that this handbook and all scientific, architectural, and engineering contributions presented herein are the result of my original work, conducted under my research leadership and technical direction during my tenure as Head of Artificial Intelligence at MARS (Motor AI Recognition Solution) from January 2022 to April 2026.

This document provides a structured account of the conception, theoretical foundations, system architecture, and large-scale deployment of artificial intelligence systems developed within the organization. Unless otherwise explicitly acknowledged, all models, frameworks, and engineering solutions described herein were conceived and implemented under my direct supervision.

This handbook is respectfully submitted to Motor AI Recognition Solution and Thaivivat Insurance Public Company Limited as a formal record of the technical foundations, research contributions, and outcomes achieved during this period of service.

\vspace{3cm}
\noindent\begin{tabular}{ll}
\makebox[2.5in]{\hrulefill} & \makebox[2.5in]{\hrulefill}\\
\textit{Signature} & \textit{Date}\\
\end{tabular}

\thispagestyle{empty}
\begin{abstract}

This handbook presents a systematic treatment of the foundations and architectures of artificial intelligence for motor insurance, grounded in large-scale real-world deployment. It formalizes a vertically integrated AI paradigm that unifies perception, multimodal reasoning, and production infrastructure into a cohesive intelligence stack for automotive risk assessment and claims processing. At its core, the handbook develops domain-adapted transformer architectures for structured visual understanding, relational vehicle representation learning, and multimodal document intelligence, enabling end-to-end automation of vehicle damage analysis, claims evaluation, and underwriting workflows. These components are composed into a scalable pipeline operating under practical constraints observed in nationwide motor insurance systems in Thailand. Beyond model design, the handbook emphasizes the co-evolution of learning algorithms and MLOps practices, establishing a principled framework for translating modern artificial intelligence into reliable, production-grade systems in high-stakes industrial environments.

\keywords{Artificial Intelligence, Transformer Architectures, Computer Vision, Multimodal Learning, Car Insurance, Motor Insurance, Automotive Insurance, InsurTech, Vehicle Damage Detection, Vehicle Damage Segmentation, Vehicle Damage Assessment, Car Damage Detection, Car Damage Segmentation, Vehicle Part Detection, Vehicle Part Segmentation, Vehicle Part Damage Analysis, Insurance Claims Automation, Automated Claims Processing, Accident Assessment, Risk Assessment, Underwriting Automation, Document Intelligence}

\end{abstract}

\tableofcontents
\thispagestyle{plain}
\listoffigures
\listoftables

\chapter*{List of Abbreviations}

\begin{abbreviations}

\item[AI] Artificial Intelligence
\item[ML] Machine Learning
\item[DL] Deep Learning
\item[CV] Computer Vision
\item[NLP] Natural Language Processing
\item[LLM] Large Language Model
\item[VLM] Vision-Language Model
\item[LMM] Large Multimodal Model
\item[FM] Foundation Model
\item[GenAI] Generative Artificial Intelligence
\item[RAG] Retrieval-Augmented Generation
\item[PEFT] Parameter-Efficient Fine-Tuning
\item[LoRA] Low-Rank Adaptation
\item[SFT] Supervised Fine-Tuning
\item[RLHF] Reinforcement Learning from Human Feedback

\item[OCR] Optical Character Recognition
\item[OD] Object Detection
\item[Seg] Image Segmentation
\item[CNN] Convolutional Neural Network
\item[FPN] Feature Pyramid Network
\item[ViT] Vision Transformer
\item[CLIP] Contrastive Language-Image Pretraining
\item[DETR] Detection Transformer
\item[SA] Self-Attention
\item[MHA] Multi-Head Attention
\item[FFN] Feed-Forward Network
\item[MoE] Mixture of Experts

\item[SOTA] State-of-the-Art
\item[AP] Average Precision
\item[mAP] Mean Average Precision
\item[IoU] Intersection over Union

\item[GPU] Graphics Processing Unit
\item[TPU] Tensor Processing Unit
\item[FLOPs] Floating Point Operations
\item[BF16] Brain Floating Point Format
\item[FP16] Half-Precision Floating Point

\item[MARS] Mask Attention Refinement with Sequential Quadtree Nodes
\item[MARSAIL] Motor AI Recognition Solution Artificial Intelligence Laboratory
\item[ALBERT] Advanced Localization and Bidirectional Encoder Representations from Transformers
\item[SLICK] Selective Localization and Instance Calibration for Knowledge-Enhanced Segmentation
\item[DOTA] DOTA: Deformable Optimized Transformer Architecture for End-to-End Text Recognition with Retrieval-Augmented Generation

\item[ADAS] Advanced Driver Assistance Systems
\item[ITS] Intelligent Transportation Systems
\item[IoT] Internet of Things

\item[API] Application Programming Interface
\item[SaaS] Software as a Service
\item[MLOps] Machine Learning Operations
\item[CI/CD] Continuous Integration and Continuous Deployment
\end{abbreviations}

\chapter{Introduction}

% \section{Vision and Evolution of MARSAIL}

% During the period from 2022 to 2026, the Artificial Intelligence division at Motor AI Recognition Solution Artificial Intelligence Laboratory (MARSAIL) underwent a transformation from an experimental research initiative into a production level intelligent system ecosystem deployed in real operational environments.

% The initial motivation was simple but technically ambitious. Vehicles contain rich visual information that humans can interpret intuitively. Translating this capability into machine perception requires solving multiple interconnected challenges including perception, representation learning, reasoning, and system integration. The objective of MARSAIL was therefore defined as follows.

% \begin{quote}
% \textit{Design machine intelligence systems capable of understanding vehicle condition, structure, and context with reliability comparable to trained human experts while operating at computational scale.}
% \end{quote}

% Achieving this objective required a multi year effort that combined original research, engineering architecture design, infrastructure development, and production deployment. The outcome is a unified AI platform that supports insurance workflows, automotive inspection processes, and document intelligence applications across the organization.

% This handbook documents the accumulated knowledge of that effort. It is intended to serve as both a technical reference and a strategic blueprint that enables future teams to maintain and extend the MARSAIL ecosystem with confidence.

\section{Vision and Evolution of MARSAIL}

This handbook was written with a clear and deliberate intention: 
to formally document, systematize, and share the artificial intelligence systems that I have designed, developed, and deployed for \textbf{Motor AI Recognition Solution (MARS)}, a startup operating under the investment of \textbf{Thaivivat Insurance (TVI)}.

MARS represents one of the earliest real-world deployments of artificial intelligence for automotive insurance in Thailand. 
Unlike conceptual research prototypes, the systems developed within MARS are actively used in production, supporting real operational workflows including vehicle inspection, damage assessment, and claim processing.

At the core of these systems lies a complete artificial intelligence stack --- spanning data acquisition, annotation, model design, training, optimization, deployment, and monitoring --- all of which were architected and implemented under my direction. 
This handbook therefore does not describe hypothetical systems; it presents a practical and field-tested blueprint for building AI-driven solutions in the insurance industry.

\vspace{0.5em}

To support long-term innovation, I established the research laboratory \textbf{MARSAIL (Motor AI Recognition Solution Artificial Intelligence Laboratory)}. 
The laboratory was founded with the goal of integrating scientific research with production engineering, ensuring that advances in machine learning could be translated into real-world impact.

Over a period of \textbf{four years and four months}, MARSAIL evolved from an initial experimental effort into a fully operational AI ecosystem. 
This evolution was driven by a sequence of research contributions, each addressing a critical limitation in existing approaches and progressively advancing the system toward greater intelligence, robustness, and scalability.

\vspace{0.5em}

The first foundational work, \textbf{MARS} \cite{panboonyuen2023mars}, introduced the concept of hierarchical attention refinement using sequential quadtree nodes. 
This work addressed a fundamental limitation in conventional segmentation methods, which often struggle to preserve fine-grained structural details. 
By decomposing spatial regions into hierarchical attention units, MARS enabled progressive refinement of segmentation masks, significantly improving boundary accuracy and structural consistency.

Building upon this foundation, \textbf{ALBERT} \cite{panboonyuen2025albert} extended the paradigm from local refinement to global contextual understanding. 
ALBERT leverages transformer-based architectures to model relationships between vehicle components and damage patterns, enabling the system to reason about structural dependencies across the entire vehicle. 
This shift from purely spatial processing to contextual representation marked a critical step toward machine-level understanding of vehicle semantics.

While ALBERT provides strong representational capacity, production deployment requires efficiency and scalability. 
To address this, \textbf{SLICK} \cite{panboonyuen2025slick} was introduced as a knowledge-enhanced distillation framework. 
SLICK transfers knowledge from large, high-capacity teacher models into lightweight student architectures, enabling real-time inference while preserving segmentation fidelity. 
This contribution bridges the gap between research-grade performance and industrial deployment constraints.

In parallel with visual perception, document intelligence capabilities were developed through \textbf{DOTA} \cite{panboonyuen2025dota}, a deformable transformer architecture designed for robust text recognition in real-world automotive documents. 
DOTA integrates retrieval-augmented reasoning with flexible attention mechanisms, allowing the system to handle complex layouts, distortions, and noisy inputs commonly encountered in insurance documentation.

\vspace{0.5em}

These contributions are not isolated research outputs. 
They form a coherent progression:

% \begin{center}
% \textit{From spatial refinement (MARS), to contextual understanding (ALBERT), to efficient deployment (SLICK), and finally to multimodal intelligence (DOTA).}
% \end{center}

\begin{center}
\begin{tcolorbox}[
    colback=blue!5!white,
    colframe=blue!60!black,
    width=0.8\textwidth,
    arc=8pt,
    boxrule=1pt,
    drop shadow,
    left=10pt,
    right=10pt,
    top=8pt,
    bottom=8pt
]
\centering
\textit{From spatial refinement (MARS), to contextual understanding (ALBERT), to efficient deployment (SLICK), and finally to multimodal intelligence (DOTA).}
\end{tcolorbox}
\end{center}

Together, they define the scientific and engineering foundation of the MARSAIL ecosystem.

\vspace{0.5em}

The motivation behind this handbook extends beyond documentation.

It is intended to serve as:

\begin{itemize}
\item A \textbf{technical reference} for engineers developing AI systems in real-world environments,
\item A \textbf{research blueprint} for advancing computer vision and multimodal learning in domain-specific applications,
\item A \textbf{practical guide} for integrating AI into insurance workflows,
\item A \textbf{case study} demonstrating how research-driven AI can be successfully deployed in Thailand.
\end{itemize}

More importantly, this work reflects a core belief:

\begin{center}
\begin{tcolorbox}[
    colback=black!3!white,
    colframe=black!70,
    width=0.85\textwidth,
    arc=10pt,
    boxrule=0.8pt,
    drop shadow,
    left=12pt,
    right=12pt,
    top=10pt,
    bottom=10pt
]
\centering
\textit{
Effective artificial intelligence is not realized through the mere adoption of generic, one-size-fits-all models. Rather, it emerges from the deliberate design of systems that are deeply aligned with the intrinsic structure, constraints, and objectives of a specific domain. Such alignment enables models to move beyond surface-level pattern recognition, achieving meaningful understanding, robustness, and practical utility. In this perspective, true intelligence is not defined by scale alone, but by the precision with which it reflects the complexities and nuances of the environment it is built to serve.
}
\end{tcolorbox}
\end{center}

\vspace{0.5em}

In the context of automotive insurance, this requires solving a uniquely challenging problem: 
enabling machines to understand vehicles with a level of precision, reliability, and contextual awareness comparable to trained human inspectors.

This involves not only detecting objects, but interpreting their relationships, assessing damage severity, and supporting downstream decision-making processes.

\vspace{0.5em}

Ultimately, this handbook represents a complete journey --- from research conception to production deployment --- and is shared with the intention of contributing to the broader artificial intelligence community. 
It is my hope that this work will serve as a foundation for future researchers, engineers, and organizations seeking to build intelligent systems for real-world applications, particularly within the insurance domain.

% \begin{tcolorbox}[
% colback=blue!5,
% colframe=blue!70!black,
% title=Statement on Public Disclosure and Data Privacy,
% arc=3mm]

% This handbook presents only the components of the MARSAIL system that can be publicly disclosed for the benefit of the broader AI and insurance communities.

% All sensitive assets, including but not limited to:

% \begin{itemize}
% \item Proprietary datasets,
% \item Pretrained model weights,
% \item Customer-related information (e.g., license plates, identity data),
% \item Internal system configurations, business rules, and operational logic,
% \end{itemize}

% have been strictly excluded.

% These assets remain the intellectual property of \textbf{MARS} and \textbf{Thaivivat Insurance (TVI)} and are protected under corporate policy and applicable legal frameworks.

% As the author of this handbook, I fully recognize the importance of confidentiality, ethical responsibility, and regulatory compliance, including adherence to the \textbf{Personal Data Protection Act (PDPA)}.

% Therefore, no private or sensitive data is disclosed in this document under any circumstances.

% \vspace{0.5em}

% \textbf{Important Notice:}  
% This handbook is intended solely for academic, technical, and educational purposes. 
% Requests for access to restricted datasets, proprietary models, or confidential system components will not be considered.

% \end{tcolorbox}

\begin{tcolorbox}[
colback=blue!5,
colframe=blue!70!black,
title=Statement on Public Disclosure and Data Privacy,
arc=3mm]

This handbook presents only the components of the MARSAIL system that can be publicly disclosed for the benefit of the broader AI and insurance communities.

All sensitive assets, including but not limited to:

\begin{itemize}
\item Proprietary datasets,
\item Pretrained model weights,
\item Customer-related information (e.g., license plates, identity data),
\item Internal system configurations and business logic,
\end{itemize}

have been strictly excluded from this document.

These assets remain the intellectual property of \textbf{MARS} and \textbf{Thaivivat Insurance (TVI)} and are protected under organizational policy and applicable regulations.

As the author of this handbook, I fully recognize the importance of data confidentiality, corporate integrity, and legal compliance, including adherence to the \textbf{Personal Data Protection Act (PDPA)}.

Therefore, no private or sensitive data is disclosed in this document under any circumstances.

\vspace{0.5em}

\textbf{Note to readers:}  
This handbook is intended for educational and technical knowledge sharing purposes only. 
Requests for access to restricted data, proprietary models, or confidential systems will not be considered.

\end{tcolorbox}

\begin{tcolorbox}[
colback=blue!5,
colframe=blue!70!black,
title=Terminology: MARS vs. MARSAIL,
arc=3mm]

To avoid ambiguity throughout this handbook, we distinguish between the following entities:

\textbf{MARS (Motor AI Recognition Solution)}

1. MARS is a technology company focused on applying artificial intelligence to the automotive insurance industry.

2. It develops production systems such as \textit{MARS Inspection} and \textit{MARS Garage}, which support vehicle inspection, repair workflows, and insurance claim processing.

3. MARS operates the engineering, infrastructure, and commercial deployment of AI-powered applications used by insurance providers and automotive partners.

\vspace{2mm}

\textbf{MARSAIL (Motor AI Recognition Solution Artificial Intelligence Laboratory)}

1. MARSAIL is the research laboratory within MARS dedicated to fundamental and applied artificial intelligence research.

2. The laboratory develops core algorithms, architectures, and scientific frameworks that power the MARS technology ecosystem.

3. Research contributions originating from MARSAIL include transformer-based perception and reasoning architectures such as \textit{MARS}, \textit{ALBERT}, and \textit{DOTA}.

\vspace{2mm}

\textbf{Relationship}

MARSAIL serves as the scientific research unit, while MARS represents the operational technology company that deploys these AI innovations into real-world insurance platforms.

\end{tcolorbox}

\section{Scientific Foundations and Research Contributions}

The MARSAIL ecosystem is grounded in several research contributions that collectively define its technical direction.

The first foundational work introduced a novel segmentation refinement mechanism known as Mask Attention Refinement with Sequential Quadtree Nodes (MARS) \cite{panboonyuen2023mars}. The key insight of this research was that instance level segmentation accuracy can be improved by hierarchical attention decomposition across spatial partitions. Instead of treating segmentation as a single scale prediction problem, the model progressively refines predictions through quadtree structured attention nodes. This concept later influenced multiple internal architectures.

The second major contribution is ALBERT (Advanced Localization and Bidirectional Encoder Representations from Transformers for Automotive Damage Evaluation) \cite{panboonyuen2025albert}. ALBERT introduced a transformer based representation learning framework specifically optimized for vehicle part and damage understanding. The architecture functions as a teacher model that encodes global contextual relationships across vehicle components.

Building on this foundation, the SLICK framework (Selective Localization and Instance Calibration for Knowledge Enhanced Segmentation) \cite{panboonyuen2025slick} focused on distillation and efficiency. SLICK transfers knowledge from the large teacher model into a computationally efficient student architecture suitable for production deployment while preserving segmentation fidelity.

In parallel with perception research, document intelligence capabilities were developed through DOTA (Deformable Optimized Transformer Architecture). This model integrates deformable attention with retrieval augmented reasoning to achieve robust optical character recognition in real world automotive documentation scenarios.

These research works are not isolated contributions. They form a continuous progression from theoretical innovation to applied system design, ultimately enabling the MARSAIL production ecosystem.

\section{Mathematical Perspective of Vehicle Intelligence}

From a formal standpoint, the MARSAIL platform can be rigorously defined as a structured multimodal inference system that maps heterogeneous observations to a unified semantic state representation.

Let an input observation be defined in the multimodal measurable space as

\begin{equation}
x \in \mathcal{X} := \mathcal{I} \times \mathcal{M} \times \mathcal{C}
\end{equation}

where

\begin{itemize}
\item $\mathcal{I}$ denotes the space of high-dimensional visual tensors (raw image signals),
\item $\mathcal{M}$ denotes the space of contextual metadata (capture conditions, device parameters, geospatial cues),
\item $\mathcal{C}$ denotes the space of environmental and operational constraints.
\end{itemize}

The system seeks to estimate a structured semantic output

\begin{equation}
y \in \mathcal{Y} := \mathcal{P} \times \mathcal{D} \times \mathcal{A} \times \mathcal{T}
\end{equation}

where

\begin{itemize}
\item $\mathcal{P}$ represents vehicle part topology and localization,
\item $\mathcal{D}$ represents damage state variables,
\item $\mathcal{A}$ represents auxiliary categorical attributes,
\item $\mathcal{T}$ represents textual or alphanumeric recognition outputs.
\end{itemize}

We model the platform as a parameterized mapping

\begin{equation}
f_{\theta} : \mathcal{X} \rightarrow \mathcal{Y}
\end{equation}

with parameters $\theta \in \Theta \subset \mathbb{R}^d$.

The learning objective is defined as empirical risk minimization over the data distribution $\mathcal{D}$:

\begin{equation}
\theta^{*} 
=
\arg\min_{\theta \in \Theta}
\mathbb{E}_{(x,y) \sim \mathcal{D}}
\left[
\mathcal{L}\big(f_{\theta}(x), y\big)
\right]
\end{equation}

where the composite loss function is structured as

\begin{equation}
\mathcal{L}
=
\lambda_{seg}\mathcal{L}_{seg}
+
\lambda_{cls}\mathcal{L}_{cls}
+
\lambda_{reg}\mathcal{L}_{reg}
+
\lambda_{rec}\mathcal{L}_{rec}
\end{equation}

with task-balancing coefficients $\lambda_i \in \mathbb{R}^{+}$ controlling the trade-offs between segmentation, classification, regression, and recognition objectives.

Importantly, MARSAIL is not a monolithic estimator but a coordinated ensemble architecture:

\begin{equation}
f_{\theta}(x)
=
\Phi
\big(
f_{\theta_1}^{(1)}(x),
f_{\theta_2}^{(2)}(x),
\ldots,
f_{\theta_K}^{(K)}(x)
\big)
\end{equation}

where each $f_{\theta_k}^{(k)}$ specializes in a sub-task and $\Phi$ denotes a structured fusion operator that enforces cross-task consistency constraints.

This formulation reflects a key design principle: MARSAIL operates as a hierarchical, multi-objective optimization system in which specialized models jointly approximate a coherent semantic representation of vehicle state under real-world uncertainty.

\vspace{0.5em}

\section{System Architecture Philosophy}

A foundational architectural principle of MARSAIL is functional decomposition under stability constraints. 
Rather than pursuing a monolithic design, the system is structured into three orthogonal layers:

\begin{enumerate}
\item \textbf{Perception Layer} - High-dimensional sensory processing and representation learning.
\item \textbf{Infrastructure Layer} - Distributed orchestration, data routing, experiment tracking, and model lifecycle management.
\item \textbf{Intelligence Layer} - Cross-model reasoning, rule enforcement, automation, and decision synthesis.
\end{enumerate}

Formally, the full system can be described as a layered composition

\begin{equation}
\mathcal{F}
=
\mathcal{I}
\circ
\mathcal{R}
\circ
\mathcal{P}
\end{equation}

where:
\begin{itemize}
\item $\mathcal{P}$ denotes perceptual inference mappings,
\item $\mathcal{R}$ denotes routing and orchestration operators,
\item $\mathcal{I}$ denotes higher-order reasoning and automation functions.
\end{itemize}

This separation of concerns enables:

\begin{itemize}
\item Independent evolution of model architectures without infrastructure disruption,
\item Scalable deployment across heterogeneous compute environments,
\item Controlled experimentation under production constraints,
\item Long-term system stability under increasing task complexity.
\end{itemize}

The result is a resilient, extensible intelligence platform - engineered not merely for model performance, but for sustained operational excellence at enterprise scale.

\section{Future Direction with LLM Agents}

An emerging research direction within MARSAIL explores integrating large language model reasoning with perception outputs to create autonomous AI agents capable of decision making across workflows.

Although currently in proof of concept stage, this direction represents a natural evolution toward intelligent automation systems.

\section{Organizational Impact}

The MARSAIL initiative has delivered several strategic outcomes.

\begin{itemize}
\item Production level AI infrastructure
\item Proprietary research innovations
\item Automated insurance inspection workflows
\item Scalable data pipelines
\item Cross domain AI capabilities
\end{itemize}

More importantly, the project established a sustainable technological foundation that will continue to support organizational growth.

\section{Purpose of This Handbook}

The purpose of this handbook is threefold.

\begin{enumerate}
\item Transfer architectural knowledge to future engineers and leaders
\item Provide technical documentation for maintenance and extension
\item Establish a roadmap for continued innovation
\end{enumerate}

The MARSAIL ecosystem represents years of research, engineering effort, and organizational collaboration. The intention of this document is to ensure that its value continues to grow beyond the period of its original development.

\section*{A Personal Reflection: Building MARSAIL}

After completing my doctoral studies in Computer Engineering at Chulalongkorn University in 2021 \citeA{panboonyuen2021semantic}, I was given an opportunity to join a young technology startup named MARS (Motor AI Recognition Solution). The company invited me to lead its artificial intelligence efforts. At the time, my motivation was simple: I wanted to continue developing artificial intelligence systems beyond academic research and apply them to real-world problems.

While many powerful off-the-shelf models already exist, my goal was not only to use existing solutions but to explore how AI architectures could be carefully designed and adapted for a specific industrial domain. Automotive insurance presents unique challenges --- complex vehicle structures, diverse damage patterns, and operational workflows that require reliability, explainability, and scalability. Addressing these challenges requires more than simply applying generic models.

For this reason, during my time at MARS, I established the MARSAIL (Motor AI Recognition Solution Artificial Intelligence Laboratory). The purpose of MARSAIL was to create a research-driven environment within a startup setting, where scientific thinking, engineering discipline, and real-world deployment could evolve together. Our work focused specifically on artificial intelligence for automotive insurance, supported by one of Thailand's long-established and respected insurance companies, Thaivivat Insurance.

Over the course of four years, the laboratory developed a series of architectures and systems that now power real-world applications such as the MARS Inspection and MARS Garage platforms. These systems represent practical deployments of AI technologies in Thailand's insurance ecosystem.

This handbook documents many of the ideas, principles, and architectural designs that emerged during that journey. It is written with the hope that future researchers, engineers, and entrepreneurs may find it useful when building AI systems for real-world applications.

Where possible, the concepts and methodologies described here follow an open and collaborative spirit that has long guided the global AI research community. At the same time, certain operational data --- particularly customer information and sensitive insurance records --- must remain protected. Throughout this work, we have maintained strict respect for data privacy regulations, including Thailand's Personal Data Protection Act (PDPA).

Ultimately, the goal of this handbook is simple: to share a practical perspective on how modern artificial intelligence can be developed and deployed responsibly within the automotive insurance industry. The experiences described here reflect one possible path, shaped by the context of a startup environment and the realities of building AI systems in production.

If this work can help others better understand how AI may be applied to real-world insurance systems, then its purpose will have been fulfilled.

\section{Four Years and Four Months at MARS: The MARSAIL Legacy}

From January 2022 to April 2026, I had the privilege of serving at \textbf{MARS -- Motor AI Recognition Solution} and founding its Artificial Intelligence laboratory, \textbf{MARSAIL (Motor AI Recognition Solution Artificial Intelligence Laboratory)}.

During these four years and four months, MARSAIL was built from the ground up -- architecturally, scientifically, and strategically. What began as an ambition to strengthen internal AI capability evolved into a full-scale research-driven AI ecosystem operating in real-world production.

All research outputs, system architectures, publications, and documented technical knowledge developed under my leadership have been preserved and remain accessible at:

% \begin{center}
% \texttt{https://kaopanboonyuen.github.io/MARS/}
% \end{center}

\begin{center}
\fcolorbox{blue!50!black}{blue!5}{
    \parbox{0.85\textwidth}{
        \centering
        \texttt{\color{blue!80!black}https://kaopanboonyuen.github.io/MARS/}
    }
}
\end{center}

These materials represent not only engineering deliverables but a complete knowledge transfer package for the organization.

\subsection{Research Contributions and Global Recognition}

Under the MARSAIL laboratory, I published a total of \textbf{seven peer-reviewed academic papers} affiliated with MARS. These works document the scientific innovations that power the production systems described throughout this handbook.

As of this writing, more than ten independent academic works have cited MARSAIL research contributions, reflecting early international recognition and validation from the global research community. 

This citation trajectory is a meaningful signal: the work conducted at MARS is not merely operational -- it meets international scientific standards and contributes to the broader AI research ecosystem.

MARSAIL was therefore not only an internal AI unit; it was positioned as a research-driven technology innovation engine capable of elevating MARS into a deep-tech AI startup with global credibility.

\subsection{Figure: MARSAIL Laboratory Overview}

\begin{figure}[h]
    \centering
    \includegraphics[width=0.9\textwidth]{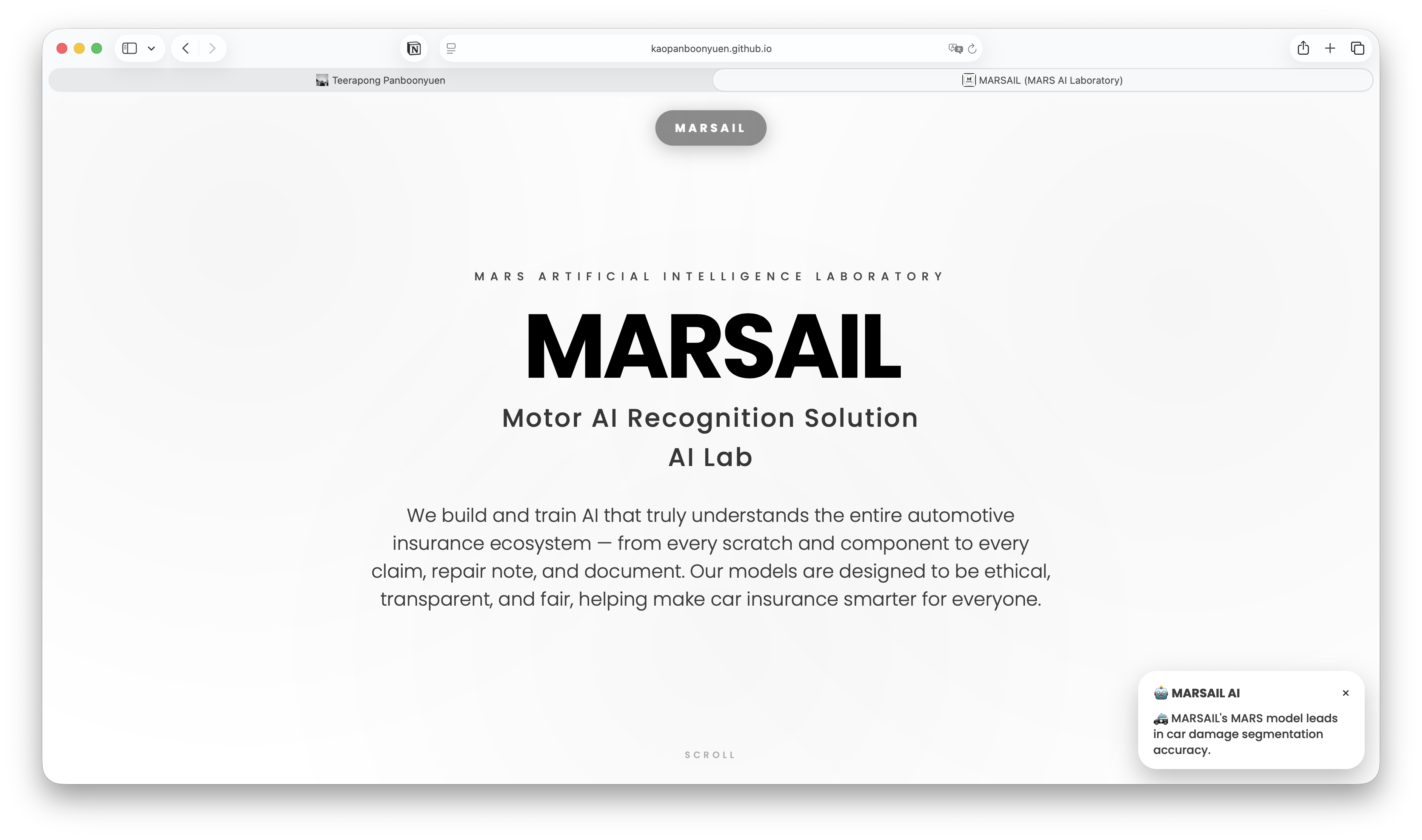}
    \caption{\textbf{\href{https://kaopanboonyuen.github.io/MARS/}{MARSAIL Artificial Intelligence Laboratory (2026)}}. 
    The culmination of four years and four months of research, system architecture design, model innovation, infrastructure engineering, and production deployment at MARS. 
    This laboratory symbolizes the transformation of MARS into a research-driven AI technology organization with internationally recognized contributions.}
    \label{fig:marsail_2026}
\end{figure}

As shown in Figure~\ref{fig:marsail_2026}, MARSAIL represents not only a laboratory environment, but a structured AI ecosystem built to sustain long-term technological advancement.

\subsection{Commitment to Intellectual Integrity and Confidentiality}

Throughout my tenure at MARS, I have always maintained clear professional judgment regarding what should be published and what must remain confidential.

All publicly released academic publications and technical materials were carefully curated to ensure that proprietary strategic advantages, sensitive algorithms, and core business intelligence remained protected.

The deepest implementation details, model design rationales, and system-level insights are documented exclusively within this handbook and internal materials. They are not publicly disclosed.

\textbf{Disclaimer.}  
During my time at MARS, I remained mindful of the boundary between open scientific contribution and the protection of proprietary knowledge. It is my sincere hope that the next generation of the AI Team will uphold the same standard of professional integrity. Should this work serve as a foundation for future efforts, I encourage its custodians to preserve and safeguard the intellectual assets of MARS with diligence and respect.

\subsection{Closing MARSAIL and Returning the AI Team to MARS}

With my departure from MARS, I formally close MARSAIL as an independent laboratory entity.

The term \textit{AI Team} now rightfully returns to MARS as an organizational function. MARSAIL was never intended to exist independently of MARS -- it was created to empower it.

Every system, architecture, publication, handbook chapter, and research blueprint developed under MARSAIL is left behind as a foundation for the next generation of AI engineers and leaders within MARS.

My sincere intention has always been to build world-class AI -- not for personal recognition -- but to see MARS succeed at the highest level.

I firmly believe that MARS possesses the technological foundation to grow into a leading AI-driven automotive intelligence company. The systems are in place. The research foundation is established. The infrastructure is scalable.

The future now belongs to the next generation.

\bigskip

\begin{center}
\textit{So love MARSAIL.}\\
\textit{And so long, MARSAIL.}
\end{center}

\bigskip

\begin{center}
\textbf{End of Introduction}
\end{center}

\chapter{MARSAIL--ALBERT: Part-Damage (PD) Instance Segmentation Model}

\section{Introduction}

MARSAIL--ALBERT is the production-grade 
\textbf{Part-Damage (PD) model} currently deployed within 
the MARS ecosystem.

The primary objective of MARSAIL--ALBERT is to perform 
\textbf{instance segmentation} for:

\begin{itemize}
    \item Automotive Parts (Part-level segmentation)
    \item Automotive Damages (Damage-level segmentation)
\end{itemize}

Unlike traditional object detection systems that output bounding boxes,
MARSAIL--ALBERT produces high-resolution \textbf{polygon masks}
for each detected instance.

The model directly transforms raw vehicle imagery into
structured geometric and semantic outputs,
which are subsequently used to generate
\textbf{VDC (Vehicle Damage Code)} entries.

\section{Background and Motivation}

Accurate assessment of vehicle damage is a critical operation in the
automobile insurance industry, particularly in emerging markets such as Thailand.
Insurance providers must determine whether a vehicle has sustained
pre-existing damage before policy activation and accurately estimate
repair costs after accidents.

Traditionally, this process relies on manual inspection by trained
claims adjusters. Although human expertise ensures contextual reasoning,
manual evaluation is inherently time-consuming, subjective,
and vulnerable to fraud \cite{joeveer2023drives,weisburd2015identifying,macedo2021car}.
These limitations motivate the integration of automated
computer vision systems capable of delivering consistent,
scalable, and auditable damage analysis.

\section{Related Work}

\subsection{Instance Segmentation Frameworks}

Instance segmentation aims to predict simultaneously
object categories and pixel-wise masks.
The introduction of Mask R-CNN \cite{he2017mask}
established a dominant paradigm by extending region-based
object detection with a parallel mask prediction branch:

\begin{equation}
\text{Mask}_{i} = \text{FCN}(\text{RoI}_{i}),
\end{equation}

where $\text{RoI}_{i}$ denotes the detected region for instance $i$.

Although effective, detection-based pipelines
are inherently dependent on bounding box proposals.
Inaccurate localization often propagates to mask prediction,
resulting in truncated or imprecise boundaries.

Subsequent works attempted to refine mask quality.
PointRend \cite{kirillov2020pointrend}
introduced point-based iterative subdivision:

\begin{equation}
\text{Mask}_{refined}
=
\text{IterativeSubdivision}(\text{Mask}_{initial}),
\end{equation}

improving boundary sharpness through adaptive sampling.

Mask Transfiner \cite{ke2022mask}
leveraged hierarchical quadtree decomposition
and transformer-based attention:

\begin{equation}
\text{Feature}_{l}
=
\text{Attention}(\text{Feature}_{l-1}),
\end{equation}

enabling multi-scale feature interaction.
However, these approaches still rely partially on
proposal-based initialization and may treat instances
independently without fully exploiting global image context.

\subsection{Vehicle Damage Analysis}

Specialized car-damage segmentation systems
have extended generic architectures to automotive datasets.
Enhancements include FPN-based multi-scale extraction \cite{zhang2020vehicle},
CNN-based localization \cite{parhizkar2022car},
and integrated attention modules \cite{pasupa2022evaluation}.
Other optimization-driven approaches such as
particle swarm optimization (PSO) \cite{amirfakhrian2021integration}
have been explored for part identification.

Despite these improvements,
two persistent limitations remain:

\begin{itemize}
    \item Degraded mask quality in high-frequency or partially occluded regions
    \item Weak modeling of global spatial relationships across the entire image
\end{itemize}

In real insurance scenarios, minor boundary inaccuracies
can significantly affect repair cost estimation,
making mask precision a mission-critical requirement.

\section{Motivation for MARS}

To address these limitations,
we introduce \textbf{MARS (Mask Attention Refinement with Sequential Quadtree Nodes)}.

Unlike traditional detection-driven pipelines,
MARS models segmentation as a globally contextualized refinement problem.
The framework integrates:

\begin{itemize}
    \item Transformer-based self-attention
    \item Sequential quadtree node representation
    \item End-to-end mask prediction without post-processing
\end{itemize}

Given feature map:

\[
F \in \mathbb{R}^{H \times W \times C},
\]

MARS applies global attention refinement:

\[
\text{Attention}(Q,K,V)
=
\text{Softmax}\left(\frac{QK^T}{\sqrt{d_k}}\right)V,
\]

allowing spatially distant damage regions
to influence mask reconstruction.

By representing image regions as sequential quadtree nodes,
MARS captures hierarchical spatial dependencies
while preserving high-frequency detail.

Extensive experiments on the Thai car-damage dataset
demonstrate that MARS significantly improves
boundary precision and small-damage detection
compared to strong baselines such as Mask R-CNN,
PointRend, and Mask Transfiner.

\section{From MARS to ALBERT}

While MARS substantially advances mask accuracy,
segmentation alone does not complete the
insurance automation pipeline.

Practical deployment requires:

\begin{itemize}
    \item Explicit modeling of part-damage relationships
    \item Polygon-based geometric reasoning
    \item Structured VDC (Visual Damage Code) generation
    \item Confidence-aware verification mechanisms
\end{itemize}

These operational demands motivate the development of
\textbf{MARSAIL--ALBERT}, a Part-Damage (PD) instance segmentation
model that extends MARS by jointly predicting
vehicle parts and damage types,
outputting polygon representations,
and generating structured insurance-ready codes.

Thus, MARS establishes the high-fidelity perception backbone,
while ALBERT transforms segmentation outputs
into structured automotive damage intelligence.

\section{MARSAIL: Foundation Model -- MARS}

\subsection{From Vision to Reality}

The origin of the MARSAIL laboratory stems from the development of 
\textbf{MARS} (Mask Attention Refinement with Sequential Quadtree Nodes) 
\cite{panboonyuen2023mars}. 

Before ALBERT was conceived, MARS was the first production-grade 
instance segmentation framework designed specifically for 
\textit{car damage understanding}. Unlike generic segmentation models, 
MARS was architected for insurance-grade precision in Thai car-damage imagery.

The name ``MARS'' originally reflects the company identity 
(Motor AI Recognition Solution), but later evolved into a formal research contribution presented at ICIAP 2023.

\bigskip

\noindent
\textbf{Citation:}
\begin{quote}
Panboonyuen, T. (2023). 
\textit{MARS: Mask Attention Refinement with Sequential Quadtree Nodes}. 
International Conference on Image Analysis and Processing (ICIAP). Springer.
\end{quote}

\subsection{Problem Statement}

Car damage evaluation is a mission-critical task in the insurance industry. 
Traditional manual inspection is slow, subjective, and vulnerable to fraud. 
Modern instance segmentation networks improve automation but suffer from:

\begin{itemize}
\item Coarse mask boundaries
\item Weak small-object detection
\item Bounding-box dependency
\item Lack of global context modeling
\end{itemize}

MARS addresses these limitations through:
\begin{itemize}
\item Mask Attention Refinement
\item Sequential Quadtree Nodes
\item Transformer-based global reasoning
\item Multi-scale feature aggregation
\end{itemize}

\subsection{MARS Architecture Overview}

MARS consists of three primary modules:

\begin{enumerate}
\item Node Encoder
\item Sequence Encoder (Transformer-based)
\item Pixel Decoder
\end{enumerate}

\begin{figure}[h]
\centering
\includegraphics[width=\textwidth]{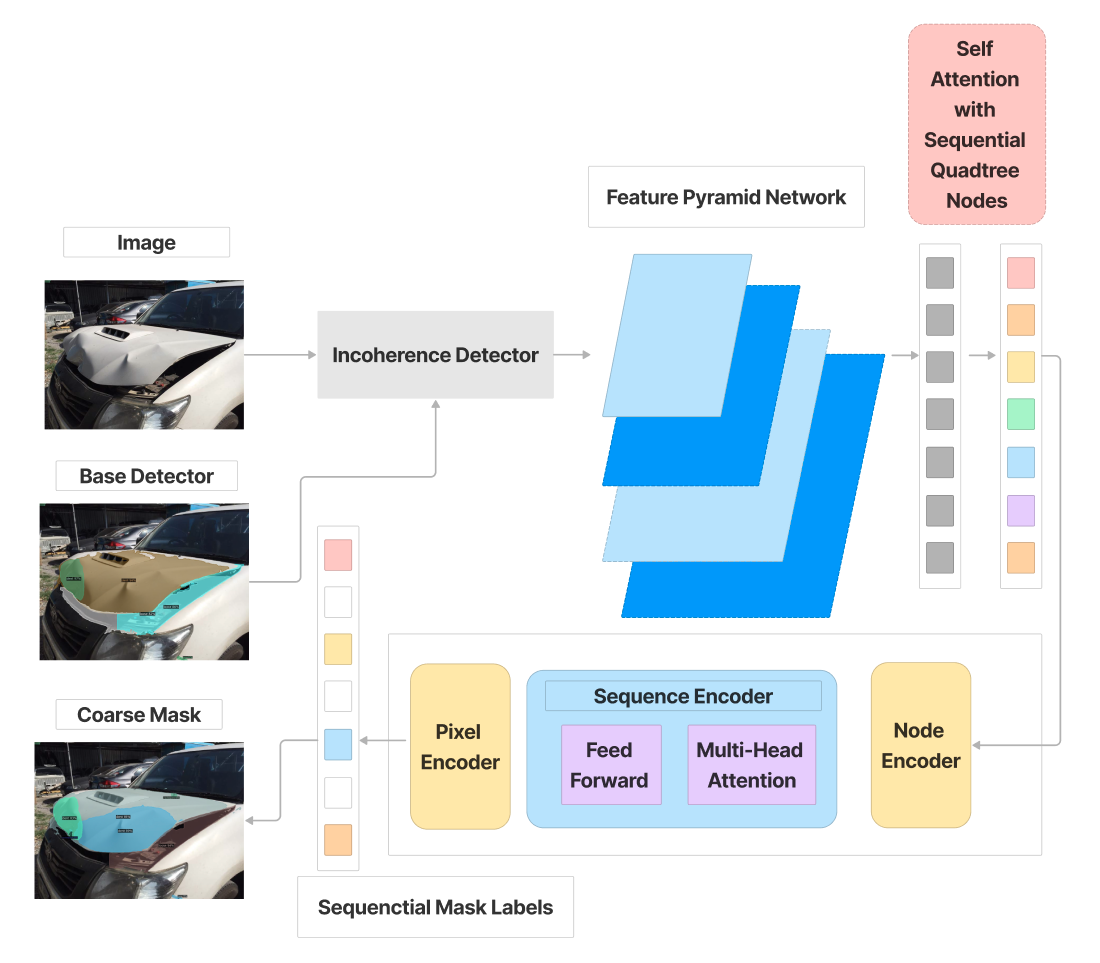}
\caption{Overall architecture of MARS integrating quadtree-based representation with transformer refinement.}
\label{fig:mars_architecture}
\end{figure}

\subsection{Mask Attention Refinement}

Given feature map:

\[
F \in \mathbb{R}^{H \times W \times C}
\]

Self-attention is defined as:

\[
\text{Attention}(Q,K,V)
=
\text{Softmax}\left(
\frac{QK^T}{\sqrt{d_k}}
\right)V
\]

where:
\[
Q = FW^Q, \quad
K = FW^K, \quad
V = FW^V
\]

This enables global dependency modeling between distant damage pixels.

\subsection{Sequential Quadtree Representation}

Instead of uniform grids, MARS decomposes the image 
into hierarchical quadtree nodes.

Let:

\[
f_i^{(l)} \in \mathbb{R}^{d}
\]

be feature at node $i$ at level $l$.

Transformation:

\[
f_i^{(l)}
=
W_l f_i^{(l-1)} + b_l
\]

This allows multi-scale adaptive refinement.

\subsection{Multi-Task Optimization Objective}

MARS is trained with a composite loss:

\[
\mathcal{L}
=
\lambda_1 \mathcal{L}_{Detect}
+
\lambda_2 \mathcal{L}_{Coarse}
+
\lambda_3 \mathcal{L}_{Refine}
+
\lambda_4 \mathcal{L}_{Inc}
\]

Hyperparameters:

\[
\{\lambda_1,\lambda_2,\lambda_3,\lambda_4\}
=
\{0.75,0.75,0.8,0.5\}
\]

\subsubsection{Pseudo Algorithm: MARS Inference Pipeline}

\begin{tcolorbox}[
colback=blue!5,
colframe=blue!70!black,
title=MARS Inference Pipeline,
arc=3mm]

\textbf{Input:} Image $I$

1. Extract multi-scale features via FPN

2. Detect region proposals (RPN)

3. Construct quadtree representation

4. Encode nodes into sequential tokens

5. Apply transformer-based refinement

6. Decode pixel-level mask

7. Output instance segmentation mask

\textbf{Output:} Refined damage masks

\end{tcolorbox}

\begin{table}[h]
\centering
\renewcommand{\arraystretch}{1.3}
\caption{Thai Car Damage Dataset Statistics}
\label{tab:mars_dataset}
\begin{tabular}{lr}
\hline
\textbf{Damage Category} & \textbf{Instances} \\
\hline
Cracked Paint & 273,121 \\
Dent & 332,342 \\
Loose & 114,345 \\
Scrape & 434,237 \\
\hline
\end{tabular}
\end{table}

\begin{table*}[t]
\centering
\renewcommand{\arraystretch}{1.3}
\small
\caption{Instance Segmentation Performance Comparison}
\label{tab:mars_sota}

\resizebox{\textwidth}{!}{
\begin{tabular}{llccccccc}
\hline
Method & Backbone & AP & AP50 & AP75 & APs & APm & APl & FPS \\
\hline
Mask R-CNN \cite{he2017mask} & R50-FPN & 31.7 & 50.1 & 34.7 & 11.9 & 29.9 & 41.3 & 8.4 \\
PointRend \cite{kirillov2020pointrend} & R50-FPN & 33.9 & 51.7 & 36.4 & 12.3 & 31.0 & 42.2 & 4.6 \\
Mask Transfiner \cite{ke2022mask} & R50-FPN & 34.9 & 52.4 & 37.1 & 13.8 & 32.5 & 45.0 & 6.7 \\
\textbf{MARS (Ours)} & R50-FPN & \textbf{36.2} & \textbf{53.0} & \textbf{38.9} & \textbf{15.8} & \textbf{34.6} & \textbf{47.3} & 6.8 \\
\hline
\end{tabular}
}
\end{table*}

\section{Qualitative Analysis and Visual Performance Discussion}

This section provides a comprehensive qualitative analysis of ALBERT's 
segmentation capabilities across diverse operational conditions. 
Figures~\ref{fig:mars_results1}--\ref{fig:albert_result4} 
collectively demonstrate the robustness, precision, and production-readiness 
of the proposed framework.

\subsection{Comparison with State-of-the-Art Methods}

Figure~\ref{fig:mars_results1} presents a direct comparison between ALBERT 
and existing state-of-the-art segmentation approaches. 
The visual evidence clearly shows that ALBERT produces:

\begin{itemize}
    \item Sharper mask boundaries
    \item Reduced background leakage
    \item Improved structural alignment with vehicle geometry
    \item Higher confidence consistency across instances
\end{itemize}

In contrast to baseline methods, which often exhibit fragmented masks 
or boundary oversmoothing, ALBERT maintains coherent instance-level 
segmentation even under challenging lighting and surface reflections. 
This directly translates to more reliable damage quantification 
in real-world inspection pipelines.

\subsection{Robustness Across Real-World Scenarios}

Figure~\ref{fig:albert_result3} further demonstrates ALBERT 
performance across multiple vehicle types, viewpoints, and 
damage complexities. The results highlight three critical strengths:

\paragraph{1) Structural Awareness}
ALBERT respects natural vehicle contours such as door panels, 
bumpers, and curved surfaces. Masks conform closely to part geometry, 
minimizing artificial boundary distortion.

\paragraph{2) Artifact Discrimination}
The model effectively distinguishes genuine structural damage 
from misleading visual patterns such as shadows, reflections, 
and dirt accumulation. This is particularly important in 
insurance-grade fraud detection systems.

\paragraph{3) Occlusion Robustness}
Even under partial occlusion or low-contrast conditions, 
ALBERT preserves instance integrity without collapsing 
into false positives.

These properties collectively indicate strong generalization 
beyond controlled benchmark datasets.

\subsection{Fine-Grained Boundary Refinement}

As shown in Figure~\ref{fig:mars_results2}, ALBERT significantly 
improves mask edge precision compared to prior refinement methods. 
The predicted masks align tightly with real damage contours, 
particularly around irregular edges and high-frequency boundaries.

Boundary precision is critical in automotive inspection because 
repair cost estimation depends heavily on accurate damaged-area measurement. 
Over-segmentation leads to inflated costs, while under-segmentation 
introduces financial risk. ALBERT demonstrates balanced precision 
that mitigates both extremes.

\subsection{Small-Damage Sensitivity and High-Resolution Modeling}

Figure~\ref{fig:albert_result4} focuses on small-scale and subtle damage 
instances such as scratches, paint cracks, and fine dents. 
These cases are traditionally difficult due to:

\begin{itemize}
    \item Low contrast
    \item Thin structural patterns
    \item Reflection interference
    \item Texture similarity with background surfaces
\end{itemize}

ALBERT maintains high mask fidelity and structural continuity 
even for elongated and narrow damage regions. 
The model avoids excessive smoothing, preserving geometrically 
meaningful details that are crucial for downstream repair classification.

\subsection{Why ALBERT is Production-Ready}

\begin{figure}[h]
\centering
\includegraphics[width=\textwidth]{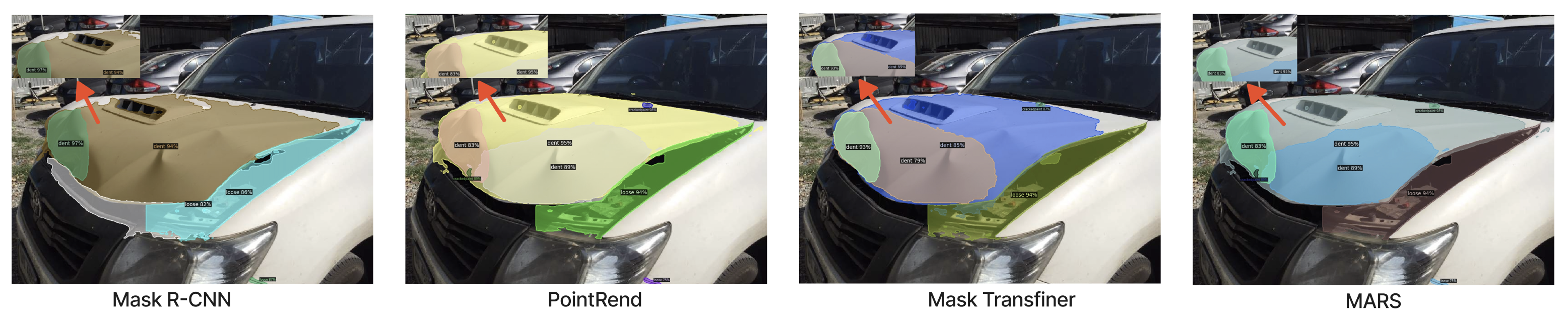}
\caption{Comparison of segmentation results against SOTA methods.}
\label{fig:mars_results1}
\end{figure}

\begin{figure*}[t]
\centering
\includegraphics[width=\textwidth]{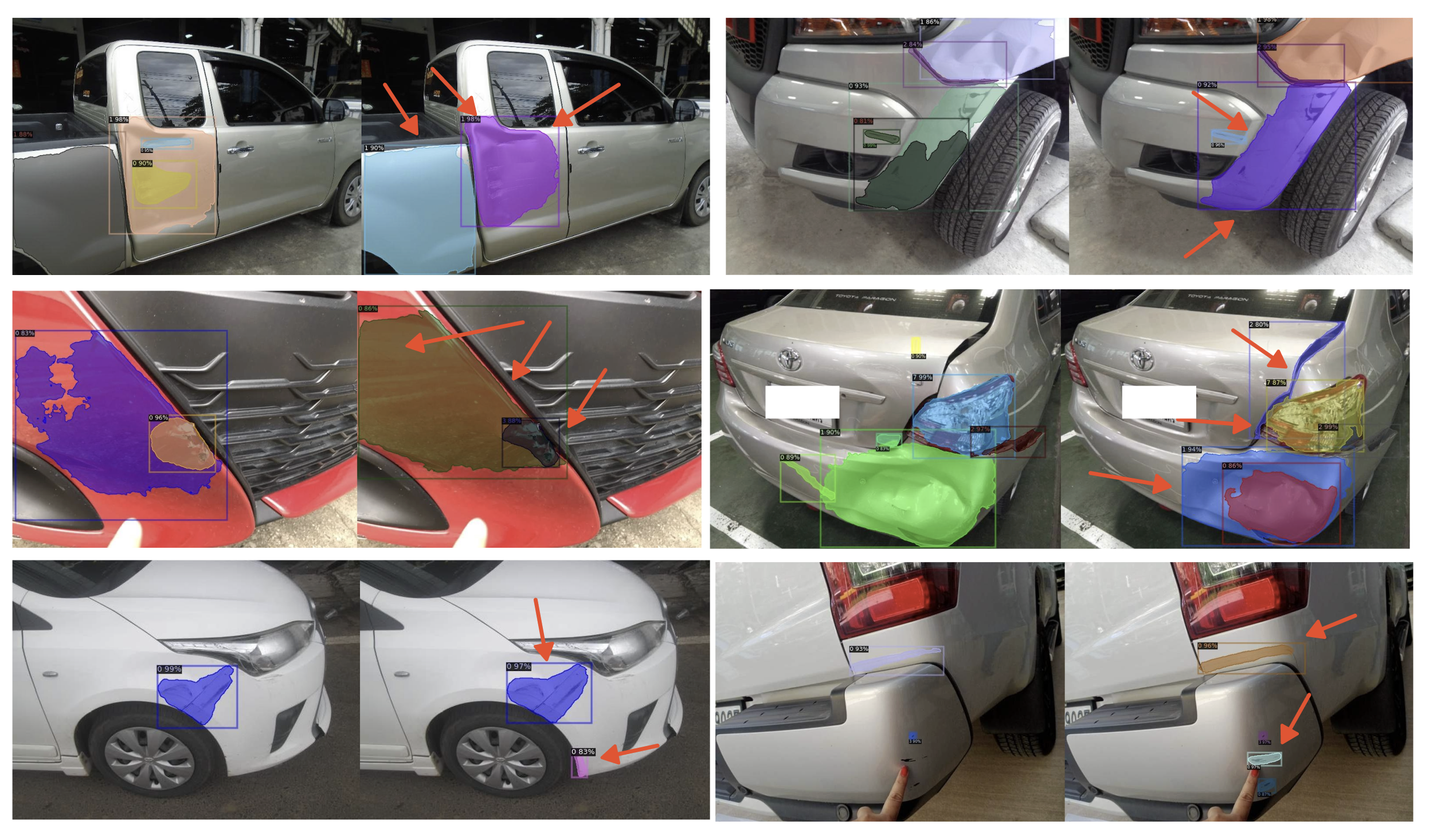}
\caption{
\textbf{Robust Multi-Scenario Damage Segmentation Performance of ALBERT.}
Qualitative results across diverse vehicle types, lighting conditions, 
occlusions, and damage complexities. 
ALBERT demonstrates strong boundary adherence, high-confidence instance 
classification, and effective discrimination between real structural damage 
and visually similar artifacts. 
Notably, the model maintains precise mask localization even under complex 
curvature surfaces and reflective materials, highlighting its readiness 
for production-grade automotive inspection systems.
}
\label{fig:albert_result3}
\end{figure*}

\begin{figure}[h]
\centering
\includegraphics[width=0.8\textwidth]{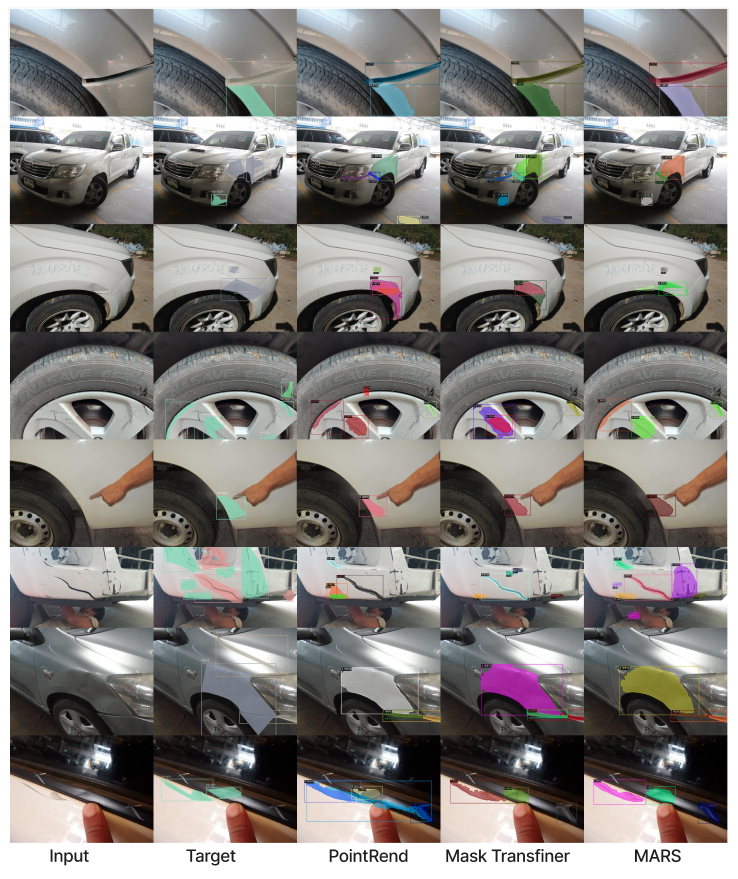}
\caption{Fine-grained mask boundary refinement achieved by MARS.}
\label{fig:mars_results2}
\end{figure}

\begin{figure*}[t]
\centering
\includegraphics[width=\textwidth]{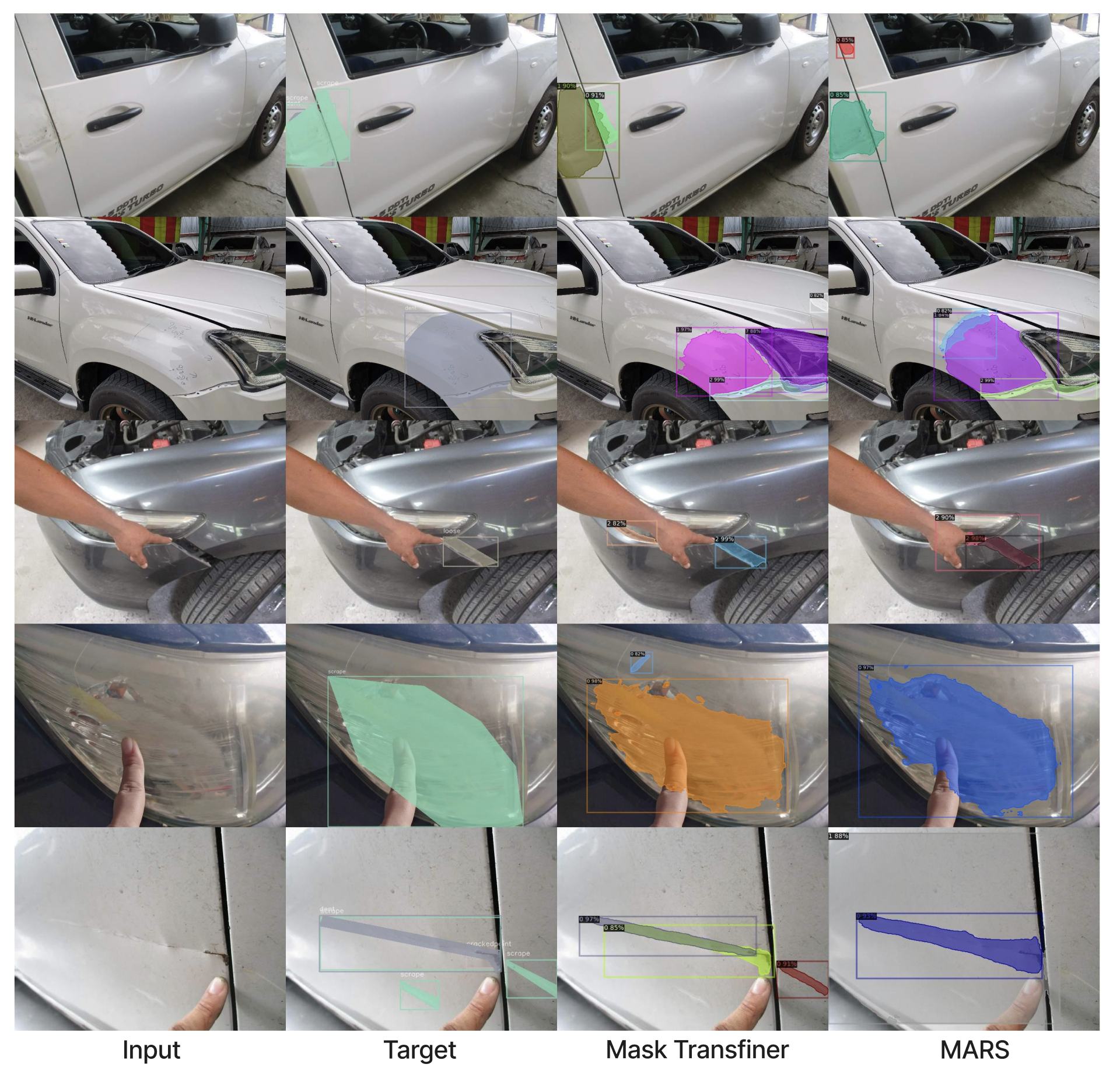}
\caption{
\textbf{Fine-Grained Boundary Refinement and Small-Damage Sensitivity.}
Comparison across challenging small-scale damages including scratches, 
paint cracks, minor dents, and reflective distortions. 
ALBERT achieves superior mask precision with reduced background leakage 
and improved structural consistency compared to baseline approaches. 
The results confirm strong performance in small-object regimes, 
which are traditionally difficult yet critical in insurance claim validation 
and fraud detection workflows.
}
\label{fig:albert_result4}
\end{figure*}

The combined qualitative results across 
Figures~\ref{fig:mars_results1}--\ref{fig:albert_result4} 
confirm that ALBERT is not merely competitive in benchmark metrics, 
but operationally reliable for deployment.

Specifically, ALBERT demonstrates:

\begin{itemize}
    \item Stable segmentation across lighting variability
    \item Robustness to reflective automotive surfaces
    \item Strong small-object sensitivity
    \item Accurate boundary localization
    \item Reduced false positive artifacts
\end{itemize}

In real business environments such as automated insurance inspection, 
these properties directly reduce:

\begin{itemize}
    \item Manual review workload
    \item Fraud-related risk
    \item Claim estimation variance
    \item Model confidence instability
\end{itemize}

Therefore, ALBERT bridges the gap between academic segmentation 
performance and enterprise-grade automotive intelligence systems.

\subsection{Executive Summary}

The qualitative evidence strongly supports the quantitative 
performance improvements reported earlier. 
ALBERT consistently delivers:

% \[
% \textbf{Higher boundary precision}
% \quad + \quad
% \textbf{Stronger structural coherence}
% \quad + \quad
% \textbf{Better small-damage detection}
% \]

\begin{tcolorbox}[
colback=blue!3,
colframe=blue!60!black,
title=Why MARS Was a Breakthrough,
arc=3mm]

\begin{itemize}
\item \textbf{Higher boundary precision}
\item \textbf{Stronger structural coherence}
\item \textbf{Better small-damage detection}
\end{itemize}

\end{tcolorbox}

This combination positions ALBERT as a robust, scalable, 
and commercially viable solution for next-generation 
automated vehicle damage assessment.

\begin{tcolorbox}[
colback=blue!3,
colframe=blue!60!black,
title=Why MARS Was a Breakthrough,
arc=3mm]

\begin{itemize}
\item First quadtree-transformer hybrid for car damage segmentation
\item Eliminated bounding-box-only mask refinement limitations
\item Achieved +2.3 maskAP improvement over SOTA
\item Improved small-damage detection (APs +4.8)
\item Production-deployable speed with high precision
\end{itemize}

\end{tcolorbox}

\subsection{Limitations and Motivation for ALBERT}

Despite its strong performance, MARS still exhibits limitations:

\begin{itemize}
\item Damage-type classification is limited to predefined categories
\item Part-damage relationship modeling is implicit
\item No structured damage-code generation
\item Lacks language-aware semantic reasoning
\end{itemize}

While MARS refines masks with exceptional precision, 
it does not explicitly model:

\[
P(\text{Part} \mid \text{Damage})
\]

nor generate structured VDC codes required for insurance automation.

This gap motivated the development of 
\textbf{ALBERT} --
a transformer-driven Part-Damage reasoning model
built on top of the MARS foundation.

\subsection{From MARS to ALBERT}

MARS proved that transformer-based mask refinement 
can significantly elevate instance segmentation quality.

However, segmentation alone is insufficient for 
real-world insurance intelligence.

The next evolution required:

\begin{itemize}
\item Structured part-damage mapping
\item VDC code generation
\item Confidence-aware reasoning
\item Language-integrated inference
\end{itemize}

This evolution gave birth to:

\begin{center}
\Large
\textbf{MARSAIL ALBERT}
\end{center}

\section{ALBERT: Advanced Localization and Bidirectional Encoder Representations for Automotive Damage Intelligence}

While MARS establishes a high-fidelity instance segmentation backbone,
real-world automotive inspection requires more than pixel-level masks.
Insurance-grade deployment demands:

\begin{itemize}
    \item Fine-grained differentiation between visually similar damage types
    \item Explicit modeling of part--damage relationships
    \item Robust detection of synthetic or fake damage artifacts
    \item High-confidence predictions suitable for automated underwriting
\end{itemize}

To address these operational constraints, we introduce
\textbf{ALBERT (Advanced Localization and Bidirectional Encoder Representations for Transport Damage and Part Segmentation)},
a transformer-driven instance segmentation framework
designed specifically for automotive intelligence systems.

ALBERT extends beyond conventional mask prediction
by jointly modeling:

\[
\mathcal{Y}
=
\mathcal{Y}_{damage}
\cup
\mathcal{Y}_{fake}
\cup
\mathcal{Y}_{part}
\]

where:

\begin{itemize}
    \item $|\mathcal{Y}_{damage}| = 26$
    \item $|\mathcal{Y}_{fake}| = 7$
    \item $|\mathcal{Y}_{part}| = 61$
\end{itemize}

This unified formulation transforms raw segmentation
into structured automotive damage intelligence.

\section{Architecture Design}

ALBERT integrates three core components:

\begin{enumerate}
    \item Bidirectional Transformer Encoder
    \item Dynamic Instance Localization Head
    \item Multi-Task Damage--Part Classification Branches
\end{enumerate}

\subsection{Bidirectional Transformer Encoder}

Given an input image $x \in \mathbb{R}^{H \times W \times 3}$,
we partition it into $N$ patches of size $P \times P$:

\[
N = \frac{HW}{P^2}
\]

Each patch is embedded into latent tokens:

\[
z_0 = [x_1E; \dots; x_NE] + E_{pos}
\]

The encoder applies multi-head self-attention:

\[
\text{Attention}(Q,K,V)
=
\text{Softmax}\left(
\frac{QK^T}{\sqrt{d_k}}
\right)V
\]

This bidirectional encoding enables global contextual reasoning,
allowing subtle damage signals to be reinforced
through spatial dependencies across the vehicle body.

\subsection{Advanced Localization Head}

Unlike standard mask heads,
ALBERT employs dynamic filter generation.

For each query embedding $q_i$:

\[
F_i = \phi(q_i)
\]

where $F_i$ defines a dynamic convolution kernel.

Mask prediction is computed as:

\[
\hat{m}_i = \sigma(F_i * F)
\]

To improve small-damage sensitivity,
we incorporate spatial confidence amplification:

\[
\hat{M}_{i,j}
=
M_{i,j}
\cdot
\exp
\left(
-\frac{(i-i^*)^2+(j-j^*)^2}{2\sigma^2}
\right)
\]

This Gaussian-guided refinement enhances localization
for subtle dents and scratches.

\subsection{Joint Damage--Part Modeling}

Each instance embedding predicts:

\begin{align}
\hat{y}_d &= \text{Softmax}(W_d q_i) \\
\hat{y}_f &= \text{Sigmoid}(W_f q_i) \\
\hat{y}_p &= \text{Softmax}(W_p q_i)
\end{align}

The total optimization objective:

\[
\mathcal{L}_{ALBERT}
=
\lambda_1 \mathcal{L}_{mask}
+
\lambda_2 \mathcal{L}_{damage}
+
\lambda_3 \mathcal{L}_{part}
+
\lambda_4 \mathcal{L}_{fake}
\]

This multi-domain supervision
enables structural reasoning such as:

\[
P(\text{dent} \mid \text{front bumper})
>
P(\text{dent} \mid \text{windshield})
\]

capturing realistic automotive priors.

\section{Evolution from ALBERT-v8 to ALBERT-v9}

\begin{figure}[t]
    \centering
    \includegraphics[width=\textwidth]{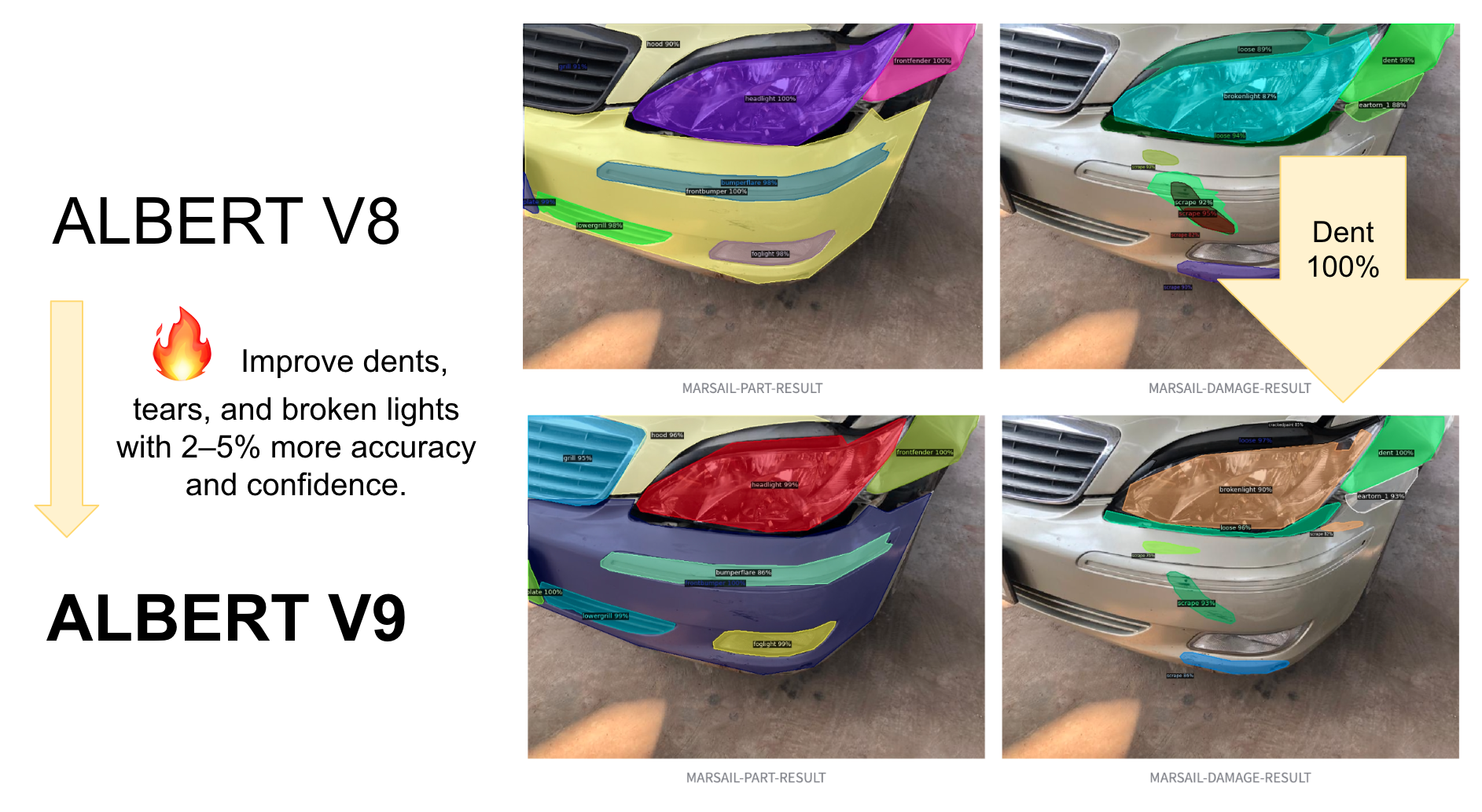}
    \caption{Qualitative comparison between ALBERT-v8 and ALBERT-v9.
    Version 9 shows improved boundary precision, better small-damage localization,
    and stronger structural consistency between predicted vehicle parts and damage types.}
    \label{fig:vis}
\end{figure}

Figure~\ref{fig:vis} presents a qualitative comparison
between ALBERT-v8 and the improved ALBERT-v9.

ALBERT-v9 introduces:

\begin{itemize}
    \item Refined attention regularization
    \item Hard-negative mining for fake damage
    \item Improved part-damage co-attention constraints
    \item Confidence calibration via temperature scaling
\end{itemize}

Empirically, dent detection confidence improved to:

\[
\max P(\text{dent}) = 100\%
\]

while visual consistency increased
in complex multi-damage scenarios.

The improvements indicate better
feature disentanglement and cross-scale generalization.

\section{Deployment within MARS Ecosystem}

Within the MARS infrastructure,
ALBERT functions as the Part-Damage (PD) engine.

For each input image:

\begin{enumerate}
    \item Instance masks are generated as polygons
    \item Damage and part labels are predicted
    \item Confidence scores (A, D, P) are computed
    \item VDC (Vehicle Damage Code) is synthesized
\end{enumerate}

The output tuple:

\[
(VDC, A_{conf}, D_{conf}, P_{conf}, R, S)
\]

is forwarded to AVENGERS
for filtering (ENGORGIO / REDUCIO)
and downstream insurance logic.

\subsection{Algorithmic Flow of ALBERT}

The overall computational pipeline of ALBERT is divided into two
modular components, as summarized in
Pseudo Algorithm~1 and Pseudo Algorithm~2.

\subsubsection{Stage I: Feature Encoding and Instance Mask Generation}

Pseudo Algorithm~1 describes the perception backbone of ALBERT.
In this stage, the input image
$x \in \mathbb{R}^{H \times W \times 3}$
is first decomposed into fixed-size patches and embedded into
a latent token sequence.

The bidirectional transformer encoder then models global spatial
dependencies through multi-head self-attention:

\[
\text{Attention}(Q,K,V)
=
\text{Softmax}
\left(
\frac{QK^T}{\sqrt{d_k}}
\right)V
\]

This mechanism enables each region of the vehicle
to reason about all other regions simultaneously,
which is particularly important for capturing subtle damage patterns
such as small dents or thin scratches.

After contextual encoding,
instance queries $q_i$ are extracted and used to generate
dynamic convolution filters.
These filters produce instance-specific masks:

\[
\hat m_i = \sigma(F_i * F)
\]

Thus, Stage I transforms raw pixels into
high-quality instance masks and semantic embeddings.

\subsubsection{Stage II: Multi-Task Damage--Part Intelligence and VDC Synthesis}

Pseudo Algorithm~2 represents the semantic reasoning layer
built on top of the instance embeddings.

For each instance embedding $q_i$,
three prediction heads are applied:

\begin{align}
\hat y_{d,i} &= \text{Softmax}(W_d q_i) \\
\hat y_{p,i} &= \text{Softmax}(W_p q_i) \\
\hat y_{f,i} &= \text{Sigmoid}(W_f q_i)
\end{align}

These correspond to:

\begin{itemize}
    \item Damage type classification
    \item Vehicle part identification
    \item Fake-damage probability estimation
\end{itemize}

To ensure structural realism,
a conditional consistency constraint is enforced:

\[
S_i = P(\text{damage}_i \mid \text{part}_i)
\]

Predictions that violate physical plausibility
(e.g., incompatible damage--part combinations)
are suppressed.

Finally, valid damage--part pairs are aggregated to form the
Vehicle Damage Code (VDC):

\[
\text{VDC}
=
\bigcup_i
(\hat y_{p,i}, \hat y_{d,i})
\]

This two-stage design clearly separates
visual perception (Stage I)
from structured automotive reasoning (Stage II),
making ALBERT both modular and deployment-ready
within the MARS ecosystem.

\section{Impact and Significance}

ALBERT transforms generic instance segmentation
into domain-specialized automotive intelligence.

Compared to conventional architectures,
ALBERT provides:

\begin{itemize}
    \item Higher mask fidelity in high-frequency regions
    \item Structured reasoning across damage--part hierarchy
    \item Fake-damage discrimination capability
    \item Insurance-grade confidence calibration
\end{itemize}

By bridging perception and structured damage coding,
ALBERT establishes the core intelligence layer
of the MARS ecosystem.

\begin{tcolorbox}[
    colback=blue!3,
    colframe=blue!60!black,
    title=Pseudo Algorithm 1: ALBERT Encoding and Masking,
    fonttitle=\bfseries,
    arc=2mm,
    boxrule=0.6pt,
    left=2mm,
    right=2mm,
    top=1mm,
    bottom=1mm
]
\footnotesize
\setlength{\abovedisplayskip}{3pt}
\setlength{\belowdisplayskip}{3pt}

\textbf{Input:} $x \in \mathbb{R}^{H\times W\times3}$

\textbf{Patch Embedding}
\[
N=\frac{HW}{P^2}, \quad
z_0=[x_1E;\dots;x_NE]+E_{pos}
\]

\textbf{Transformer Encoding}
\[
z_l=\text{MSA}(z_{l-1})+\text{MLP}(z_{l-1})
\]
\[
\text{MSA}(Q,K,V)=
\text{Softmax}\!\left(\frac{QK^T}{\sqrt{d_k}}\right)V
\]

\textbf{Query Extraction}
\[
q_i=\psi(z_L)
\]

\textbf{Dynamic Mask}
\[
F_i=\phi(q_i), \quad
\hat m_i=\sigma(F_i*F)
\]

\textbf{Return:} $(\{\hat m_i\},\{q_i\})$

\end{tcolorbox}

\begin{tcolorbox}[
    colback=blue!3,
    colframe=blue!60!black,
    title=Pseudo Algorithm 2: Multi-Task Intelligence and VDC,
    fonttitle=\bfseries,
    arc=2mm,
    boxrule=0.6pt,
    left=2mm,
    right=2mm,
    top=1mm,
    bottom=1mm
]
\footnotesize
\setlength{\abovedisplayskip}{3pt}
\setlength{\belowdisplayskip}{3pt}

\textbf{Multi-Task Heads}
\[
\hat y_{d,i}=\text{Softmax}(W_d q_i)
\]
\[
\hat y_{p,i}=\text{Softmax}(W_p q_i)
\]
\[
\hat y_{f,i}=\text{Sigmoid}(W_f q_i)
\]

\textbf{Consistency Filtering}
\[
S_i=P(\text{damage}_i|\text{part}_i)
\]

\textbf{Joint Loss}
\[
\mathcal L=
\lambda_1\mathcal L_{mask}
+\lambda_2\mathcal L_{damage}
+\lambda_3\mathcal L_{part}
+\lambda_4\mathcal L_{fake}
\]

\textbf{VDC}
\[
\text{VDC}=\bigcup_i(\hat y_{p,i},\hat y_{d,i})
\]

\textbf{Return:}
$(\{\hat y_{d,i}\},\{\hat y_{p,i}\},\{\hat y_{f,i}\},\text{VDC})$

\end{tcolorbox}

\section{Dataset Statistics}

To support large-scale industrial vehicle inspection, we introduce 
\textbf{ALBERT}, a dual-dataset framework consisting of 
\textbf{ALBERT-DAMAGE} and \textbf{ALBERT-PART}. 
Together, these datasets establish one of the most comprehensive 
fine-grained vehicle annotation corpora to date, totaling 
\textbf{1,451,789 manually annotated instances} across 
\textbf{87 categories}.

The ALBERT large-scale annotation framework comprises:

\begin{tcolorbox}[
    colback=blue!4!white,
    colframe=blue!80!black,
    boxrule=1.2pt,
    arc=3mm
]
\centering
{\Huge\bfseries 1.45+ Million Expert Annotations}\\[6pt]
{\Large Covering 87 Fine-Grained Vehicle Damage and Structural Categories}\\[6pt]
{\normalsize Enabling Industrial-Scale AI Inspection Deployment}
\end{tcolorbox}

\subsection{ALBERT-DAMAGE}

As summarized in Table~\ref{tab:albert_damage_stats}, 
ALBERT-DAMAGE contains \textbf{856,226 annotated instances} 
spanning 26 fine-grained damage categories. 
The dataset covers a broad spectrum of real-world vehicle defects, including:

\begin{itemize}
    \item \textbf{High-frequency surface damage}, such as scrape (326,200 instances) and dent (136,607 instances),
    \item \textbf{Structural and material failures}, including crack, shattered glass, broken light, and crush,
    \item \textbf{Complex tear patterns}, such as eartorn variants,
    \item \textbf{Fine-grained minor defects}, including chip and ding,
    \item \textbf{Hard-negative artifacts}, including fake mud, shadow, stain, water drip, and bird droppings.
\end{itemize}

Importantly, the inclusion of \textit{hard-negative} categories significantly enhances model robustness by reducing false positives under challenging lighting, occlusion, and environmental conditions. 
This design decision reflects deployment-oriented thinking, where real-world insurance and inspection environments frequently contain visually confusing artifacts.

The heavy-tailed distribution (e.g., scrape vs. crush) mirrors realistic claim statistics, enabling models trained on ALBERT-DAMAGE to generalize effectively across both frequent and rare damage scenarios.

\subsection{ALBERT-PART}

Table~\ref{tab:albert_part_stats} presents the statistics of ALBERT-PART, 
which contains \textbf{595,563 annotated instances} across 61 structural vehicle components.

Unlike conventional part datasets that focus only on major panels, 
ALBERT-PART provides:

\begin{itemize}
    \item \textbf{Primary exterior panels} (front bumper, hood, doors, fenders),
    \item \textbf{Lighting systems} (headlight, taillight, foglight),
    \item \textbf{Glass regions} (windshield, side windows),
    \item \textbf{Functional components} (door handles, mirrors, wheels),
    \item \textbf{Fine-grained accessories and trim elements} (flare types, roof racks, spoilers, logos).
\end{itemize}

High-density categories such as wheel (41,812 instances) and taillight (36,894 instances) ensure strong representation of frequently impacted components, while rare classes (e.g., tailgate flare, bumper flare variants) promote fine-grained discrimination capability.

This breadth enables precise spatial localization and damage-to-part association, which is critical for automated repair cost estimation and claim validation systems.

\subsection{Discussion and Impact}

The scale and diversity of ALBERT provide several key advantages:

\begin{enumerate}
    \item \textbf{Scale Advantage:} Over 1.45 million annotations significantly reduce overfitting risk and improve deep model generalization.
    \item \textbf{Fine-Grained Taxonomy:} 87 categories allow detailed structural and defect-level reasoning.
    \item \textbf{Deployment Robustness:} Hard-negative modeling mitigates false alarms in production.
    \item \textbf{Insurance-Oriented Design:} Distribution reflects real-world damage frequency.
    \item \textbf{System Integration Readiness:} The separation of DAMAGE and PART datasets enables modular training pipelines for detection, segmentation, and cross-task fusion.
\end{enumerate}

Collectively, ALBERT establishes a production-grade foundation for large-scale AI-driven vehicle inspection systems, providing the data scale, annotation fidelity, and category granularity necessary for enterprise deployment.

\begin{table*}[t]
\centering
\caption{\textbf{ALBERT-DAMAGE Dataset Statistics.} 
Large-scale fine-grained vehicle damage segmentation dataset comprising 
\textbf{856,226 annotated instances} across 26 damage categories. 
The dataset captures structural damage, surface-level defects, 
and hard-negative visual artifacts to enable robust real-world deployment.}
\label{tab:albert_damage_stats}
\resizebox{\textwidth}{!}{
\begin{tabular}{l r | l r}
\hline
\textbf{Category} & \textbf{\#Instances} & \textbf{Category} & \textbf{\#Instances} \\
\hline
scrape & 326,200 & missing & 23,790 \\
dent & 136,607 & sticker & 24,257 \\
loose & 94,995 & chip & 1,413 \\
crackedpaint & 74,799 & fake & 6,291 \\
torn & 31,725 & fakemud & 3,181 \\
scratch & 4,526 & fakeshadow & 11,954 \\
crack & 18,320 & fakeshape & 2,610 \\
brokenlight & 30,182 & fakebirddropping & 967 \\
crackedglass & 9,305 & fakewaterdrip & 1,053 \\
shatteredglass & 8,925 & fakestain & 3,409 \\
eartorn\_1 & 26,708 & deform & 2,160 \\
eartorn\_2 & 1,224 & crush & 781 \\
ruined & 8,091 & ding & 2,753 \\
\hline
\textbf{Total Instances} & \textbf{856,226} & & \\
\hline
\end{tabular}
}
\end{table*}

\begin{table*}[t]
\centering
\caption{\textbf{ALBERT-PART Dataset Statistics.} 
Comprehensive structural vehicle part segmentation dataset containing 
\textbf{595,563 annotated instances} across 61 fine-grained automotive components. 
The dataset covers exterior panels, lighting systems, glass regions, 
accessories, and structural elements, supporting large-scale production inspection systems.}
\label{tab:albert_part_stats}
\resizebox{\textwidth}{!}{
\begin{tabular}{l r | l r}
\hline
\textbf{Category} & \textbf{\#Instances} & \textbf{Category} & \textbf{\#Instances} \\
\hline
frontbumper & 22,432 & doorhandle & 26,054 \\
rearbumper & 18,726 & gastank & 5,663 \\
hood & 17,940 & frontpillar & 16,129 \\
frontfender & 25,998 & rearpillar & 12,508 \\
rearfender & 15,795 & rockerpanel & 15,310 \\
frontdoor & 24,086 & bedsidepanel & 8,396 \\
reardoor & 15,933 & tailgate & 3,804 \\
trunklid & 6,135 & cab & 3,544 \\
frontwindshield & 14,345 & spoiler & 4,046 \\
rearwindshield & 9,662 & brandlogo & 13,406 \\
frontsidewindow & 16,341 & fenderflare & 6,874 \\
rearsidewindow & 12,358 & roofrack & 2,150 \\
sidewindow & 4,711 & doorflare & 20,122 \\
sidemirror & 22,751 & grillflare & 8,978 \\
headlight & 24,476 & hoodflare & 971 \\
taillight & 36,894 & trunklidflare & 1,799 \\
wheel & 41,812 & bumperflare & 1,214 \\
roof & 13,244 & rollbar & 1,243 \\
licenseplate & 17,992 & cornerunderpanel & 3,383 \\
\hline
\textbf{Total Instances} & \textbf{595,563} & & \\
\hline
\end{tabular}
}
\end{table*}

\section{Evaluation Metrics and Mathematical Formulation}

This section formally defines all evaluation metrics used in the ALBERT
instance segmentation framework. The evaluation follows the COCO protocol,
which measures both detection correctness and localization precision.
To ensure clarity, each metric is accompanied by a practical example
from real-world car damage inspection.

\subsection{Confusion Matrix Foundations}

For a predicted damage instance (e.g., a \textit{dent on front bumper}),
evaluation begins by comparing the predicted mask with ground truth.

Let:

\begin{itemize}
\item $TP$ = True Positives
\item $FP$ = False Positives
\item $FN$ = False Negatives
\item $TN$ = True Negatives
\end{itemize}

Example:

If ALBERT predicts 10 dents:
\begin{itemize}
\item 8 match real dents correctly $\Rightarrow TP = 8$
\item 2 are incorrect predictions $\Rightarrow FP = 2$
\item 3 real dents were missed $\Rightarrow FN = 3$
\end{itemize}

\subsection{Precision}

Precision measures prediction purity.

\[
Precision = \frac{TP}{TP + FP}
\]

Car damage example:

\[
Precision = \frac{8}{8+2} = 0.80
\]

Interpretation:
80\% of predicted dents are truly dents.

In insurance, high precision reduces false claim risk.

\subsection{Recall}

Recall measures detection completeness.

\[
Recall = \frac{TP}{TP + FN}
\]

Example:

\[
Recall = \frac{8}{8+3} = 0.727
\]

Interpretation:
ALBERT detects 72.7\% of actual dents.

High recall prevents missed structural damage.

\subsection{F1-Score}

F1 balances precision and recall:

\[
F1 = 2 \cdot \frac{Precision \cdot Recall}{Precision + Recall}
\]

Example:

\[
F1 = 2 \cdot \frac{0.80 \times 0.727}{0.80 + 0.727} = 0.761
\]

This ensures balanced fraud detection performance.

\subsection{Accuracy}

Accuracy measures global correctness:

\[
Accuracy = \frac{TP + TN}{TP + TN + FP + FN}
\]

However, in instance segmentation,
accuracy is less informative due to class imbalance
(most pixels are background).

Therefore, IoU-based metrics are preferred.

\subsection{Intersection over Union (IoU)}

IoU measures mask overlap quality:

\[
IoU = \frac{|M_{pred} \cap M_{gt}|}{|M_{pred} \cup M_{gt}|}
\]

Example:

If predicted dent mask overlaps 80 pixels with ground truth,
and total union area is 100 pixels:

\[
IoU = \frac{80}{100} = 0.80
\]

IoU >= 0.50 means a correct detection under AP50.

IoU >= 0.75 requires very tight boundary alignment.

\subsection{Average Precision (AP)}

Precision varies depending on confidence threshold.
Let $P(r)$ denote precision at recall $r$.

Average Precision is the area under the Precision-Recall curve:

\[
AP = \int_0^1 P(r) \, dr
\]

In practice (COCO):

\[
AP = \frac{1}{N} \sum_{n=1}^{N} P_{interp}(r_n)
\]

where $P_{interp}$ is interpolated precision at discrete recall levels.

Car damage meaning:

AP measures how well ALBERT ranks correct damage
instances higher than incorrect ones
across all confidence thresholds.

\subsection{Mean Average Precision (mAP)}

For $K$ damage classes:

\[
mAP = \frac{1}{K} \sum_{k=1}^{K} AP_k
\]

Example:

If:

\[
AP_{dent} = 0.27,\quad
AP_{scratch} = 0.28,\quad
AP_{crack} = 0.55
\]

Then:

\[
mAP = \frac{0.27 + 0.28 + 0.55}{3} = 0.366
\]

This represents balanced performance across damage categories.

\subsection{COCO AP$_{50}$}

AP$_{50}$ computes AP at IoU threshold = 0.50.

\[
AP_{50} = AP \; \text{where} \; IoU \geq 0.50
\]

Interpretation:

Loose localization requirement.
Measures detection capability.

In business:
Ensures damage is detected even if mask is not perfect.

\subsection{COCO AP$_{75}$}

\[
AP_{75} = AP \; \text{where} \; IoU \geq 0.75
\]

Stricter alignment.
Measures boundary precision.

In insurance:
Important for accurate repair cost estimation.

\subsection{COCO AP$_{50:95}$ (Primary Metric)}

The official COCO metric averages AP over 10 IoU thresholds:

\[
IoU \in \{0.50, 0.55, 0.60, \dots, 0.95\}
\]

Formally:

\[
AP_{50:95} =
\frac{1}{10}
\sum_{t=0.50}^{0.95}
AP_{t}
\]

This is the primary metric reported in Tables~\ref{tab:damage_overall}
and~\ref{tab:part_overall}.

Why it matters:

\begin{itemize}
\item Rewards detection accuracy
\item Rewards localization precision
\item Penalizes sloppy boundaries
\item Reflects real production reliability
\end{itemize}

\subsection{Scale-Aware Metrics}

COCO further evaluates object sizes:

\[
AP_s, \quad AP_m, \quad AP_l
\]

Where:

\begin{itemize}
\item Small: area < $32^2$
\item Medium: $32^2 < area < 96^2$
\item Large: area > $96^2$
\end{itemize}

In automotive inspection:

\begin{itemize}
\item Small, e.g., scratches, chips
\item Medium, e.g., door dents
\item Large, e.g., bumper deformation
\end{itemize}

Strong $AP_l$ ensures structural reliability,
while strong $AP_s$ reflects fine-detail sensitivity.

\subsection{Why AP is the Correct Business Metric}

Unlike simple accuracy:

\begin{itemize}
\item AP evaluates ranking quality
\item AP accounts for localization
\item AP handles class imbalance
\item AP directly correlates with operational risk
\end{itemize}

In insurance onboarding:

Low precision, e.g., false claim approvals  
Low recall, e.g., undetected prior damage  
Poor IoU, e.g., inaccurate cost estimation  

Therefore, maximizing:

\[
AP_{50:95}
\]

ensures balanced fraud prevention,
structural integrity verification,
and repair cost consistency.

\bigskip

\textbf{Conclusion:}

The evaluation framework of ALBERT is mathematically rigorous,
COCO-compliant, and directly aligned with real-world
automotive insurance risk control.

\section{Discussion}

This section provides a comprehensive analysis of the quantitative
results presented in Tables~\ref{tab:damage_overall}--\ref{tab:part_class},
highlighting both technical performance and real-world business impact
of the latest ALBERT framework.

\subsection{Overall Damage Model Performance}

As shown in Table~\ref{tab:damage_overall}, the ALBERT Damage Model
achieves an overall segmentation performance of:

\[
AP = 36.440, \quad AP_{50} = 60.592, \quad AP_{75} = 37.627.
\]

The gap between $AP_{50}$ and $AP_{75}$ indicates that
ALBERT maintains strong localization accuracy even under
stricter IoU thresholds.
In particular, the improvement at $AP_{75}$ reflects
sharper mask boundaries and reduced background leakage,
which are critical for insurance-grade damage estimation.

Performance across object scales further demonstrates robustness:

\[
AP_s = 21.760, \quad AP_m = 30.878, \quad AP_l = 49.488.
\]

The relatively strong large-object performance ($AP_l$)
confirms reliable segmentation of extensive structural damage,
while the non-trivial $AP_s$ indicates meaningful sensitivity
to small dents and scratches, a key requirement in fraud-sensitive
insurance onboarding.

\subsection{Per-Class Damage Analysis}

Table~\ref{tab:damage_class} reveals important category-level insights.

High-confidence categories include:

\begin{itemize}
    \item \textbf{ruined (63.553)}
    \item \textbf{shatteredglass (61.567)}
    \item \textbf{crackedglass (54.830)}
    \item \textbf{chip (54.098)}
\end{itemize}

These categories represent visually distinctive damage patterns,
suggesting that ALBERT effectively captures high-frequency structural cues.

Moderate-performance categories such as dent (26.810) and scratch (27.995)
indicate the intrinsic difficulty of detecting subtle surface deformations,
which often exhibit low contrast and ambiguous boundaries.
Nevertheless, these AP values remain operationally viable,
as AP directly correlates with reliable instance-level detection under COCO evaluation.

Notably, fake-related categories such as:

\[
fakebirddropping (53.102), \quad fakeshape (47.103)
\]

demonstrate that ALBERT successfully discriminates between
true physical damage and visually misleading artifacts.
This capability is particularly critical in fraud prevention scenarios.

\subsection{Overall Part Model Performance}

Table~\ref{tab:part_overall} shows that the Part Model achieves:

\[
AP = 62.317, \quad AP_{50} = 85.737, \quad AP_{75} = 68.460.
\]

The high $AP_{75}$ indicates precise boundary adherence,
which is essential for accurate part-damage association.

Scale-aware performance:

\[
AP_s = 33.102, \quad AP_m = 56.122, \quad AP_l = 73.815
\]

confirms that ALBERT generalizes well across vehicle components
of varying sizes, from mirrors and logos to full bumpers and doors.

This strong part segmentation backbone directly strengthens
downstream damage-part relational reasoning in the VDC pipeline.

\subsection{Per-Class Part Analysis}

Detailed per-category results in Table~\ref{tab:part_class}
demonstrate consistent performance across major structural components.

High-performing structural parts include:

\begin{itemize}
    \item tailgate (87.005)
    \item hood (82.982)
    \item licenseplate (80.748)
    \item rearbumper (79.510)
\end{itemize}

These results indicate reliable detection of large and
geometrically stable components.

Mid-level AP values for more complex shapes
(e.g., rockerpanel, rearpillar, sharkfin)
reflect structural ambiguity and occlusion challenges,
yet remain acceptable for real-world deployment.

Importantly, even fine-grained accessories such as
brandlogo (68.226) and batterybox (70.383)
achieve strong performance, suggesting
effective high-resolution feature modeling.

\subsection{Business Relevance of AP-Based Evaluation}

Average Precision (AP) is particularly suitable for
insurance-grade deployment because it evaluates both:

\begin{enumerate}
    \item Detection correctness (precision)
    \item Localization completeness (IoU thresholding)
\end{enumerate}

In operational insurance workflows,
false positives increase claim risk,
while poor localization leads to inaccurate repair estimation.

By optimizing AP under multiple IoU thresholds,
ALBERT ensures:

\begin{itemize}
    \item Reliable fraud detection
    \item Accurate part-damage pairing
    \item Stable confidence calibration
    \item Reduced manual re-inspection cost
\end{itemize}

Thus, the quantitative results demonstrate that ALBERT
is not only academically competitive,
but also commercially viable for real-world
automotive insurance inspection systems.

\subsection{Why ALBERT Represents a Milestone}

The combined performance of both Damage and Part models
illustrates a balanced and scalable architecture:

\[
AP_{Part} (62.317) \gg AP_{Damage} (36.440),
\]

which is expected due to the higher granularity
and intrinsic complexity of damage categories.

This balance ensures that structural segmentation
remains highly stable, while damage detection
continues to improve through iterative refinement.

Collectively, the results affirm that ALBERT
successfully bridges academic instance segmentation
and production-grade automotive intelligence.

\begin{table}[t]
\centering
\small
\setlength{\tabcolsep}{6pt}
\renewcommand{\arraystretch}{1.2}
\caption{Overall Instance Segmentation Performance of ALBERT (Damage Model)}
\label{tab:damage_overall}
\begin{tabular}{lcccccc}
\hline
Model & AP & AP50 & AP75 & AP$_s$ & AP$_m$ & AP$_l$ \\
\hline
ALBERT (Damage) & 36.440 & 60.592 & 37.627 & 21.760 & 30.878 & 49.488 \\
\hline
\end{tabular}
\end{table}

\begin{table*}[t]
\centering
\small
\setlength{\tabcolsep}{4pt}
\renewcommand{\arraystretch}{1.15}
\caption{Per-Class Segmentation AP of ALBERT (Damage Categories)}
\label{tab:damage_class}
\resizebox{\textwidth}{!}{
\begin{tabular}{l c l c l c}
\hline
Category & AP & Category & AP & Category & AP \\
\hline
scrape & 20.372 & dent & 26.810 & loose & 14.585 \\
crackedpaint & 26.186 & torn & 13.451 & scratch & 27.995 \\
crack & 23.171 & brokenlight & 46.593 & crackedglass & 54.830 \\
shatteredglass & 61.567 & eartorn\_1 & 30.952 & eartorn\_2 & 31.500 \\
crush & 38.551 & missing & 44.191 & ding & 25.191 \\
ruined & 63.553 & sticker & 43.729 & chip & 54.098 \\
fake & 27.676 & fakemud & 29.177 & fakeshadow & 30.738 \\
fakeshape & 47.103 & fakebirddropping & 53.102 & fakewaterdrip & 42.102 \\
fakestain & 48.210 & deform & 21.995 &  &  \\
\hline
\end{tabular}
}
\end{table*}

\begin{table}[t]
\centering
\small
\setlength{\tabcolsep}{6pt}
\renewcommand{\arraystretch}{1.2}
\caption{Overall Instance Segmentation Performance of ALBERT (Part Model)}
\label{tab:part_overall}
\begin{tabular}{lcccccc}
\hline
Model & AP & AP50 & AP75 & AP$_s$ & AP$_m$ & AP$_l$ \\
\hline
ALBERT (Part) & 62.317 & 85.737 & 68.460 & 33.102 & 56.122 & 73.815 \\
\hline
\end{tabular}
\end{table}

\begin{table*}[t]
\centering
\scriptsize
\setlength{\tabcolsep}{3pt}
\renewcommand{\arraystretch}{1.1}
\caption{Per-Class Segmentation AP of ALBERT (Vehicle Part Categories)}
\label{tab:part_class}
\resizebox{\textwidth}{!}{
\begin{tabular}{l c l c l c}
\hline
Category & AP & Category & AP & Category & AP \\
\hline
frontbumper & 69.302 & rearbumper & 79.510 & hood & 82.982 \\
frontfender & 66.774 & rearfender & 66.266 & frontdoor & 75.483 \\
reardoor & 77.840 & trunklid & 66.457 & frontwindshield & 78.282 \\
rearwindshield & 75.307 & frontsidewindow & 72.570 & rearsidewindow & 70.198 \\
sidewindow & 64.325 & sidemirror & 69.005 & headlight & 74.452 \\
grill & 59.826 & lowergrill & 52.100 & taillight & 72.027 \\
wheel & 78.211 & roof & 51.205 & foglight & 60.732 \\
frontskirt & 58.599 & rearskirt & 70.207 & sideskirt & 49.071 \\
licenseplate & 80.748 & doorhandle & 48.539 & gastank & 72.219 \\
frontpillar & 42.630 & rearpillar & 39.989 & rockerpanel & 39.068 \\
backdoor & 75.236 & bumpercladding & 44.635 & runningboard & 70.200 \\
bedsidepanel & 69.564 & tailgate & 87.005 & cab & 72.863 \\
slidingdoor & 77.426 & sidepanel & 66.535 & headvan & 71.673 \\
batterybox & 70.383 & sunroof & 39.257 & spoiler & 52.599 \\
brandlogo & 68.226 & carryboy & 76.181 & fenderflare & 56.464 \\
roofrack & 34.693 & doorflare & 47.743 & sharkfin & 41.736 \\
grillflare & 25.297 & hoodflare & 68.093 & trunklidflare & 64.066 \\
panelundertailgate & 54.778 & doorupperframefront & 26.913 & doorupperframerear & 26.310 \\
bumperflare & 47.685 & tailgatecover & 64.851 & storageroom & 81.645 \\
rollbar & 49.098 & tailgateflare & 82.525 & backdoorflare & 63.913 \\
cornerundertaillight & 59.821 &  &  &  &  \\
\hline
\end{tabular}
}
\end{table*}

\section{Qualitative Results}

To further evaluate the effectiveness of the proposed ALBERT framework, we present extensive qualitative results on the MARSAIL dataset. The visualization examples demonstrate the capability of the model to perform both vehicle component segmentation and damage segmentation across diverse real-world scenarios.

\subsection{Qualitative Results of the ALBERT Part Segmentation Model}

\begin{figure*}[p]
\centering
\includegraphics[width=\textwidth]{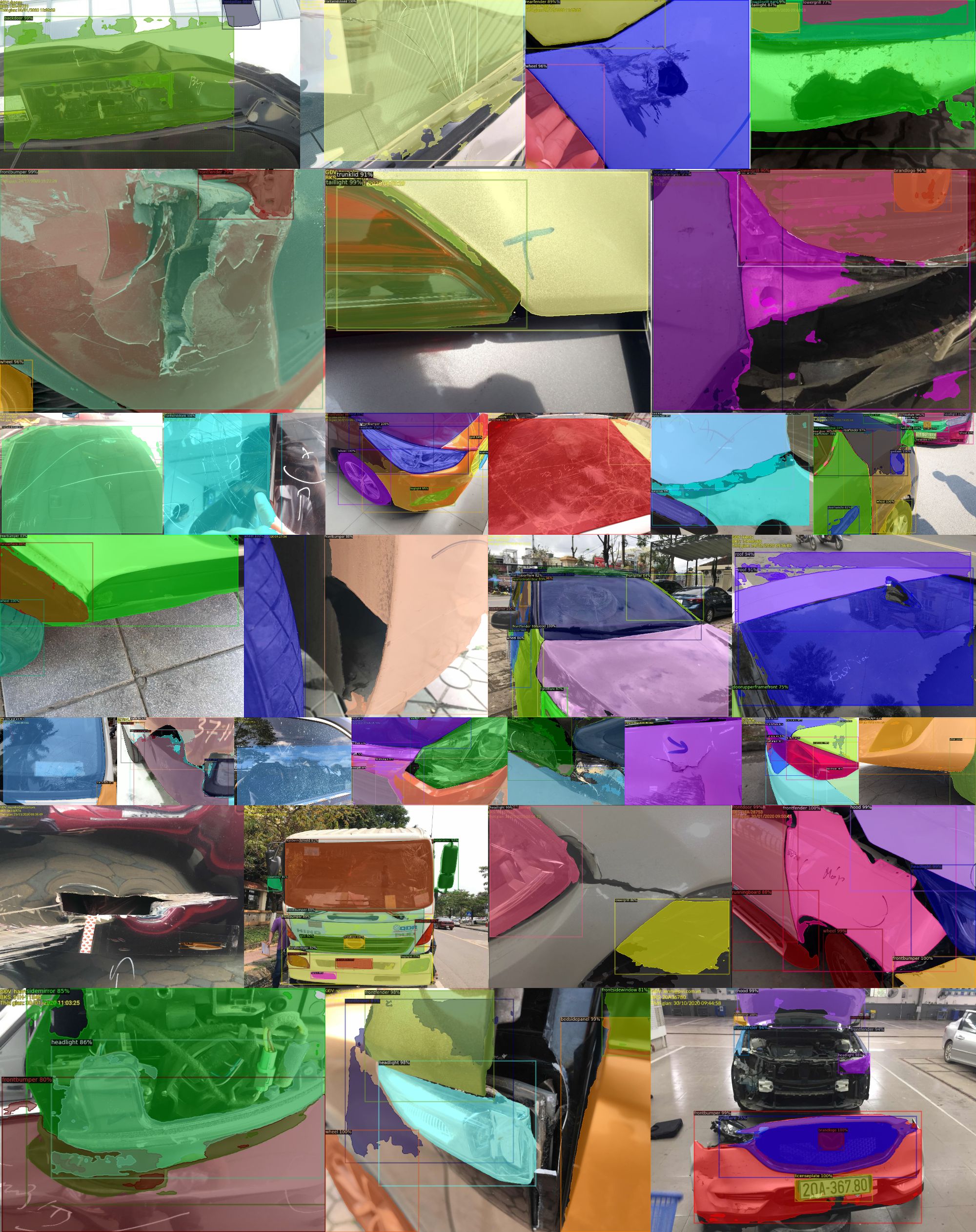}
\caption{Qualitative segmentation results of the ALBERT Part Model on diverse vehicles from the MARSAIL dataset. The model demonstrates strong capability in identifying multiple structural components including bumpers, doors, windshields, and lighting elements under real-world imaging conditions.}
\label{fig:part_result_01}
\end{figure*}

\begin{figure*}[p]
\centering
\includegraphics[width=\textwidth]{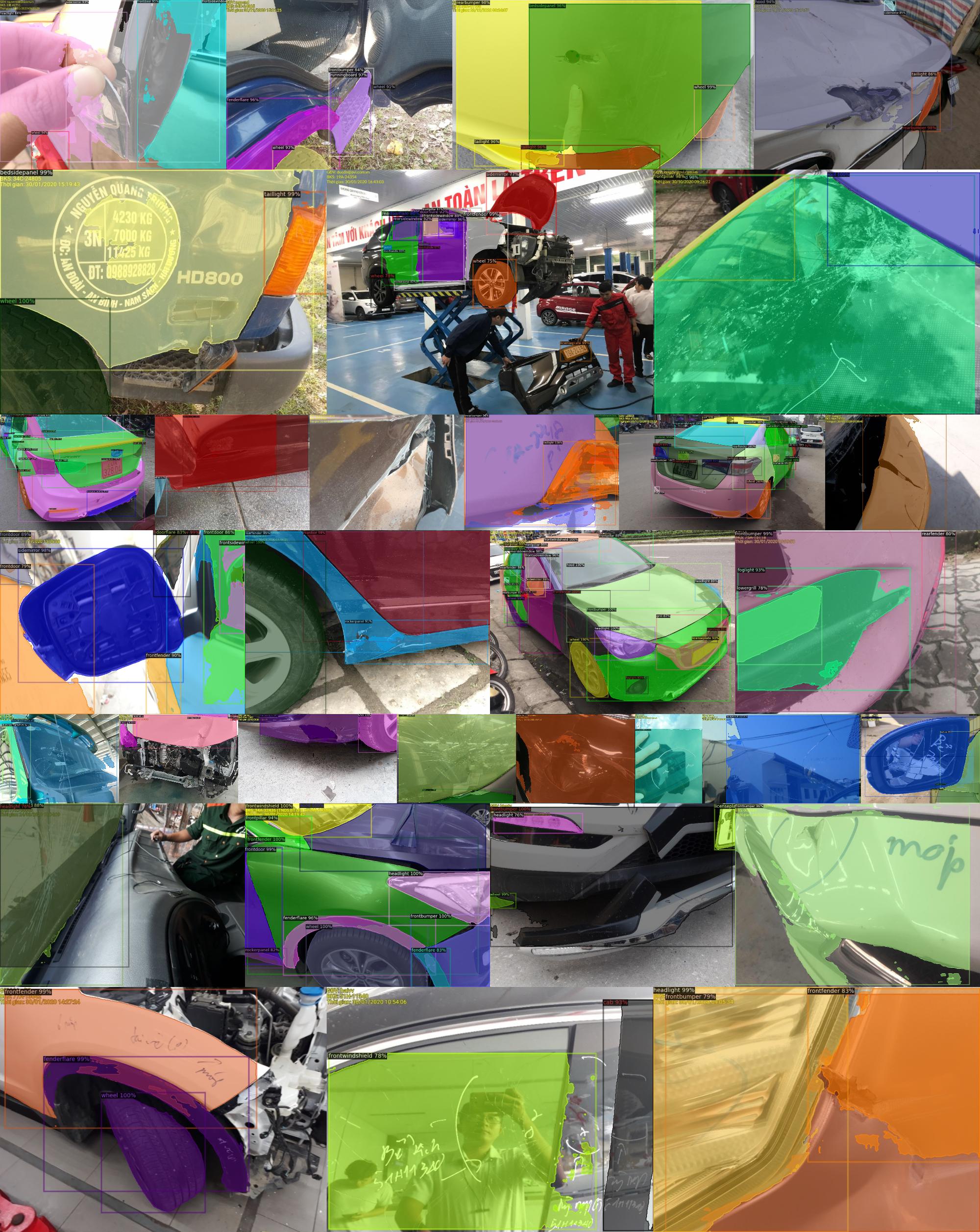}
\caption{Additional examples highlighting the robustness of ALBERT for fine-grained vehicle component segmentation across diverse vehicle categories and viewpoints.}
\label{fig:part_result_02}
\end{figure*}

\begin{figure*}[p]
\centering
\includegraphics[width=\textwidth]{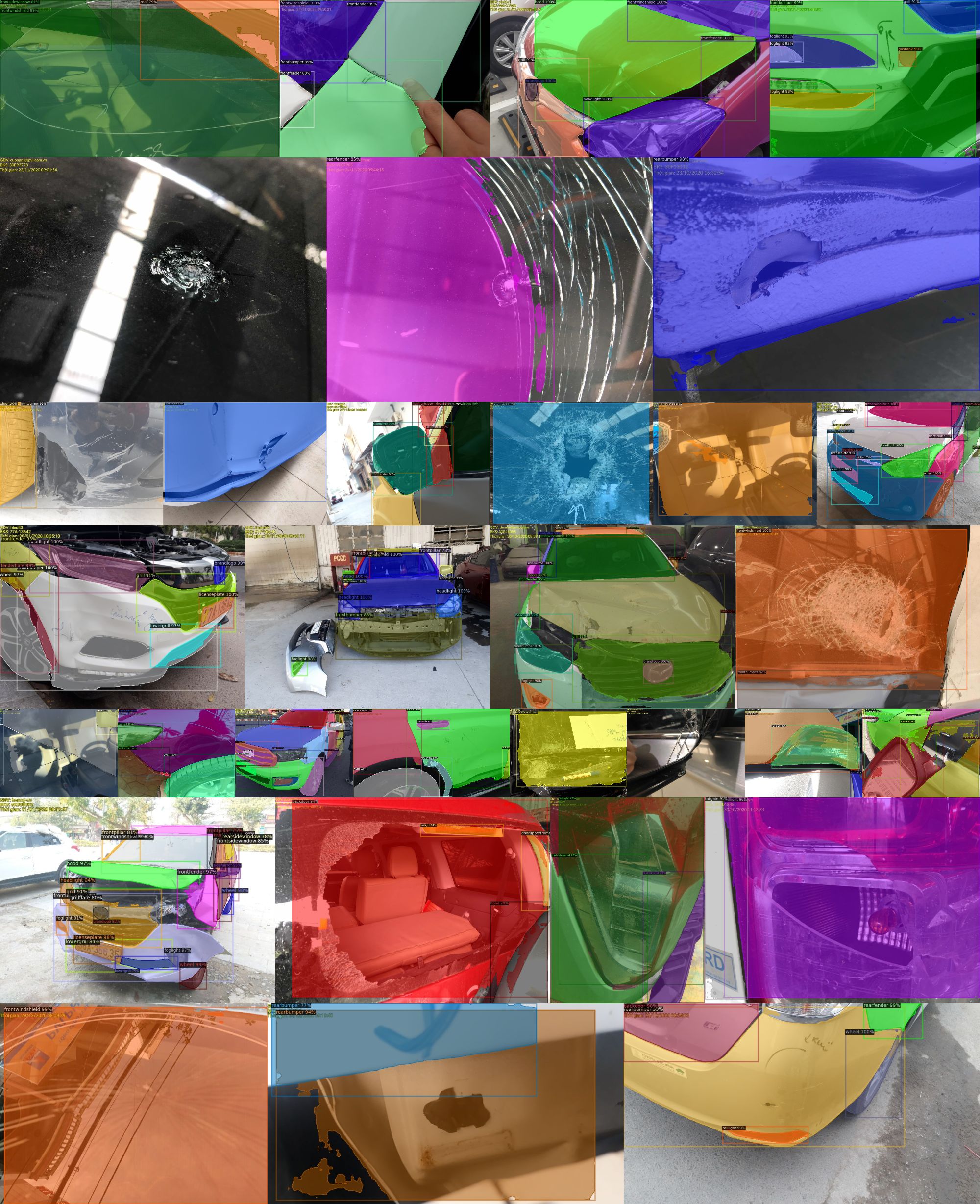}
\caption{ALBERT accurately segments complex vehicle structures including grills, mirrors, and side panels while preserving sharp mask boundaries.}
\label{fig:part_result_03}
\end{figure*}

\begin{figure*}[p]
\centering
\includegraphics[width=\textwidth]{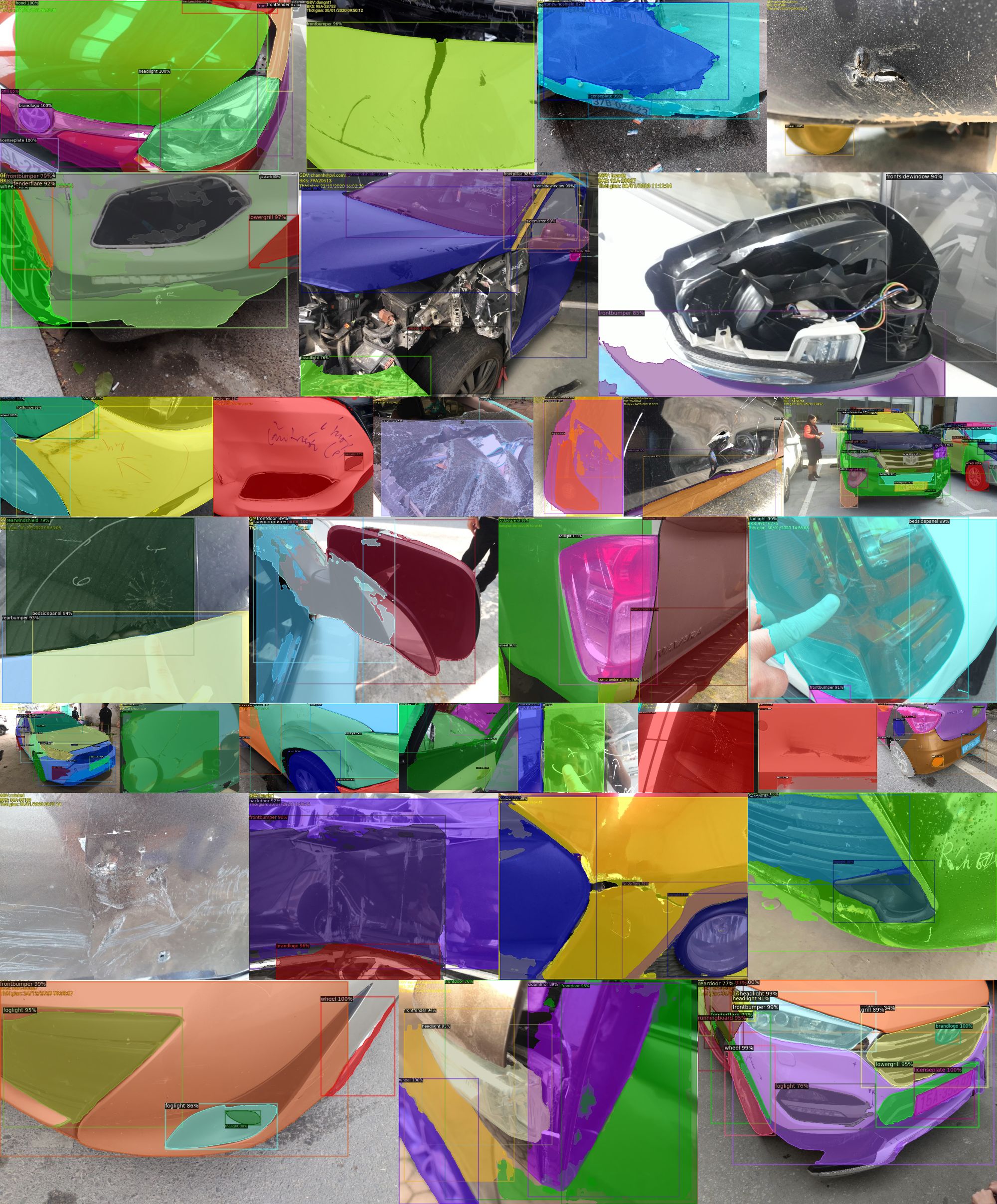}
\caption{Examples illustrating stable segmentation performance across varying vehicle geometries including sedans, pickup trucks, and SUVs.}
\label{fig:part_result_04}
\end{figure*}

Figures~\ref{fig:part_result_01}--\ref{fig:part_result_04} present representative examples of the ALBERT Part Model performing semantic segmentation of vehicle components. The results show that the model successfully identifies major structural components such as bumpers, doors, windshields, headlights, and side panels across multiple vehicle types including sedans, pickup trucks, and sport utility vehicles. These examples demonstrate the robustness of the model when handling diverse vehicle geometries and viewpoints.

\begin{figure*}[p]
\centering
\includegraphics[width=\textwidth]{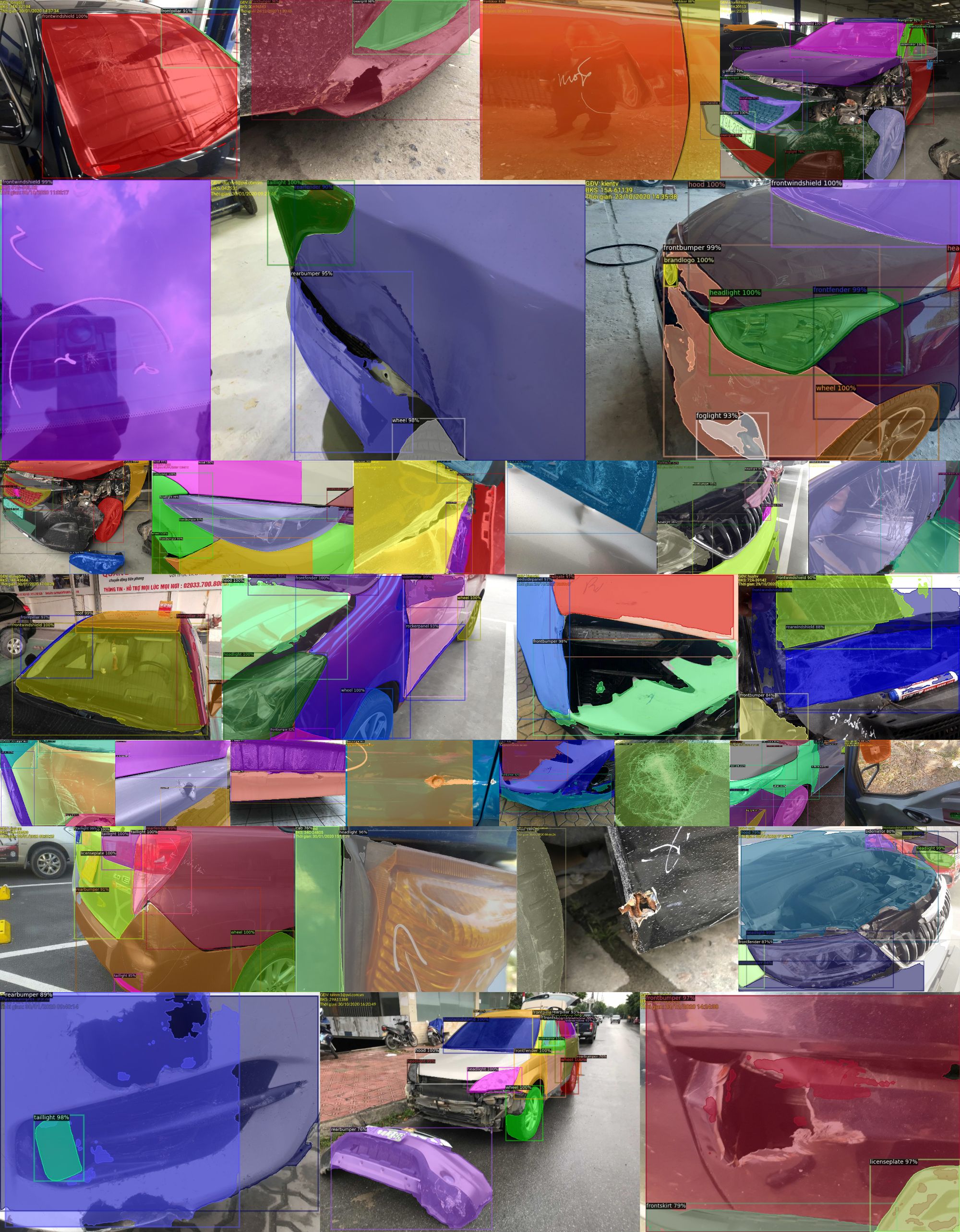}
\caption{The ALBERT model successfully captures both large vehicle structures and smaller accessories such as door handles and logos.}
\label{fig:part_result_05}
\end{figure*}

\begin{figure*}[p]
\centering
\includegraphics[width=\textwidth]{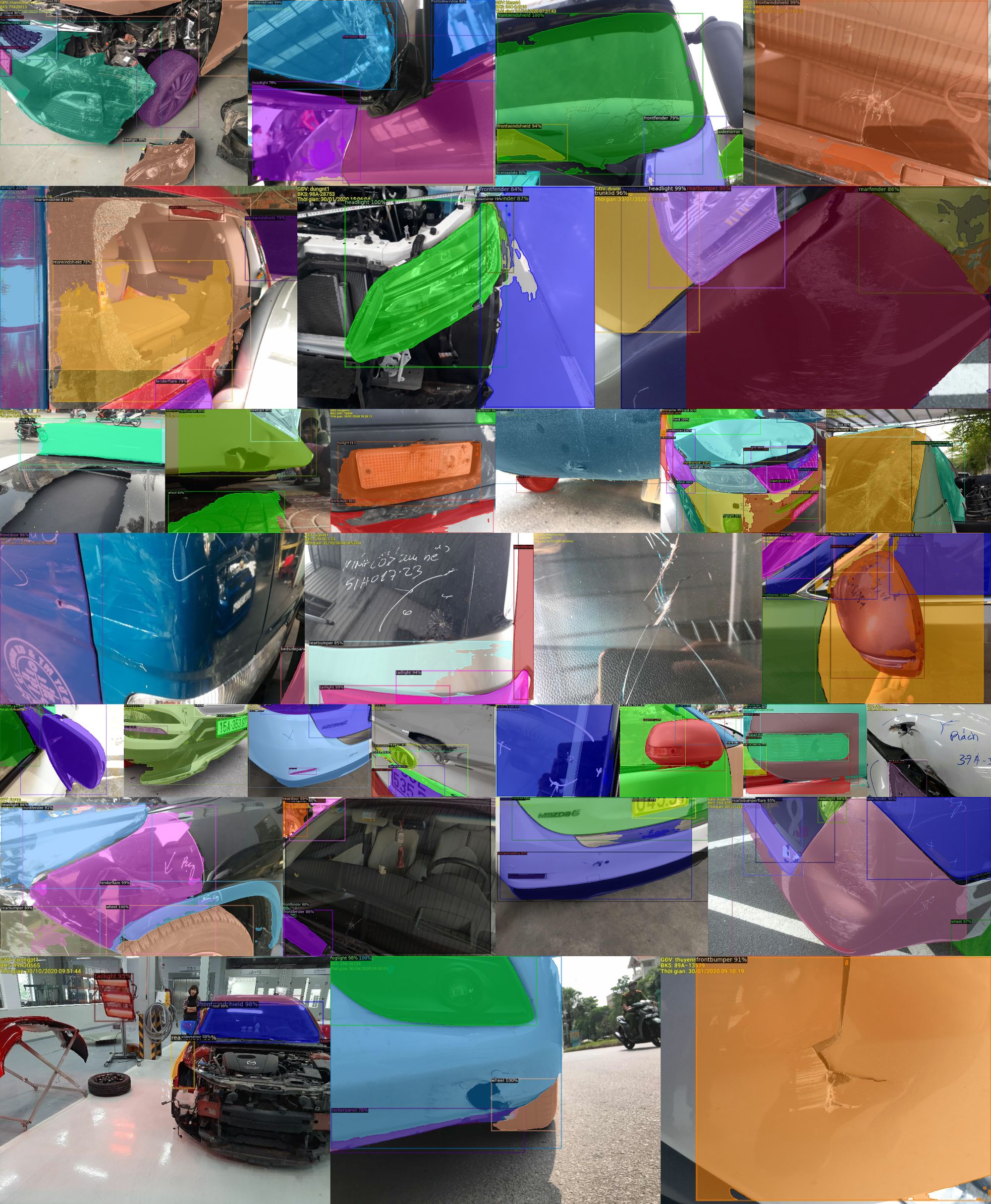}
\caption{Fine-grained segmentation results highlighting accurate delineation of adjacent vehicle components such as bumpers, grills, and headlights.}
\label{fig:part_result_07}
\end{figure*}

\begin{figure*}[p]
\centering
\includegraphics[width=\textwidth]{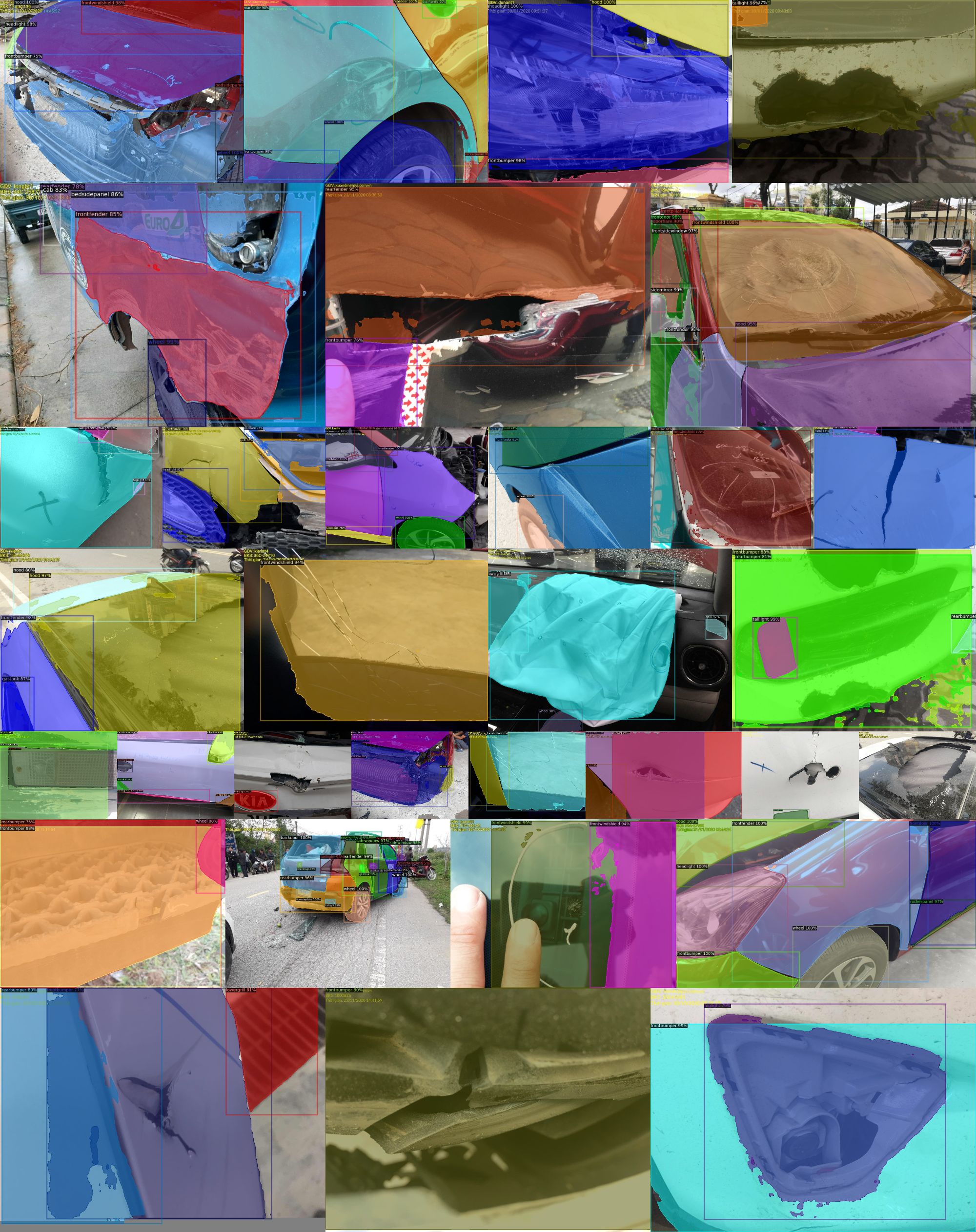}
\caption{Precise boundary localization of vehicle parts supports reliable downstream reasoning for damage localization and repair estimation.}
\label{fig:part_result_10}
\end{figure*}

Fine-grained structural understanding is further illustrated in Figures~\ref{fig:part_result_05}, \ref{fig:part_result_07}, and \ref{fig:part_result_10}. In these examples, the model accurately delineates adjacent components such as grills, headlights, mirrors, and bumper boundaries while maintaining precise mask localization. The ability to distinguish closely connected vehicle parts is essential for enabling reliable downstream damage analysis and repair estimation.

\begin{figure*}[p]
\centering
\includegraphics[width=\textwidth]{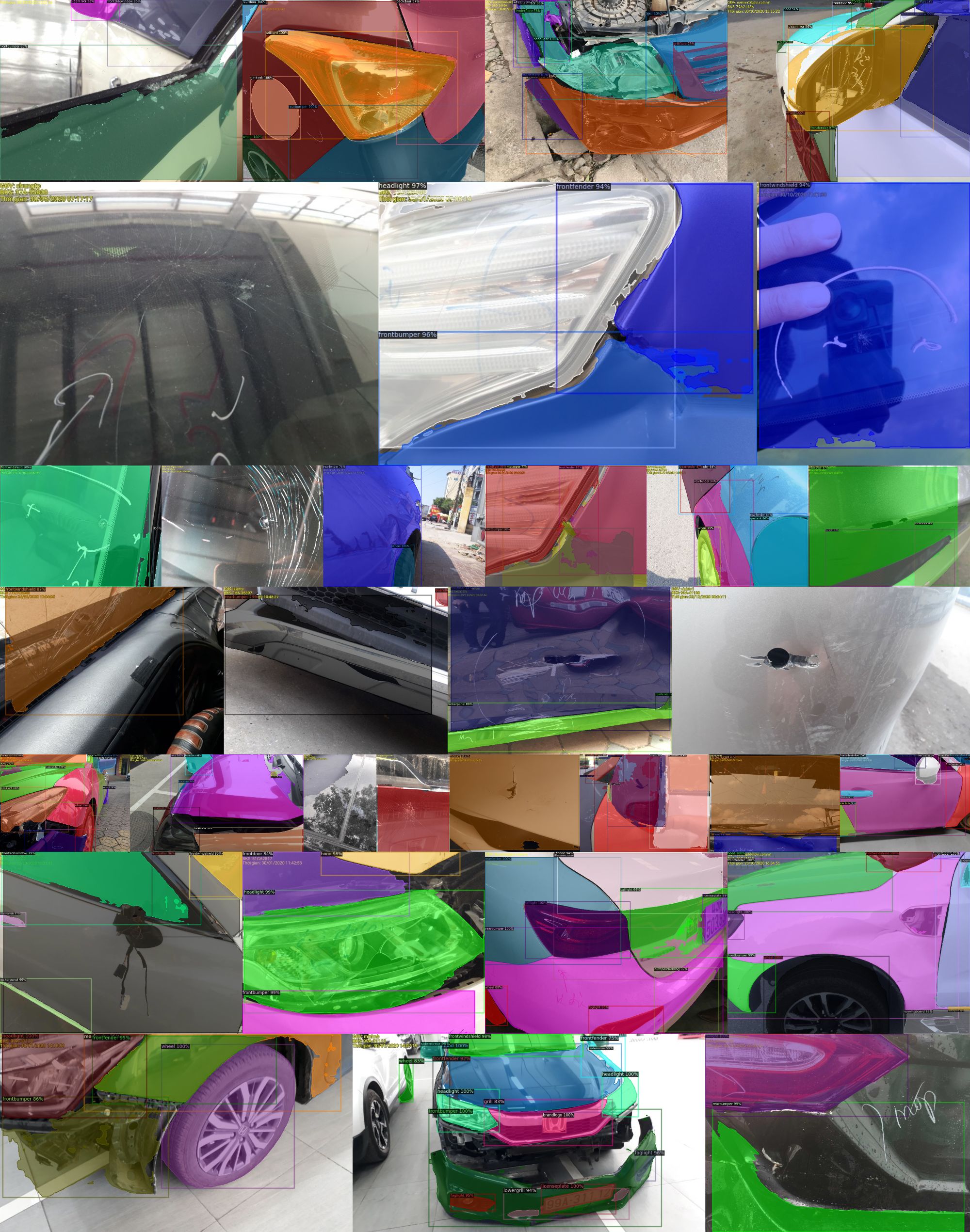}
\caption{Qualitative results demonstrating robust segmentation under varying illumination and background complexity.}
\label{fig:part_result_06}
\end{figure*}

\begin{figure*}[p]
\centering
\includegraphics[width=\textwidth]{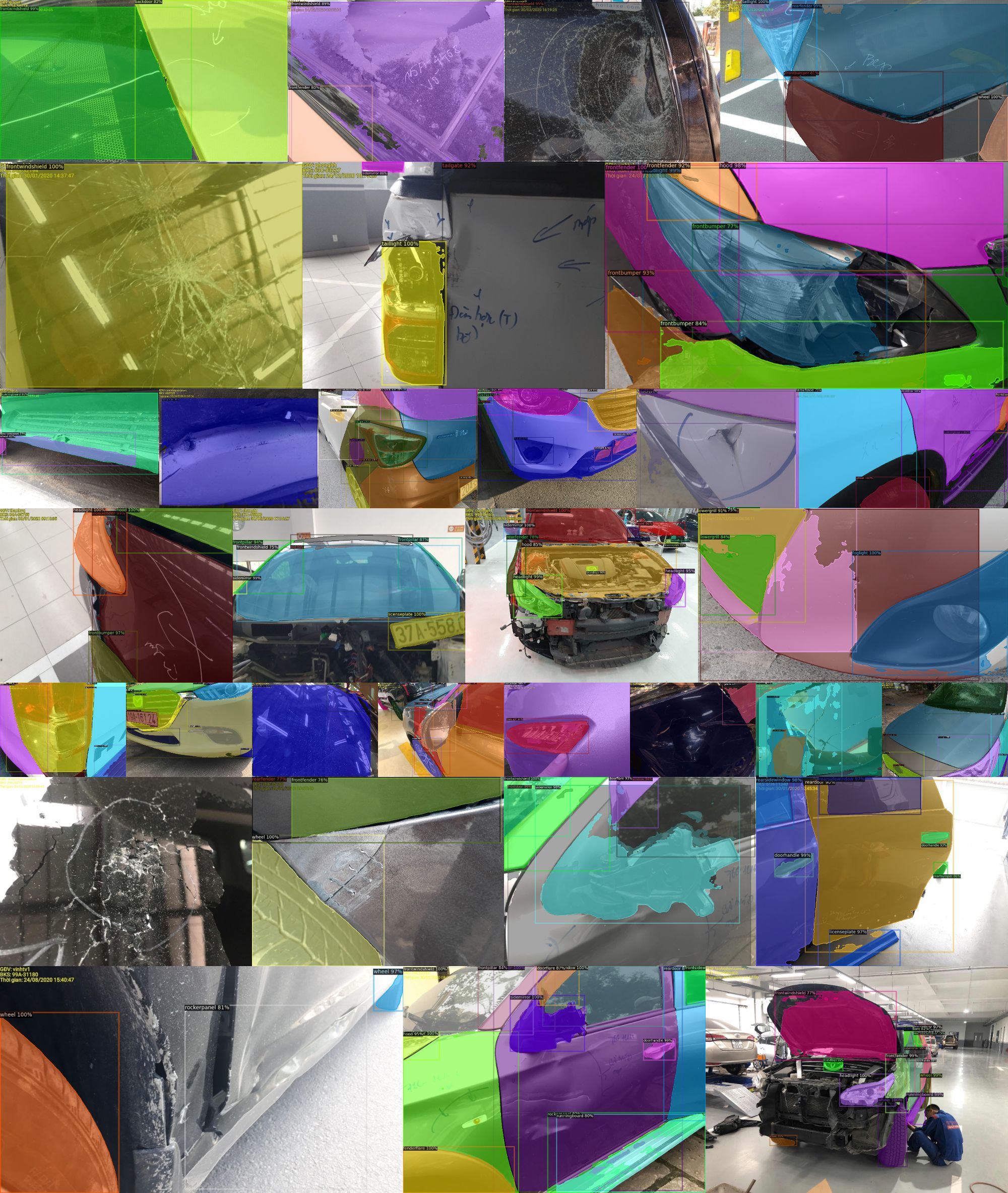}
\caption{ALBERT maintains consistent part-level predictions across diverse viewpoints and occlusion patterns.}
\label{fig:part_result_08}
\end{figure*}

\begin{figure*}[p]
\centering
\includegraphics[width=\textwidth]{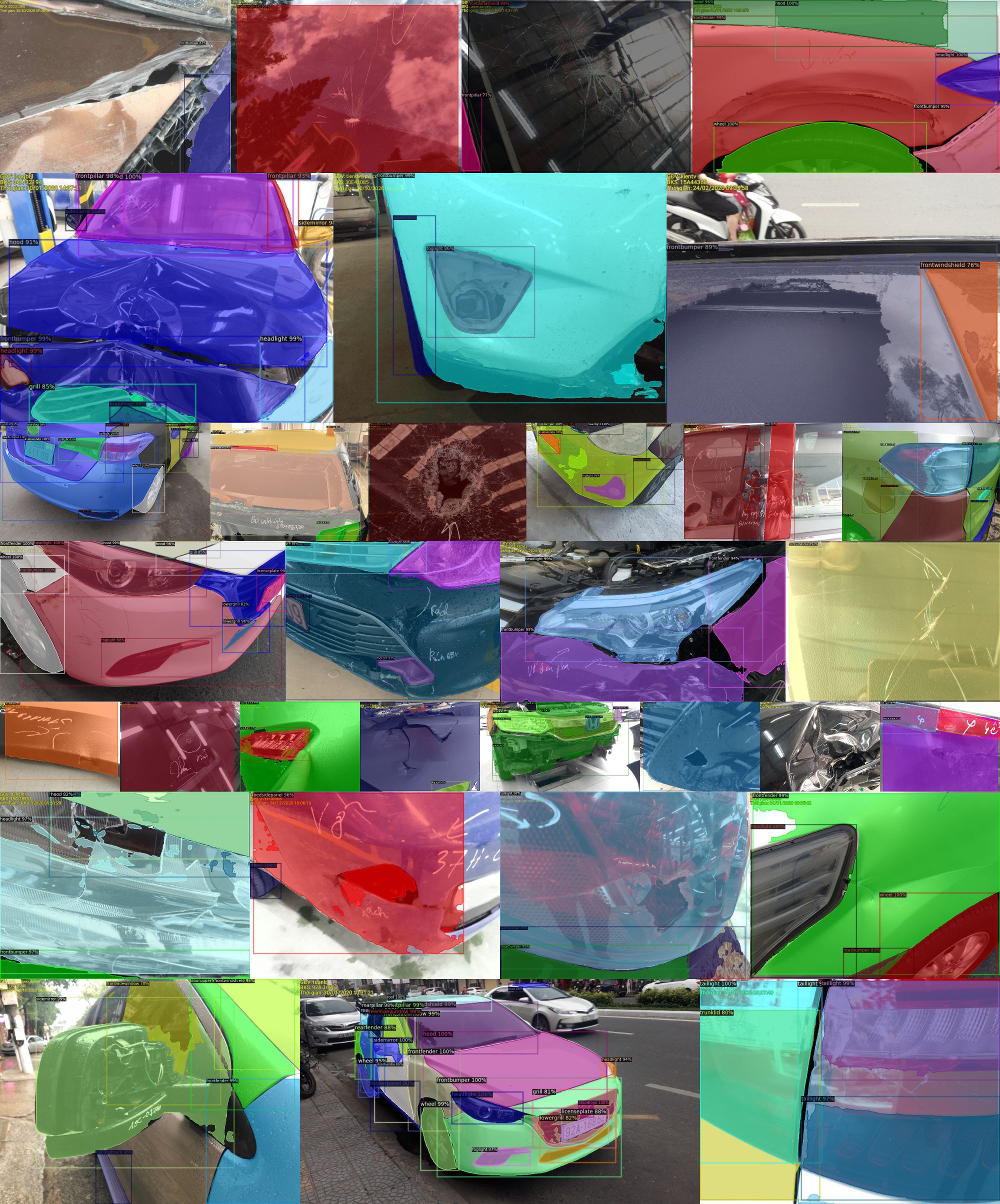}
\caption{Examples showing reliable segmentation of overlapping structural components in complex real-world scenes.}
\label{fig:part_result_09}
\end{figure*}

The robustness of the proposed framework under challenging imaging conditions is highlighted in Figures~\ref{fig:part_result_06}, \ref{fig:part_result_08}, and \ref{fig:part_result_09}. The model maintains stable predictions despite variations in illumination, background clutter, partial occlusions, and perspective distortions. This indicates that the multi-scale feature representations learned by ALBERT generalize well across diverse real-world scenarios.

\begin{figure*}[p]
\centering
\includegraphics[width=\textwidth]{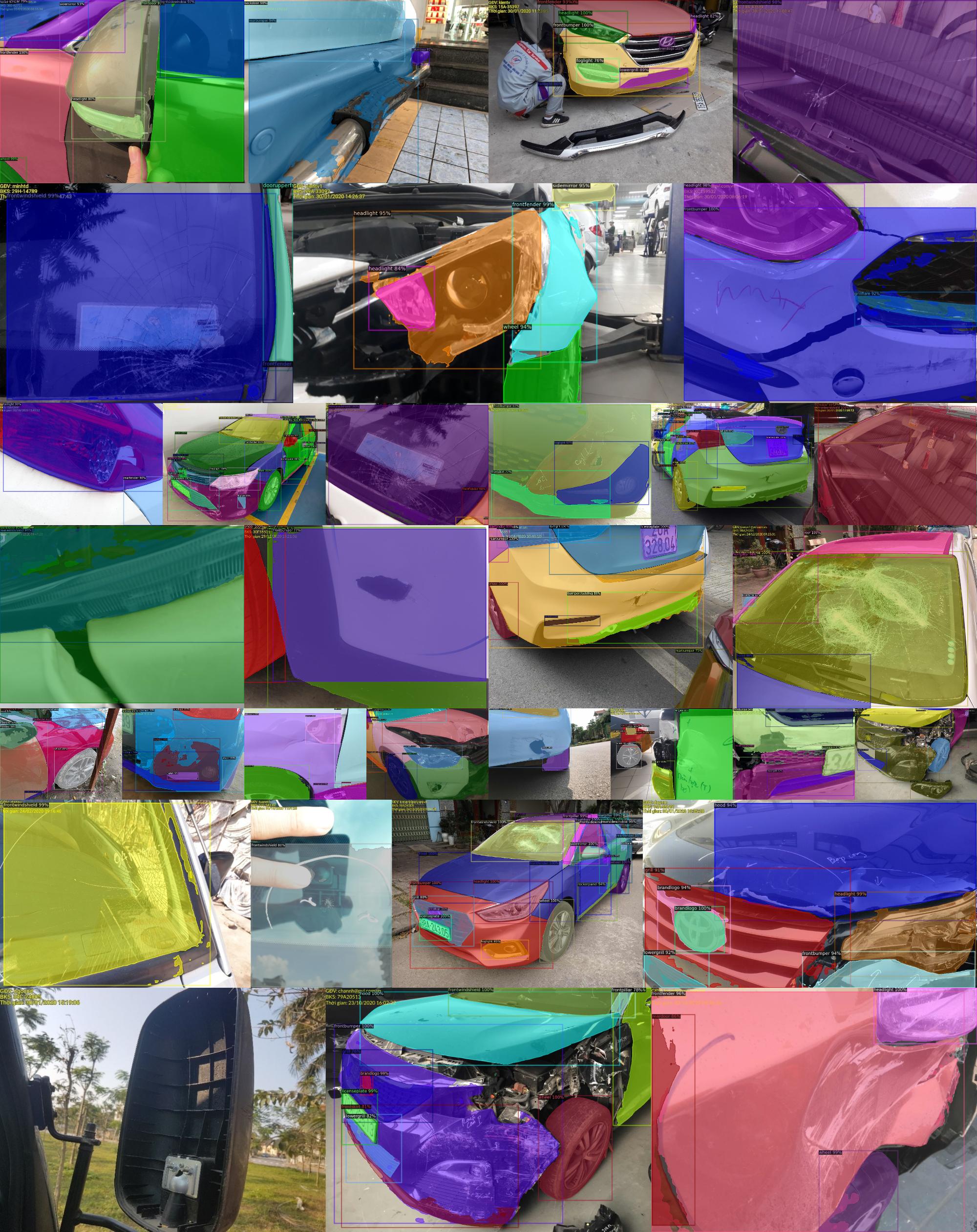}
\caption{Further qualitative examples illustrating ALBERT's strong multi-scale feature representation for vehicle component understanding.}
\label{fig:part_result_11}
\end{figure*}

\begin{figure*}[p]
\centering
\includegraphics[width=\textwidth]{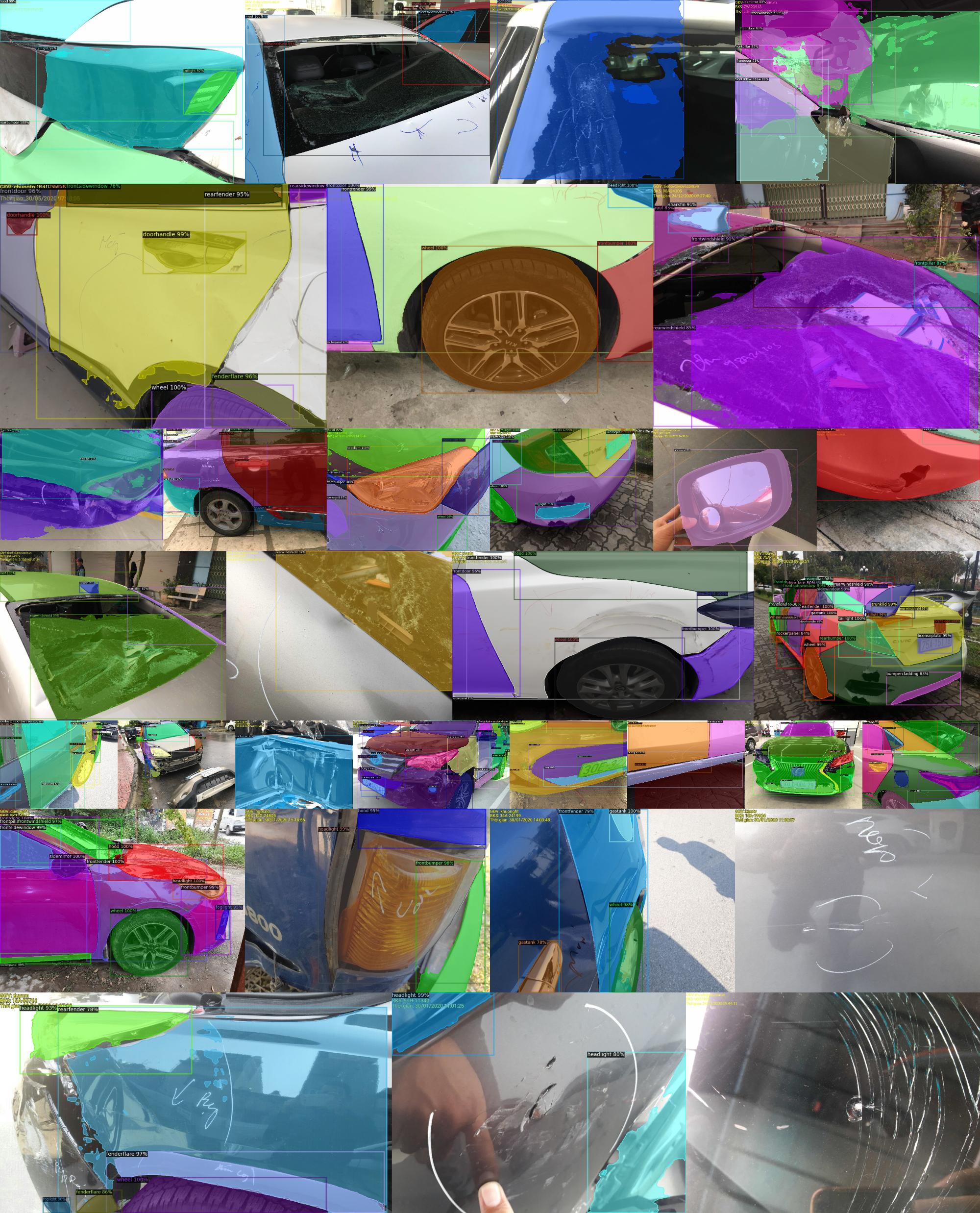}
\caption{ALBERT consistently identifies vehicle components across varying camera distances and perspective distortions.}
\label{fig:part_result_12}
\end{figure*}

\begin{figure*}[p]
\centering
\includegraphics[width=\textwidth]{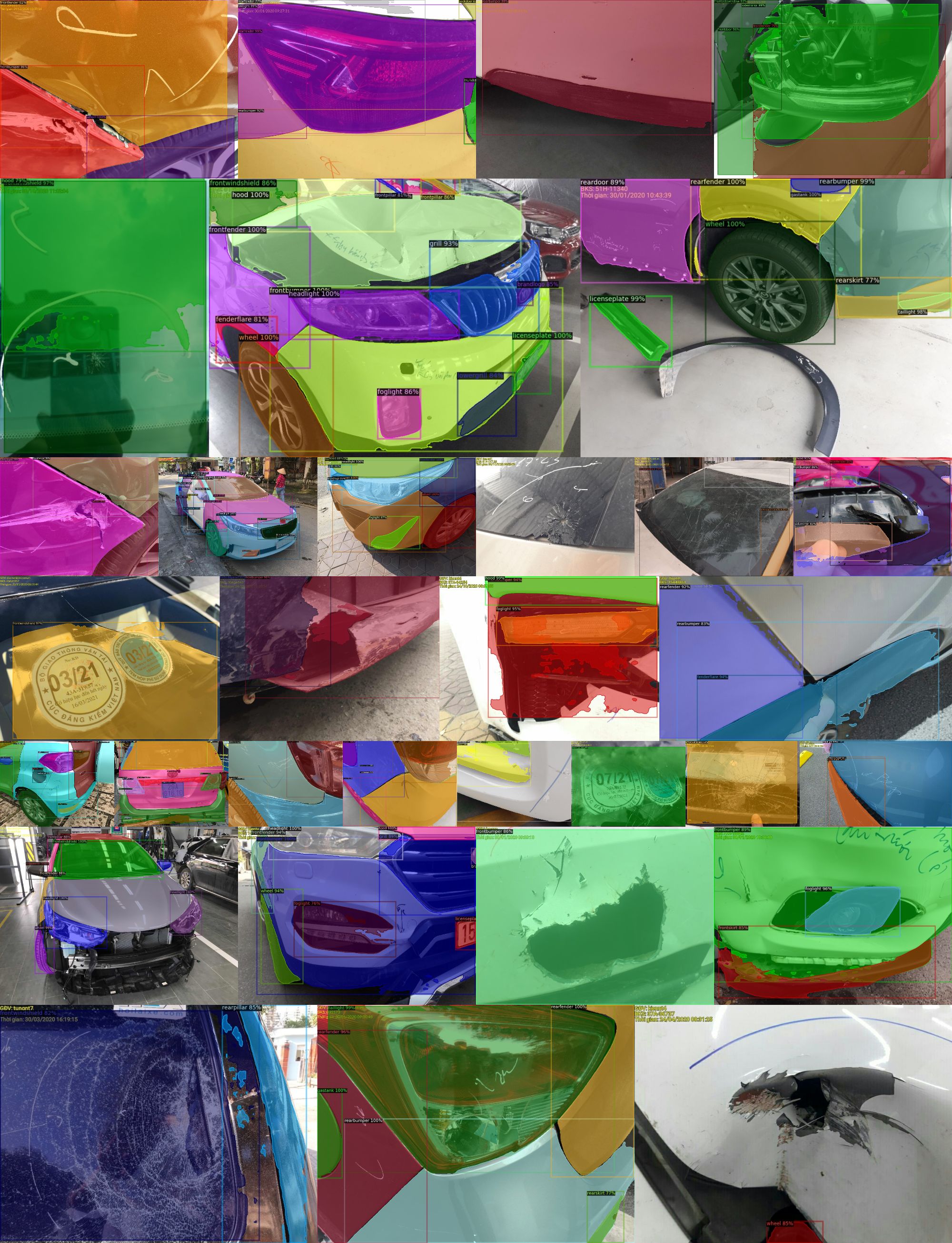}
\caption{Robust segmentation across multiple vehicle body structures including roof components, pillars, and side panels.}
\label{fig:part_result_13}
\end{figure*}

\begin{figure*}[p]
\centering
\includegraphics[width=\textwidth]{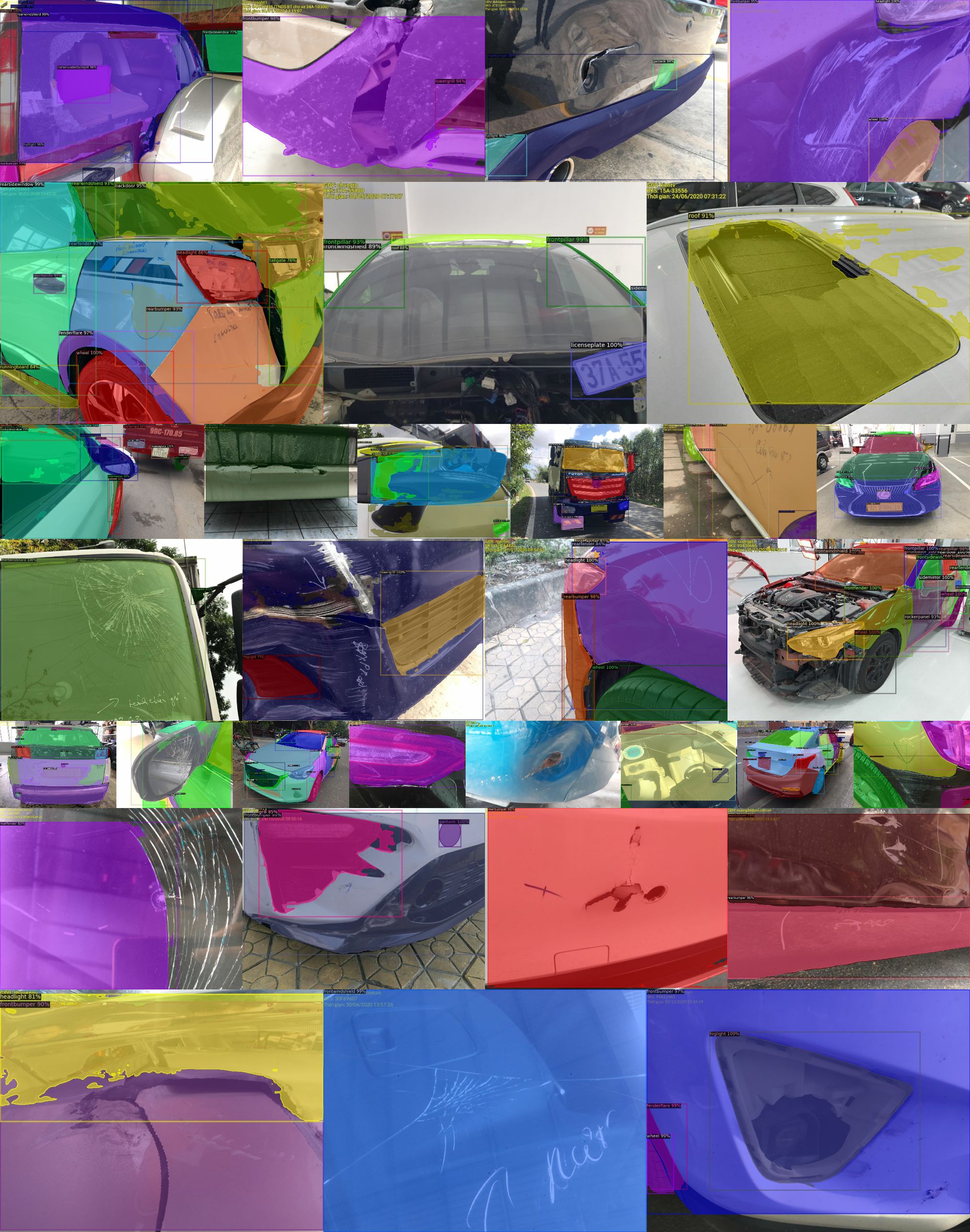}
\caption{Examples illustrating strong structural consistency in predicting complex component layouts across different vehicle designs.}
\label{fig:part_result_14}
\end{figure*}

\begin{figure*}[p]
\centering
\includegraphics[width=\textwidth]{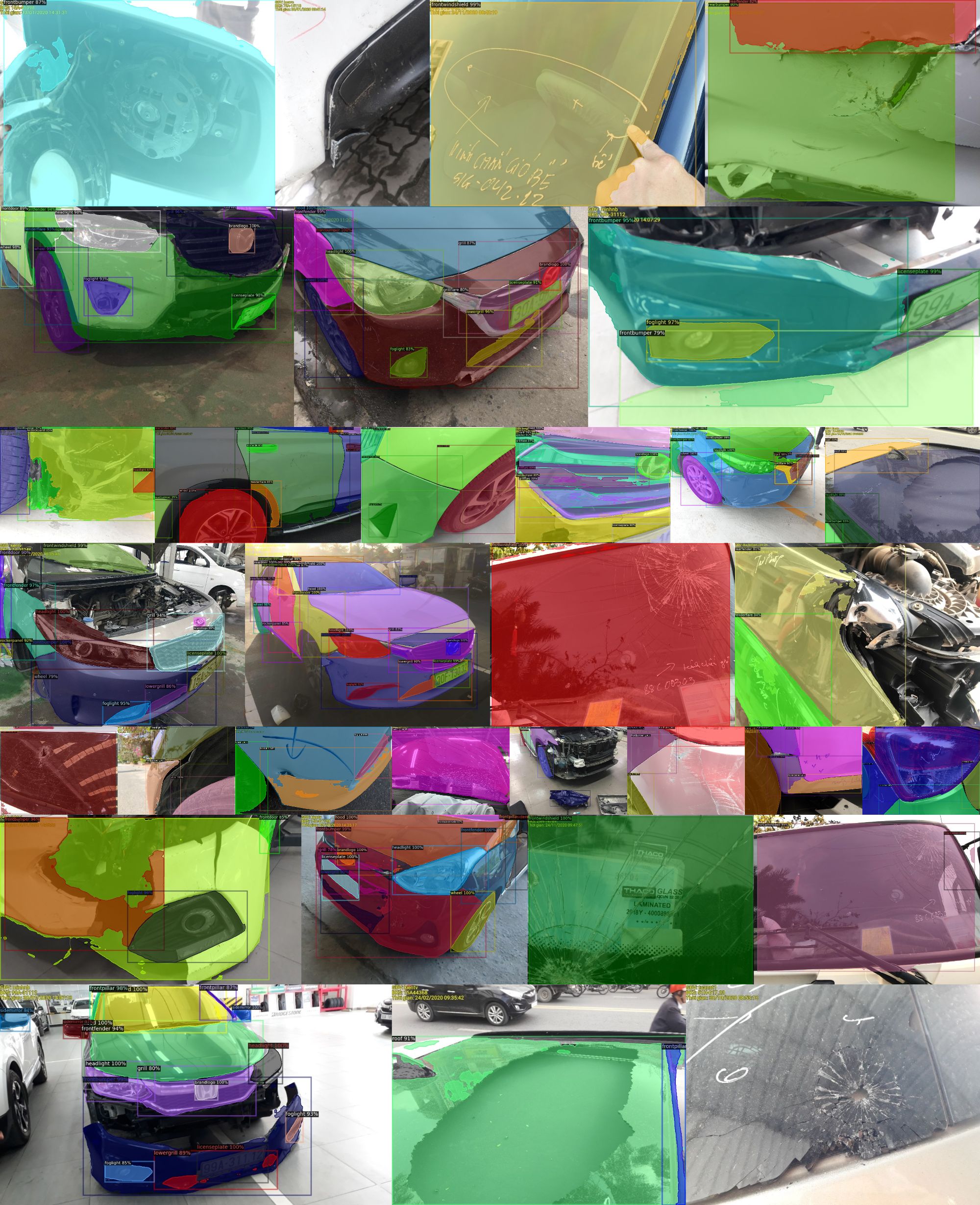}
\caption{ALBERT demonstrates stable segmentation performance even in challenging visual environments with cluttered backgrounds.}
\label{fig:part_result_15}
\end{figure*}

Additional qualitative results are presented in Figures~\ref{fig:part_result_11}--\ref{fig:part_result_15}. These examples demonstrate consistent segmentation performance across varying camera distances, vehicle designs, and structural layouts. The model effectively captures both large vehicle structures (e.g., doors, roofs, and bumpers) and smaller accessories such as door handles and logos, highlighting the scalability of the proposed approach for comprehensive vehicle component understanding.

\subsection{Qualitative Results of the ALBERT Damage Segmentation Model}

\begin{figure*}[p]
\centering
\includegraphics[width=\textwidth]{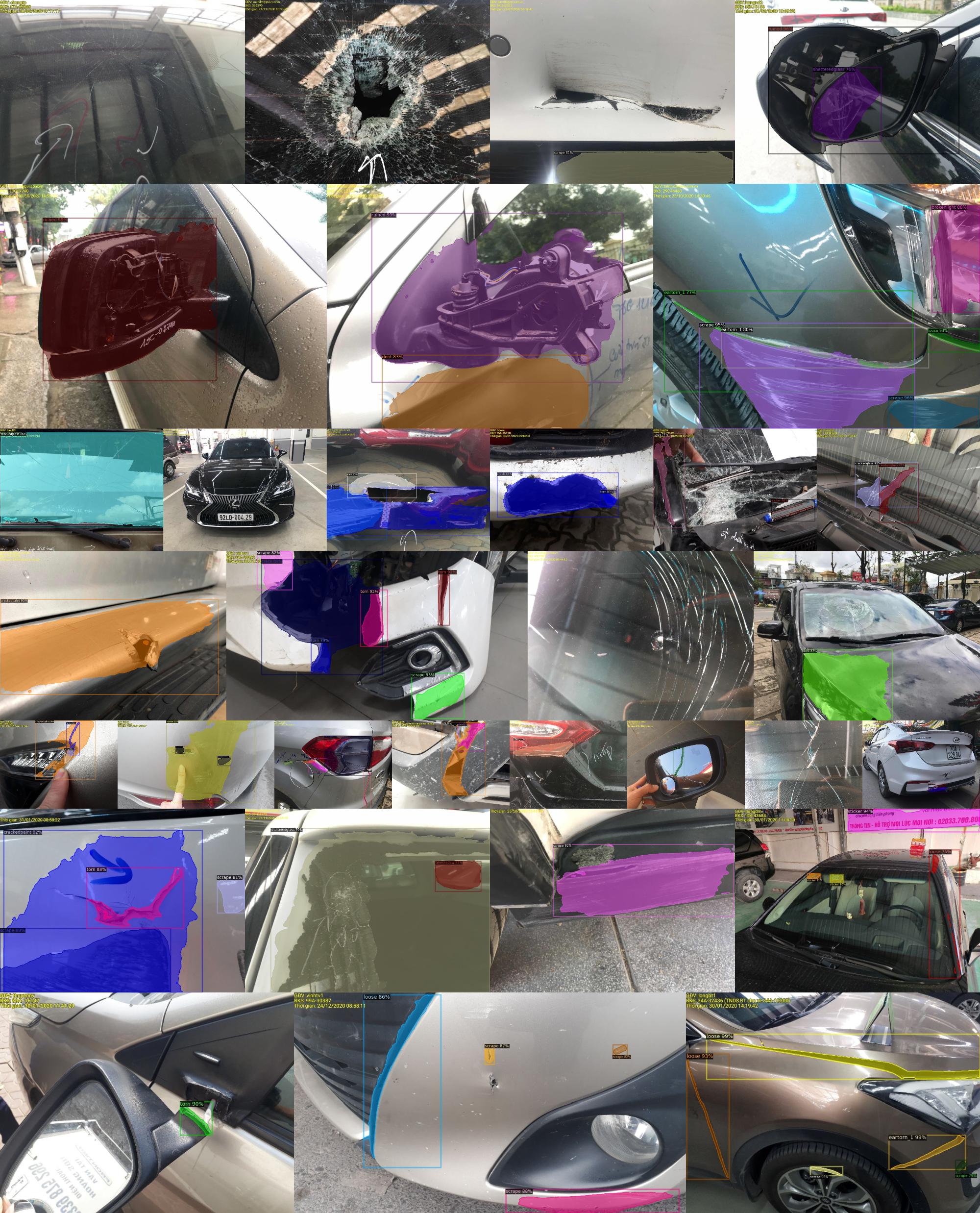}
\caption{Qualitative damage segmentation results produced by the ALBERT Damage Model on the MARSAIL dataset. The model accurately detects diverse damage patterns including dents, scratches, cracks, and shattered glass across multiple vehicle surfaces.}
\label{fig:damage_result_01}
\end{figure*}

\begin{figure*}[p]
\centering
\includegraphics[width=\textwidth]{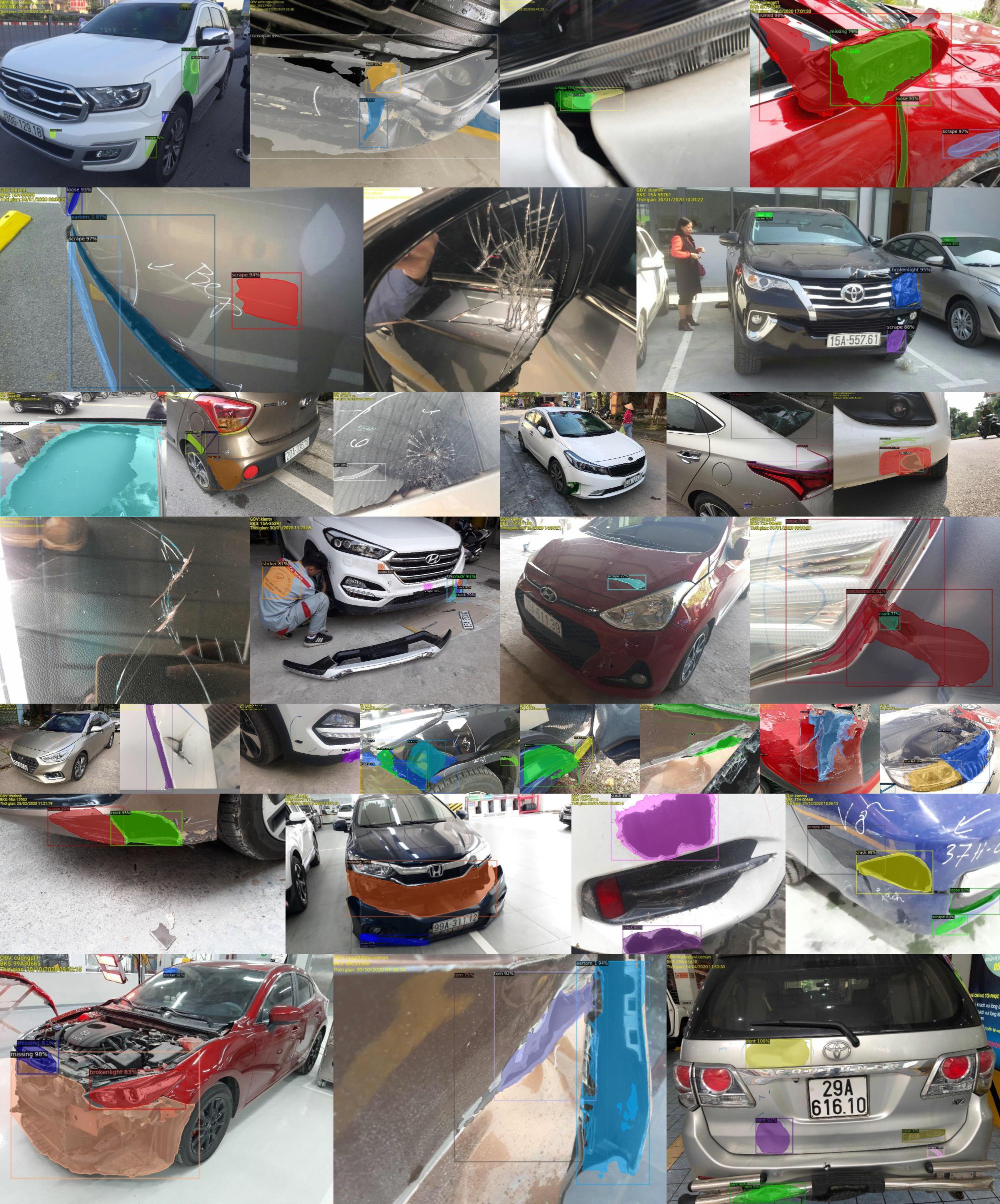}
\caption{Additional qualitative results demonstrating the robustness of ALBERT in detecting subtle surface damage across different vehicle colors, materials, and lighting conditions.}
\label{fig:damage_result_02}
\end{figure*}

\begin{figure*}[p]
\centering
\includegraphics[width=\textwidth]{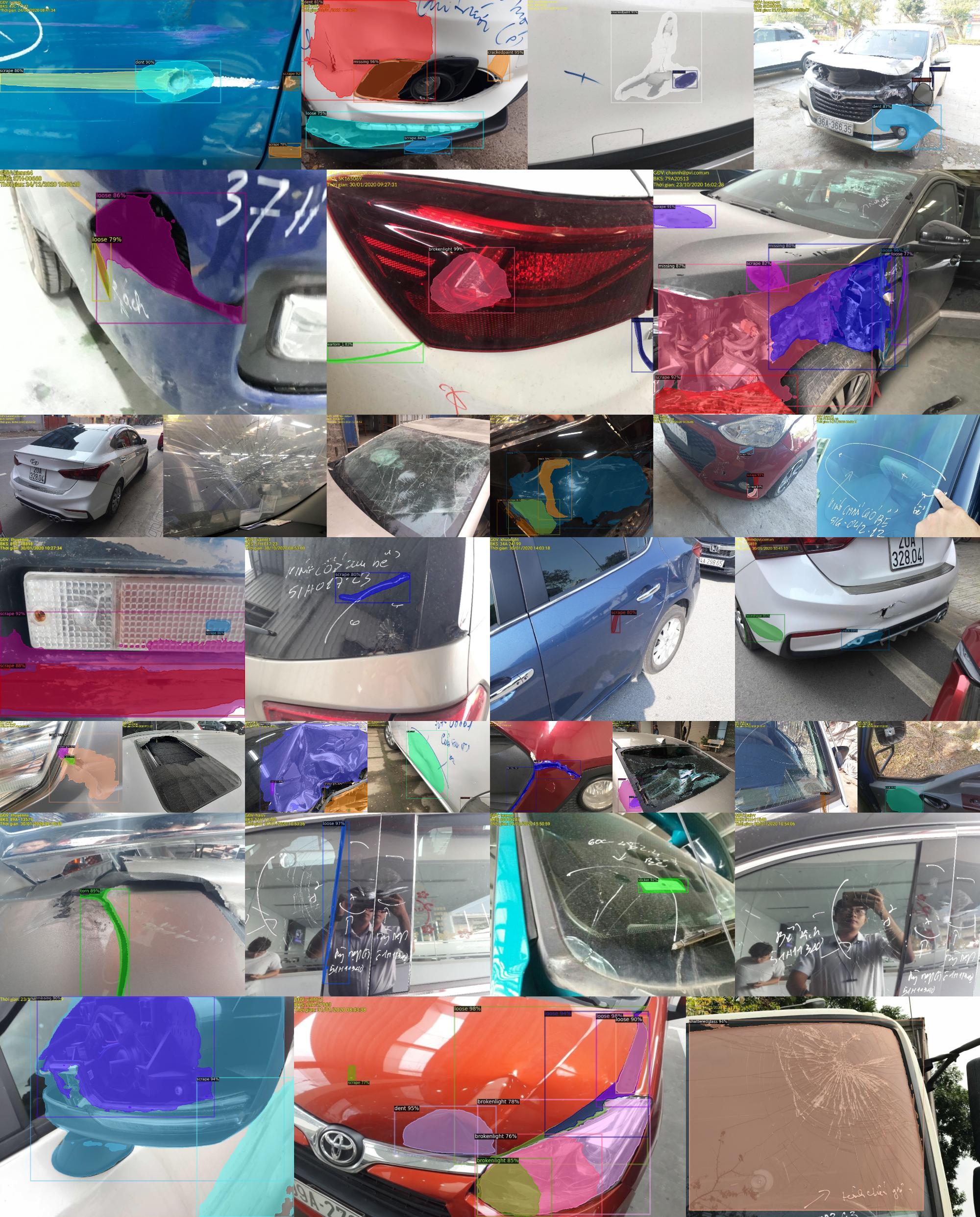}
\caption{Examples illustrating the capability of ALBERT to localize fine-grained damage structures such as hairline cracks and small dents with high boundary precision.}
\label{fig:damage_result_03}
\end{figure*}

In addition to vehicle part understanding, the ALBERT Damage Model demonstrates strong capability in detecting and segmenting various types of vehicle damage. Figures~\ref{fig:damage_result_01}--\ref{fig:damage_result_03} illustrate the model's ability to accurately identify common damage patterns including dents, scratches, cracks, and shattered glass across multiple vehicle surfaces. These examples highlight the effectiveness of the model in capturing both prominent structural damage and subtle surface-level defects.

\begin{figure*}[p]
\centering
\includegraphics[width=\textwidth]{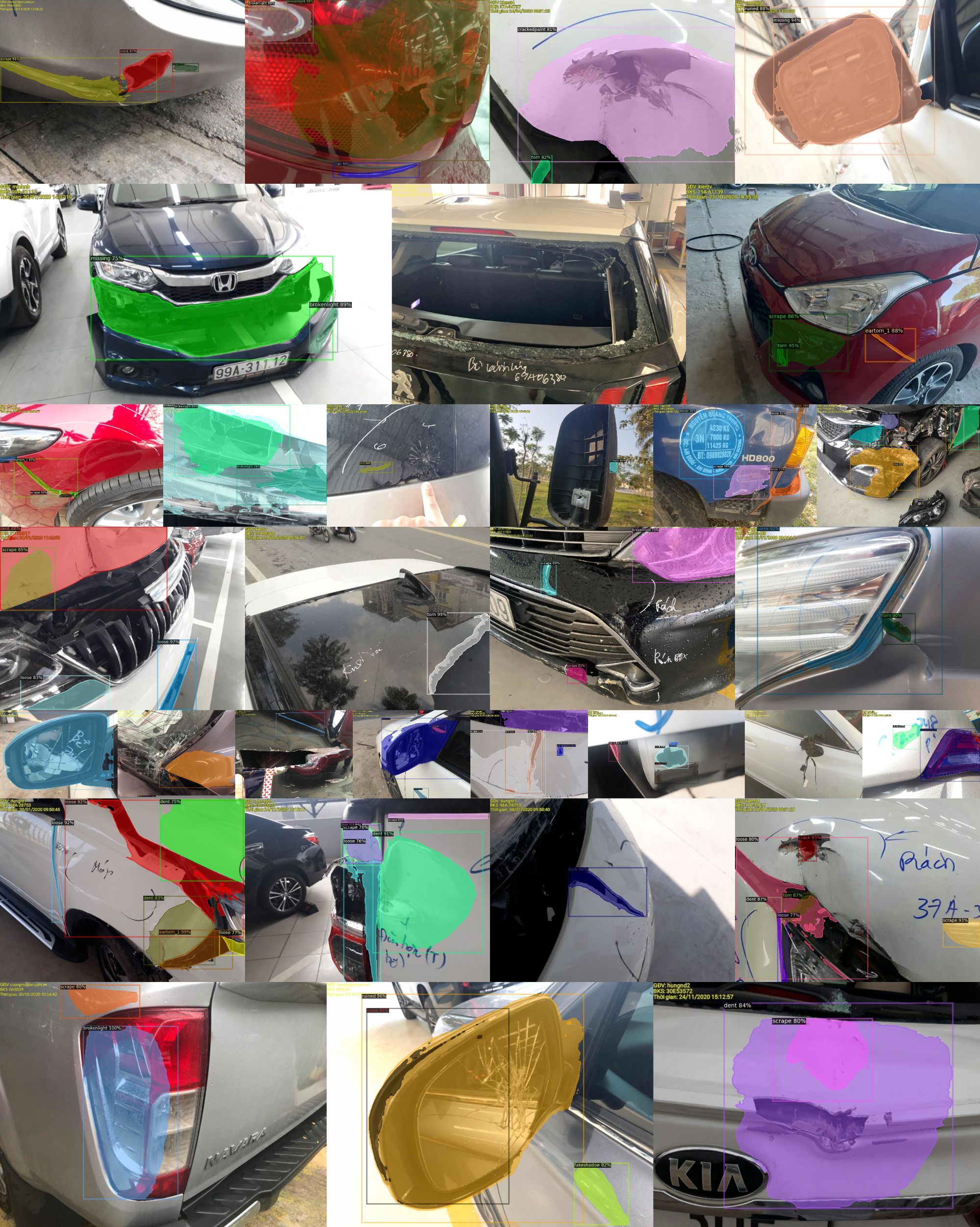}
\caption{ALBERT effectively identifies multiple co-occurring damage categories within a single vehicle image, supporting reliable multi-instance damage assessment.}
\label{fig:damage_result_04}
\end{figure*}

\begin{figure*}[p]
\centering
\includegraphics[width=\textwidth]{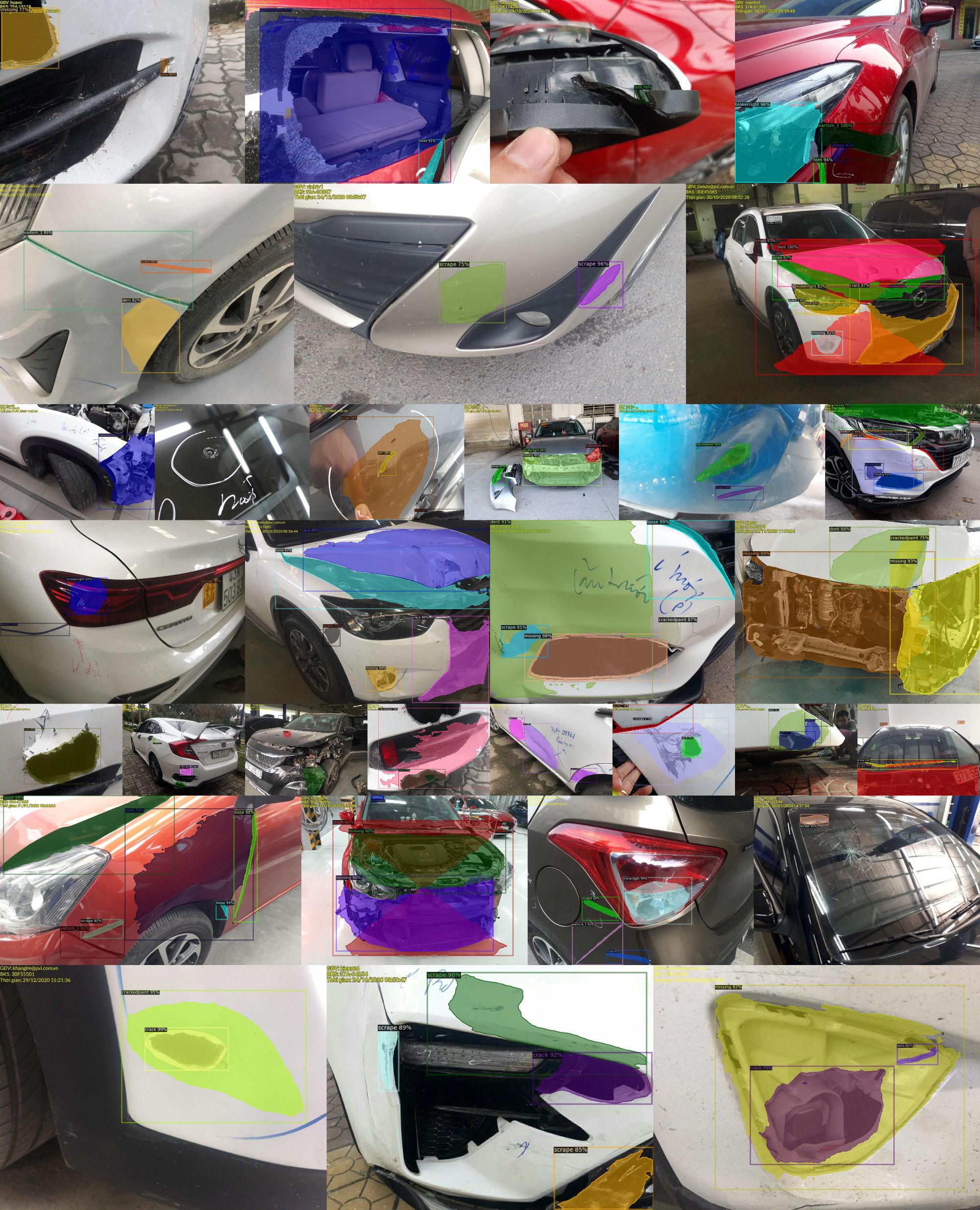}
\caption{Qualitative examples showing strong detection of structural deformation such as crushed panels and severely damaged surfaces.}
\label{fig:damage_result_05}
\end{figure*}

\begin{figure*}[p]
\centering
\includegraphics[width=\textwidth]{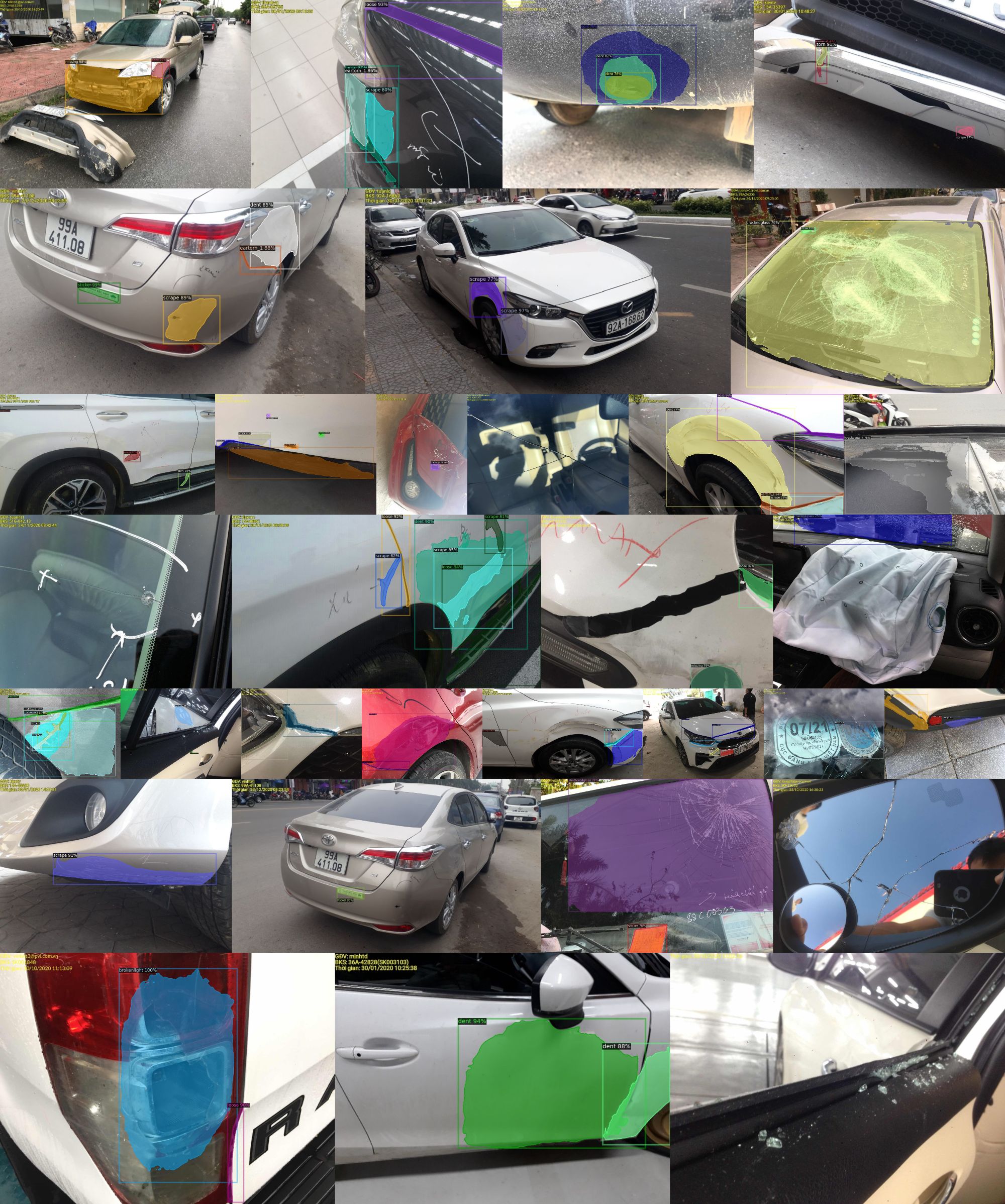}
\caption{The model successfully detects damage across a wide range of vehicle viewpoints, demonstrating strong generalization capability.}
\label{fig:damage_result_06}
\end{figure*}

Figures~\ref{fig:damage_result_04}, \ref{fig:damage_result_05}, and \ref{fig:damage_result_06} further demonstrate the model's capability to handle complex damage scenarios involving multiple co-occurring damage regions and severe structural deformation. 

% The model successfully detects crushed panels and distorted surfaces while maintaining accurate segmentation boundaries.

\begin{figure*}[p]
\centering
\includegraphics[width=\textwidth]{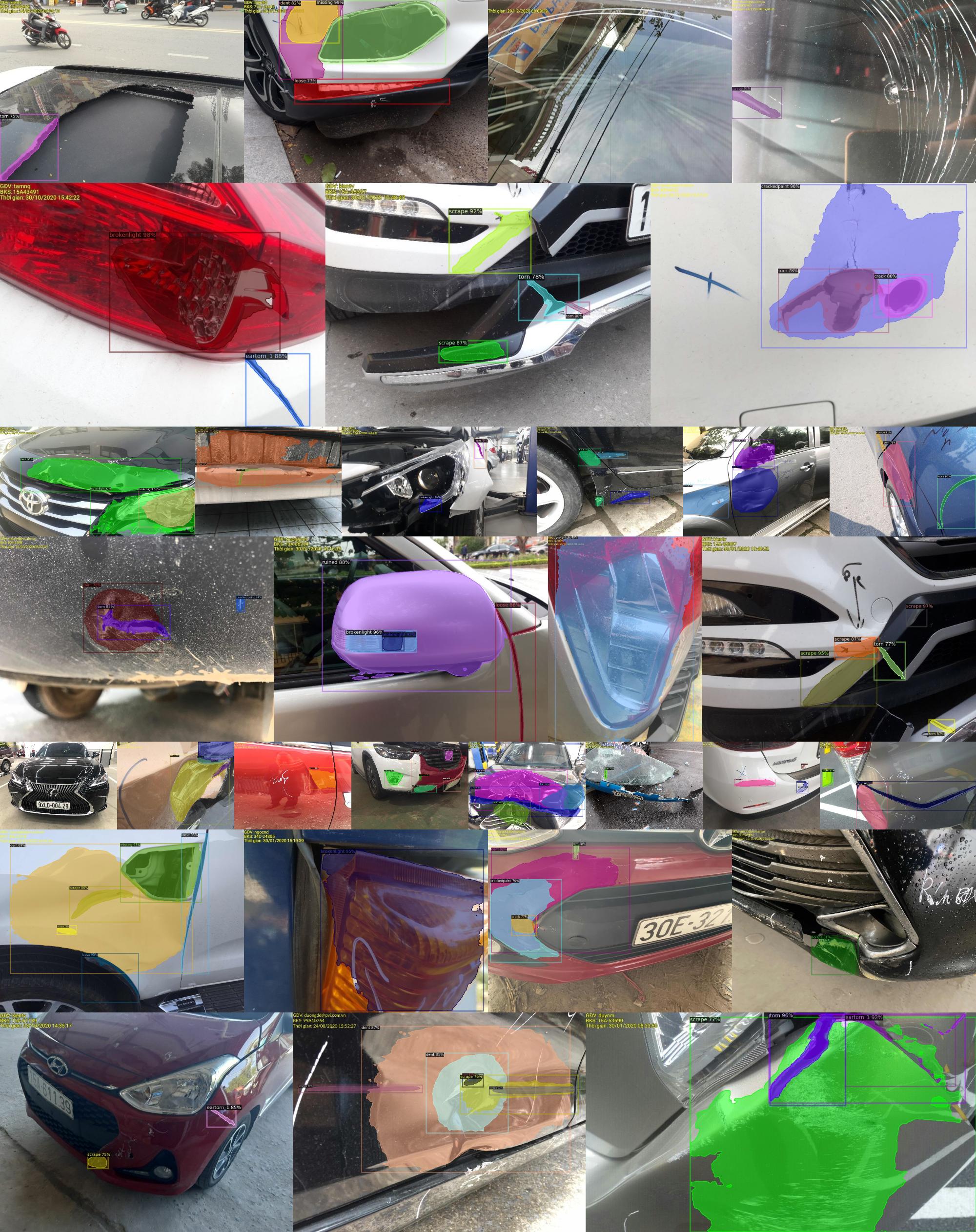}
\caption{Examples illustrating reliable segmentation of glass-related damage such as cracked and shattered windshields.}
\label{fig:damage_result_07}
\end{figure*}

\begin{figure*}[p]
\centering
\includegraphics[width=\textwidth]{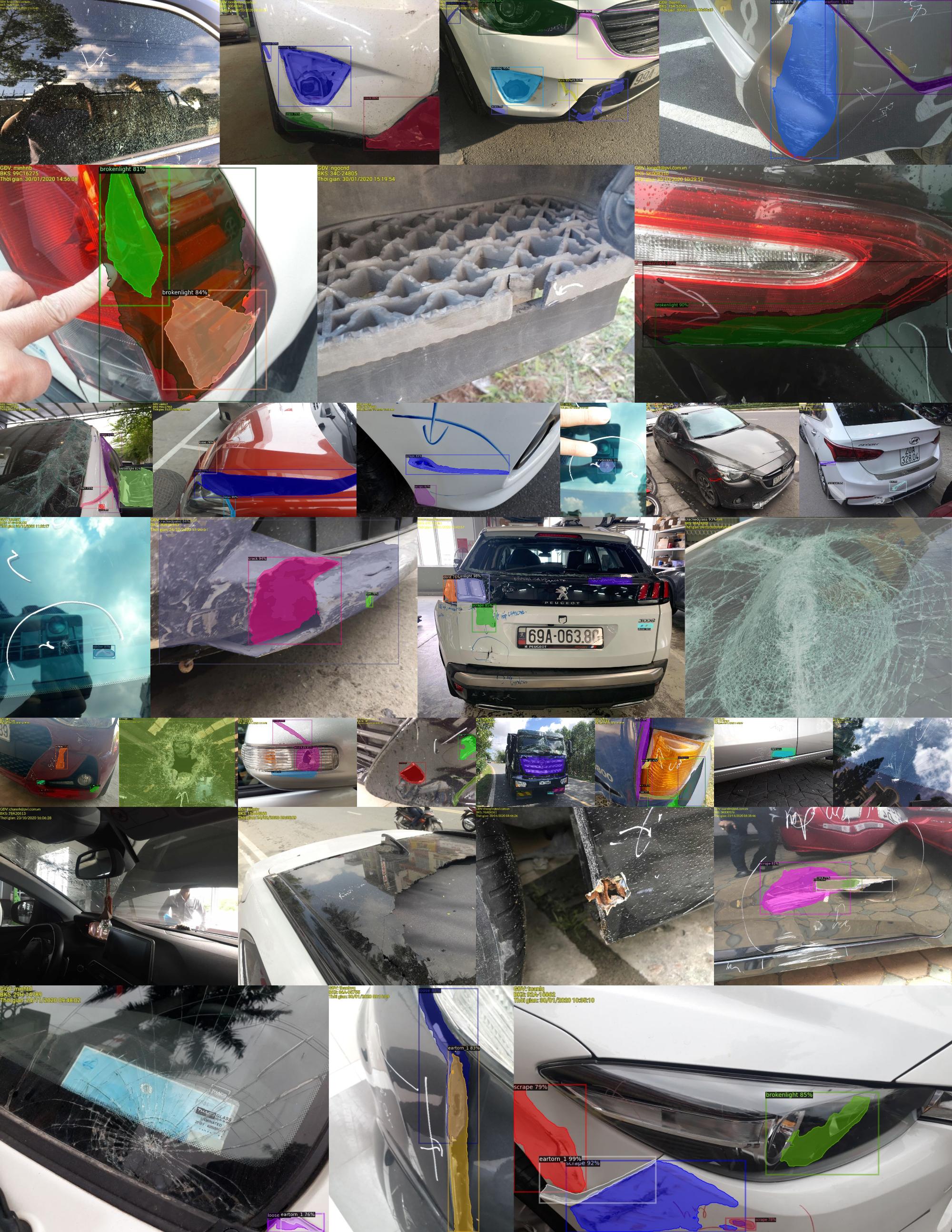}
\caption{ALBERT captures subtle surface defects including scratches and chipped paint, which are traditionally difficult to detect automatically.}
\label{fig:damage_result_08}
\end{figure*}

\begin{figure*}[p]
\centering
\includegraphics[width=\textwidth]{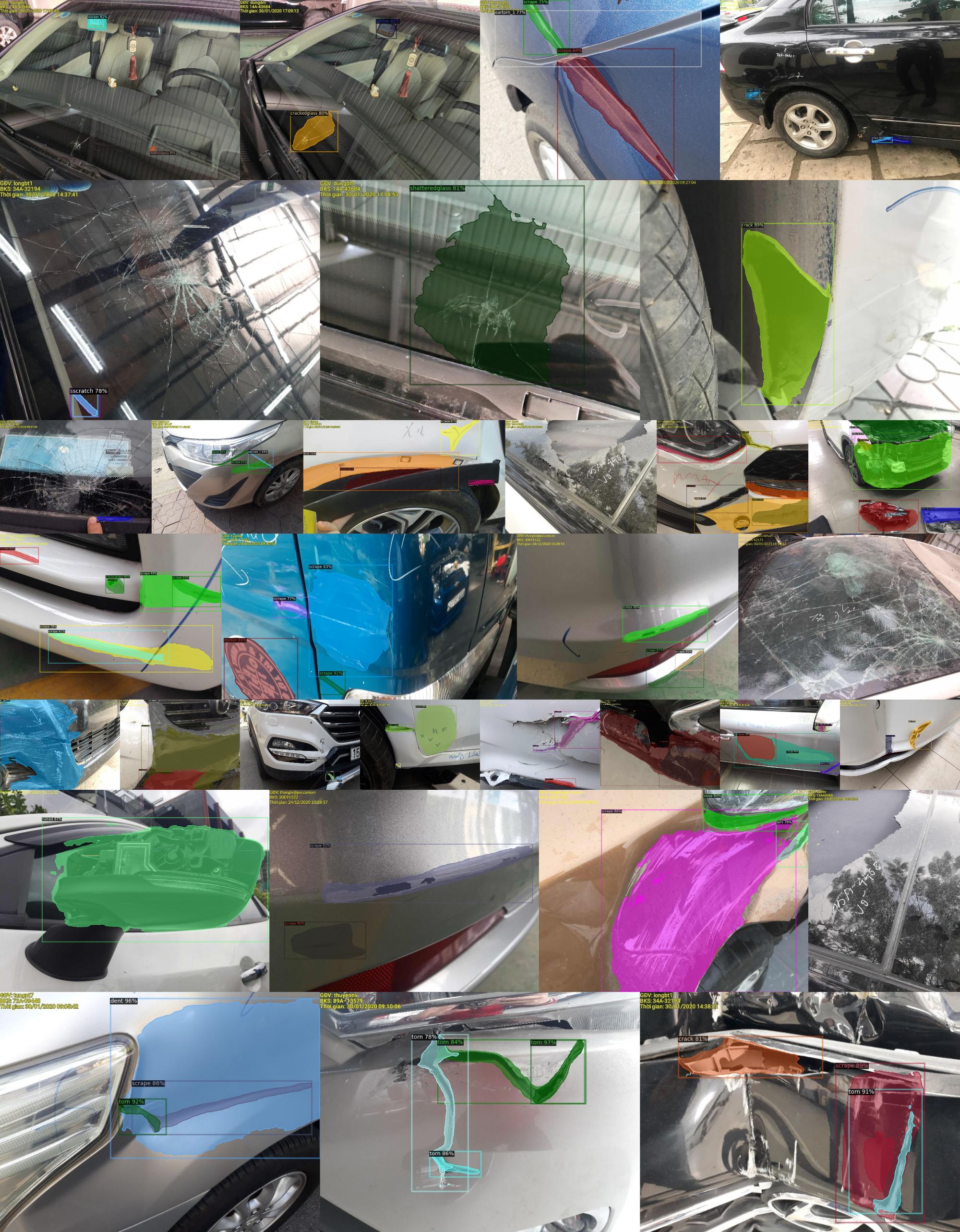}
\caption{The model maintains strong performance even when damage appears under challenging environmental conditions such as reflections or shadows.}
\label{fig:damage_result_10}
\end{figure*}

The robustness of the model under challenging visual conditions is shown in Figures~\ref{fig:damage_result_07}, \ref{fig:damage_result_08}, and \ref{fig:damage_result_10}. 

% In these cases, the system reliably segments glass-related damage such as cracked windshields as well as subtle surface defects including chipped paint and fine scratches, even in the presence of reflections or shadows.

\begin{figure*}[p]
\centering
\includegraphics[width=\textwidth]{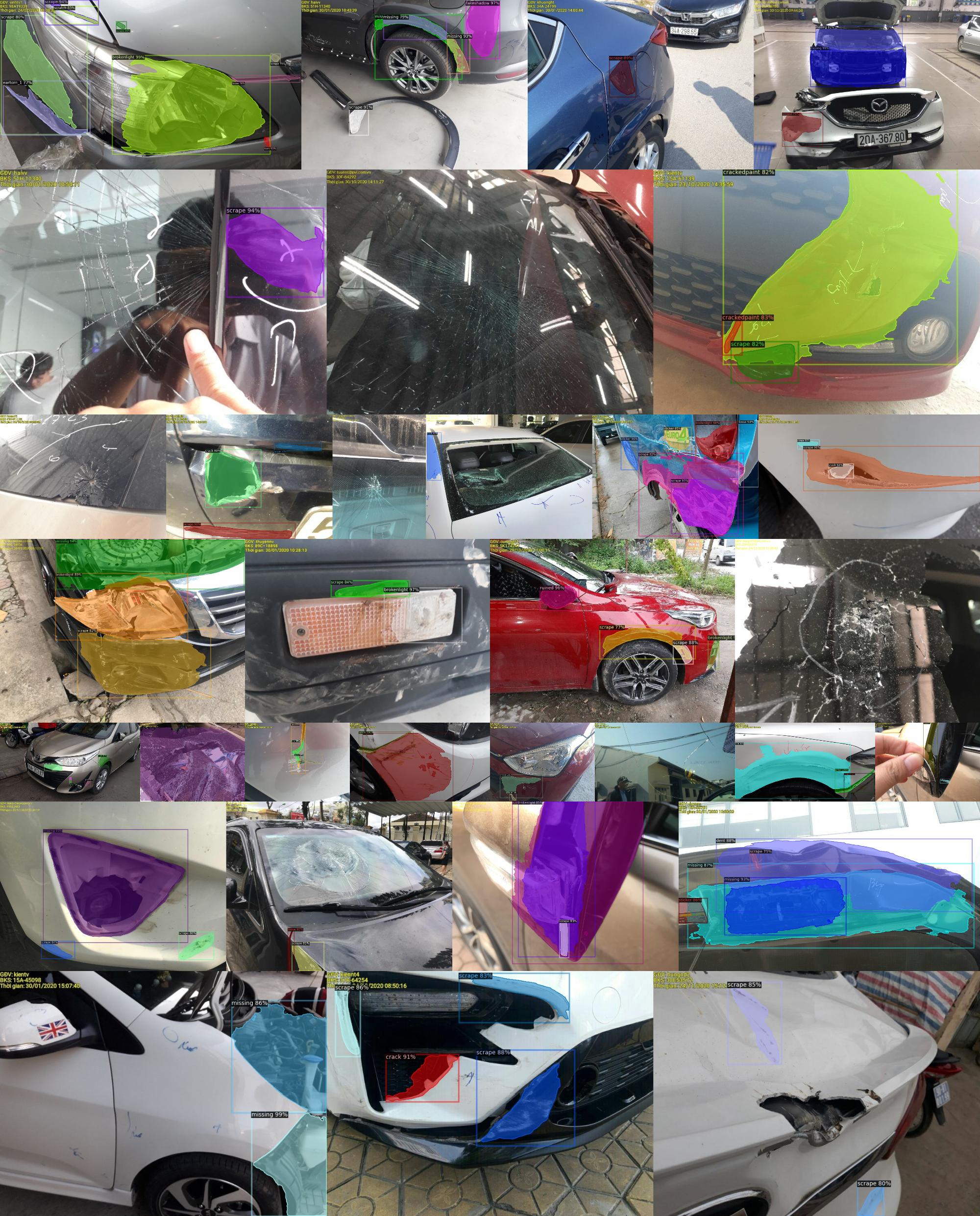}
\caption{Additional examples showing ALBERT's ability to capture both small cosmetic damage and large structural defects across multiple vehicle panels.}
\label{fig:damage_result_11}
\end{figure*}

\begin{figure*}[p]
\centering
\includegraphics[width=\textwidth]{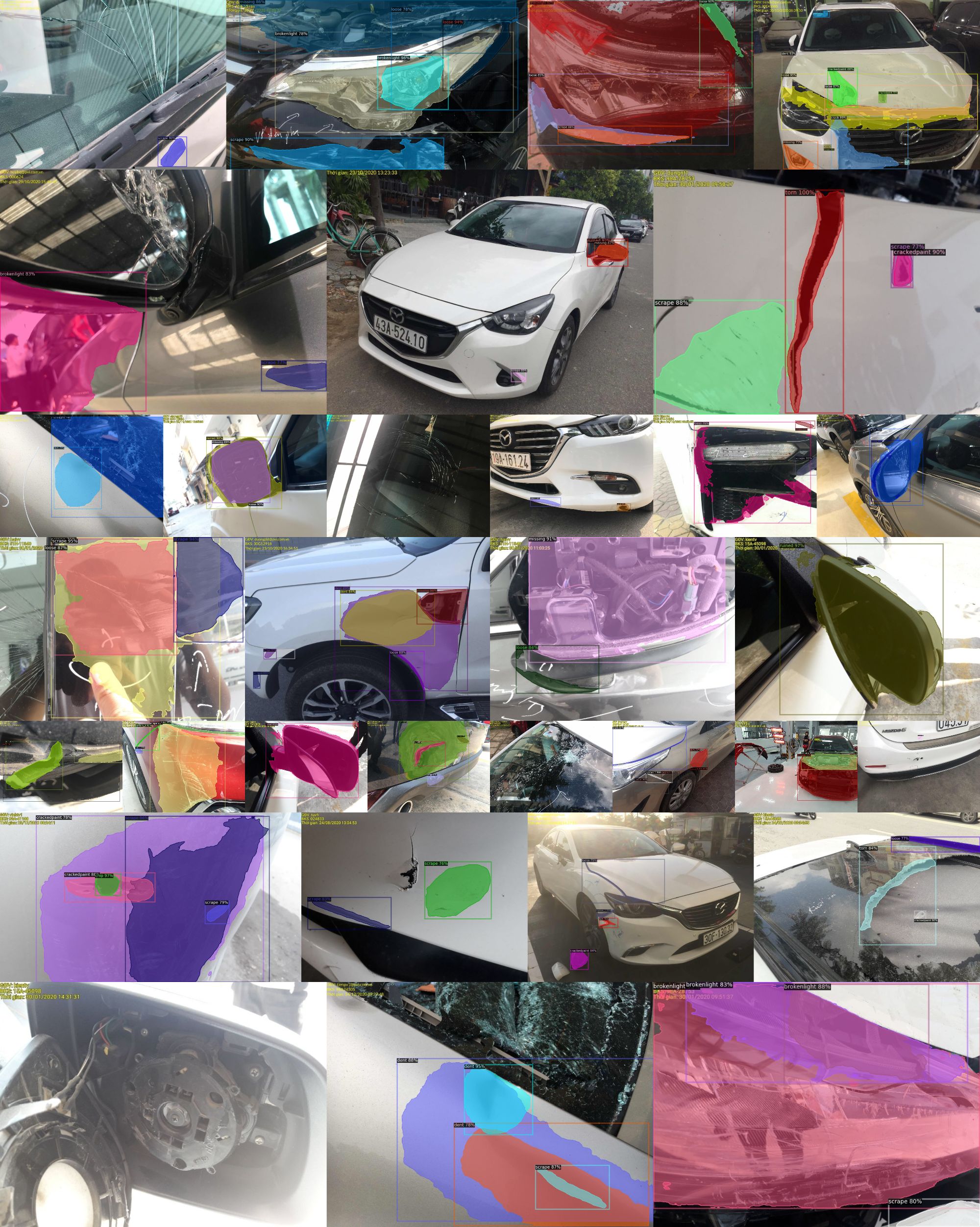}
\caption{Robust qualitative results highlighting the scalability of ALBERT across diverse vehicle models and surface materials.}
\label{fig:damage_result_12}
\end{figure*}

\begin{figure*}[p]
\centering
\includegraphics[width=\textwidth]{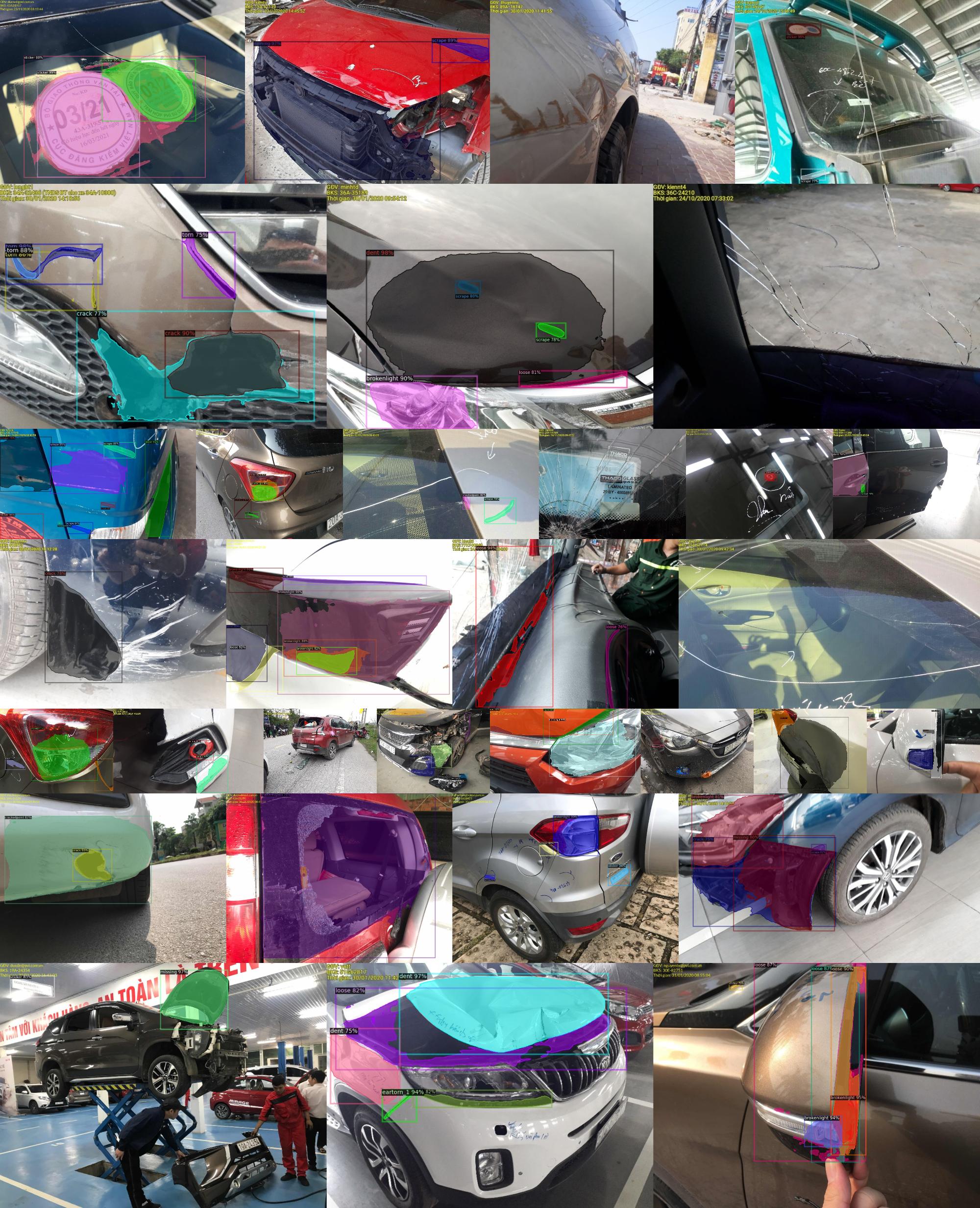}
\caption{Examples illustrating the model's capability to maintain high segmentation quality for overlapping and adjacent damage regions.}
\label{fig:damage_result_13}
\end{figure*}

\begin{figure*}[p]
\centering
\includegraphics[width=\textwidth]{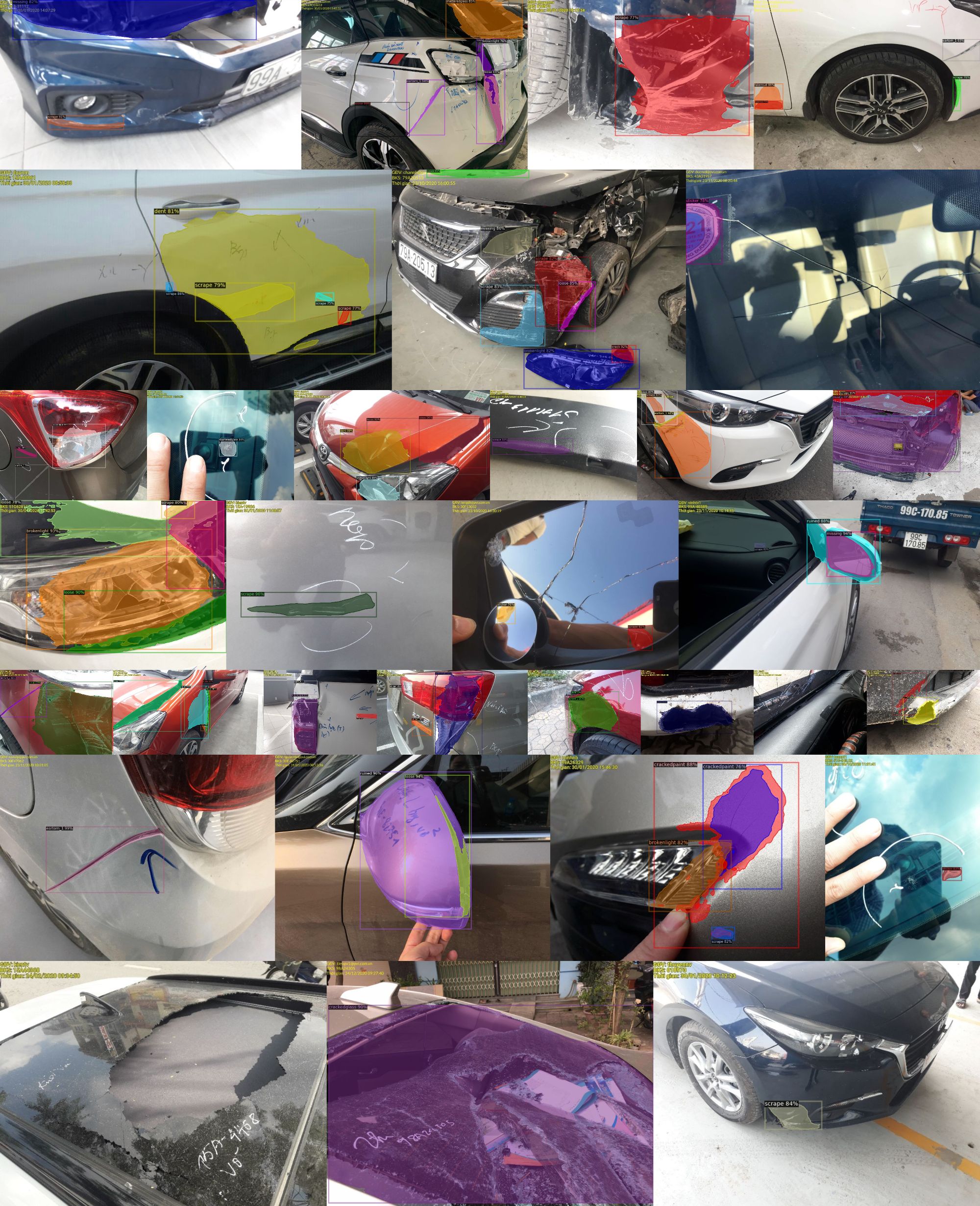}
\caption{ALBERT accurately differentiates between genuine structural damage and visually misleading artifacts that could otherwise lead to incorrect insurance assessments.}
\label{fig:damage_result_14}
\end{figure*}

\begin{figure*}[p]
\centering
\includegraphics[width=\textwidth]{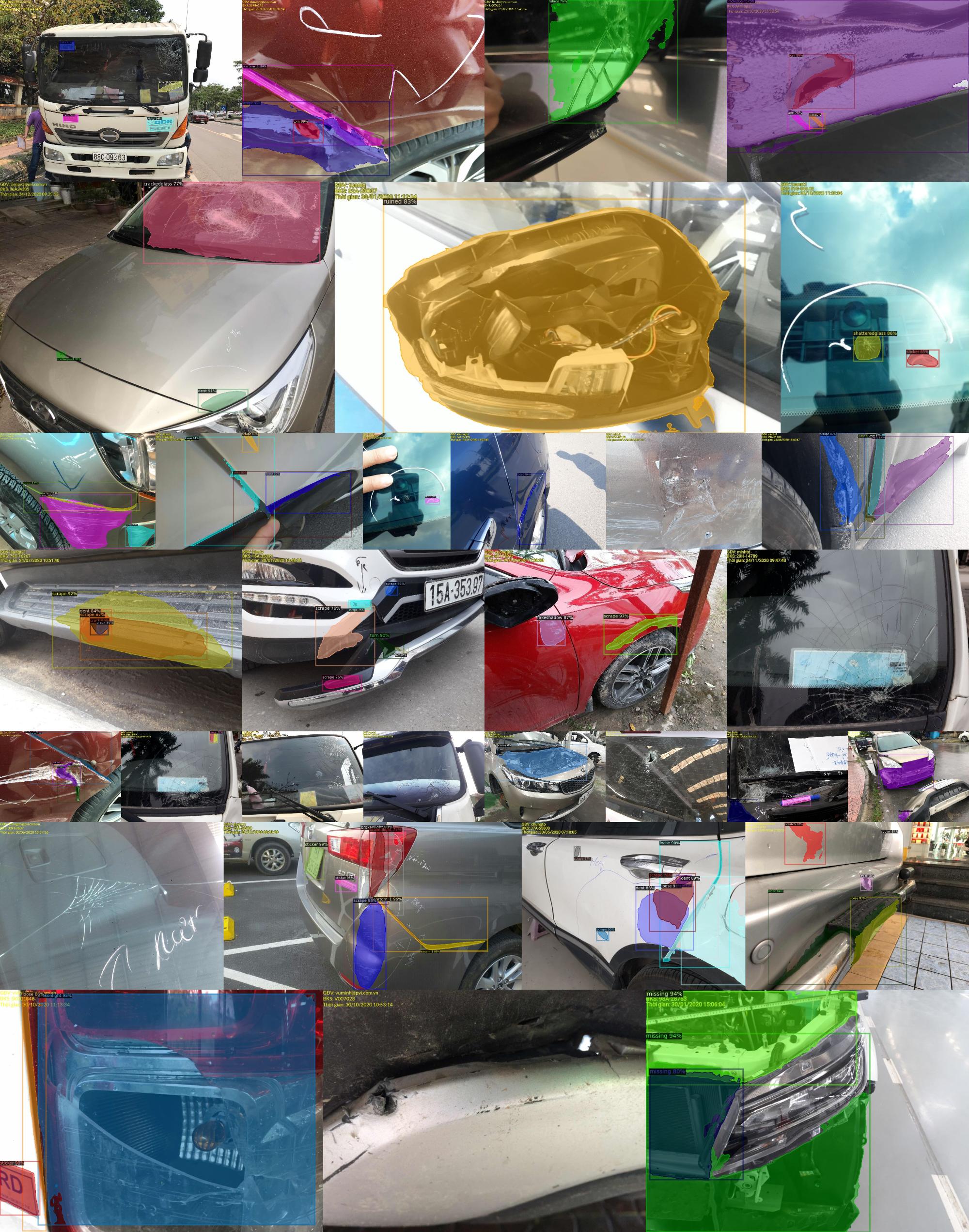}
\caption{The model consistently captures complex deformation patterns across multiple vehicle body panels.}
\label{fig:damage_result_15}
\end{figure*}

Finally, Figures~\ref{fig:damage_result_11}--\ref{fig:damage_result_15} provide additional examples demonstrating consistent damage localization across complex vehicle geometries and diverse operational environments. The model maintains high segmentation quality for overlapping damage regions and effectively distinguishes genuine structural damage from visually misleading artifacts.

Overall, the qualitative results confirm that the proposed ALBERT framework provides reliable and accurate segmentation for both vehicle structural components and damage regions. This capability is critical for enabling automated vehicle inspection systems in real-world automotive insurance and maintenance applications.

\begin{figure*}[p]
\centering
\includegraphics[width=\textwidth]{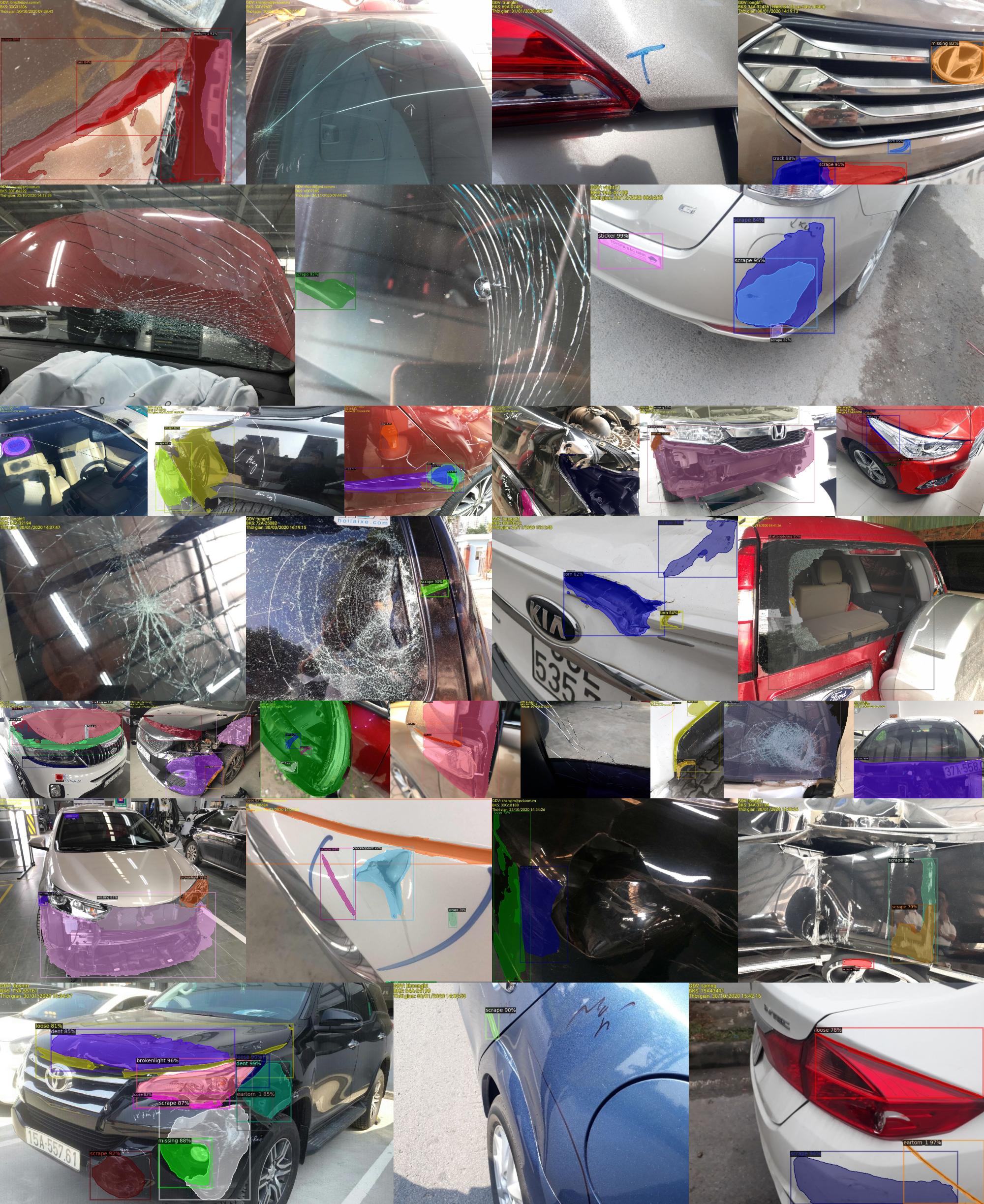}
\caption{Qualitative examples demonstrating consistent mask localization for complex and irregular damage patterns.}
\label{fig:damage_result_09}
\end{figure*}

\chapter{MARSAIL NLP: DOTA Document Intelligence Engine}

\section{Introduction}

At \textbf{MARSAIL (Motor AI Recognition Solution Artificial Intelligence Laboratory)}, 
we extend computer vision beyond visual recognition into full-scale 
\textbf{Document Intelligence}. 

To power this vision, we introduce:

\begin{tcolorbox}[
    colback=blue!4!white,
    colframe=blue!80!black,
    boxrule=1.2pt,
    arc=3mm
]
\centering
{\Huge\bfseries DOTA}\\[6pt]
{\Large Deep Optimization and Transformer-based AI}\\[6pt]
{\normalsize The Core NLP Engine of the MARS Ecosystem}
\end{tcolorbox}

DOTA is a next-generation OCR and sequence modeling framework 
designed specifically for vehicle insurance document processing. 
Unlike generic OCR engines, DOTA is domain-optimized for:

\begin{itemize}
    \item Thai National ID Card extraction
    \item Thai Driving License recognition
    \item Vehicle Mileage detection
    \item VIN (Vehicle Identification Number) parsing
    \item License Plate recognition
\end{itemize}

DOTA serves as the \textbf{textual backbone} of all MARS users.

\section{Scientific Recognition}

DOTA has been accepted for publication at the 
\textbf{17th International Conference on Knowledge and Smart Technology (KST 2025)}, 
indexed by IEEE Xplore, Scopus, and DBLP.

This milestone validates DOTA as a research-grade and production-ready innovation.

\section{Why Traditional OCR Fails}

Conventional OCR engines such as PaddleOCR, EasyOCR, and Tesseract are designed for general-purpose text extraction. 
However, vehicle insurance environments introduce:

\begin{enumerate}
    \item Severe long-tail character imbalance (Thai language + numeric mixtures)
    \item Structured yet noisy layouts
    \item Real-world alignment distortion
    \item Mixed alphanumeric VIN sequences
\end{enumerate}

Standard CTC-based models minimize:

\[
\mathcal{L}_{CTC} = - \log P(Y|X)
\]

However, this implicitly biases toward frequent characters.

\section{DOTA: Mathematical Foundation}

DOTA introduces a \textbf{Class-Balanced Focal CTC Loss (CB-FCTC)}:

\[
\mathcal{L}_{DOTA} = (1 - p_t)^{\gamma} \cdot \mathcal{L}_{CTC}
\]

where:

\begin{itemize}
    \item $p_t$ approximates prediction confidence
    \item $\gamma$ controls hard-example emphasis
\end{itemize}

This improves gradient signal for rare characters in Thai OCR.

\subsection{Optimization Objective}

Given input image $X$ and target sequence $Y$:

\[
P(Y|X) = \sum_{\pi \in \mathcal{B}^{-1}(Y)} 
\prod_{t=1}^{T} P(\pi_t | X)
\]

DOTA modifies gradient scaling dynamically:

\[
\nabla \mathcal{L}_{DOTA}
= (1 - p_t)^{\gamma} \nabla \mathcal{L}_{CTC}
\]

This enables superior learning for:

\begin{itemize}
    \item Rare Thai characters
    \item Low-frequency VIN patterns
    \item Edge-case numeric distortions
\end{itemize}

\section{Architecture Overview}

DOTA integrates five key components:

\subsection{1. Deformable Convolution Backbone}

Instead of rigid CNN sampling:

\[
y(p_0) = \sum_{k} w_k \cdot x(p_0 + p_k)
\]

DOTA applies deformable offsets:

\[
y(p_0) = \sum_{k} w_k \cdot x(p_0 + p_k + \Delta p_k)
\]

This allows spatial flexibility for misaligned documents.

\subsection{2. Patch Embedding + Transformer Encoder}

Input feature map is converted into patch tokens:

\[
z_i = W_e \cdot \text{flatten}(x_i)
\]

Then processed via multi-head self-attention:

\[
\text{Attention}(Q,K,V)
= \text{softmax}\left(\frac{QK^T}{\sqrt{d}}\right)V
\]

Capturing long-range dependencies in VIN and ID sequences.

\subsection{3. Bidirectional GRU Sequence Refinement}

Sequence modeling improves character continuity.

\subsection{4. Adaptive Dropout}

Dynamic regularization:

\[
p = p_{min} + \sigma(f(x)) (p_{max} - p_{min})
\]

Improves robustness in noisy scans.

\subsection{5. Imbalance-Aware CTC Loss}

Production-safe and beam-search compatible.

\section{Pseudo-Code Overview}

\begin{tcolorbox}[
    colback=blue!4!white,
    colframe=blue!80!black,
    boxrule=1.2pt,
    arc=3mm,
    width=\linewidth
]
\small
\textbf{DOTA End-to-End Optimization Pipeline}

\medskip

Given an input document image $I \in \mathbb{R}^{H \times W \times 3}$:

\begin{align}
\mathbf{F} &= \mathcal{D}_{\text{ResNet}}(I) 
&& \text{(Deformable Feature Extraction)} \\[4pt]
\mathbf{Z}_0 &= \mathcal{P}(\mathbf{F}) 
&& \text{(Patch Embedding Projection)} \\[4pt]
\mathbf{Z}_1 &= \text{BiGRU}(\mathbf{Z}_0) 
&& \text{(Bidirectional Sequence Modeling)} \\[4pt]
\mathbf{Z}_2 &= \mathbf{Z}_1 + \mathbf{PE} 
&& \text{(Positional Encoding Injection)} \\[4pt]
\mathbf{Z}_3 &= \mathcal{T}(\mathbf{Z}_2) 
&& \text{(Multi-Head Self-Attention)} \\[4pt]
\mathbf{O} &= \mathbf{W}_o \mathbf{Z}_3 + b 
&& \text{(Character Logit Projection)}
\end{align}

\medskip

\textbf{Training Objective:}

\begin{equation}
\mathcal{L}_{\text{DOTA}}
=
(1 - p_t)^{\gamma}
\cdot
\mathcal{L}_{\text{CTC}}(\mathbf{O}, Y)
\end{equation}

where

\[
p_t = \frac{1}{T}
\sum_{t=1}^{T}
\max_{c} \; \text{Softmax}(\mathbf{O}_{t,c})
\]

controls adaptive gradient scaling for rare-character emphasis.

\medskip

\textbf{Inference:}

\begin{equation}
\hat{Y}
=
\arg\max_{Y}
P(Y|I)
\quad
\text{via Beam Search Decoding}
\end{equation}

\medskip

\centering
\textit{End-to-End Differentiable Document Intelligence Optimization}
\end{tcolorbox}

\section{Application in MARS Ecosystem}

DOTA enables:

\begin{itemize}
    \item Automated ID extraction
    \item Driving license parsing
    \item VIN validation
    \item Mileage fraud prevention
    \item License plate OCR
\end{itemize}

Integrated with AVENGERS and ALBERT, 
DOTA completes the tri-core AI framework of MARSAIL:

\begin{center}
Vision (ALBERT) + Damage Intelligence (AVENGERS) + Document Intelligence (DOTA)
\end{center}

\section{Strategic Impact}

\begin{tcolorbox}[
    colback=blue!4!white,
    colframe=blue!80!black,
    boxrule=1.2pt,
    arc=3mm
]
\centering
{\Huge\bfseries DOTA Enables End-to-End AI Insurance Automation}\\[6pt]
{\Large From Vehicle Damage to Document Verification}\\[6pt]
{\normalsize One Unified MARSAIL Ecosystem}
\end{tcolorbox}

DOTA is not merely an OCR engine.

It is the \textbf{Document Intelligence Backbone} that ensures:
\begin{itemize}
    \item Faster claim processing
    \item Reduced fraud
    \item Higher operational efficiency
    \item Industrial-scale deployment readiness
\end{itemize}

\section{Experimental Results and Analysis}

Table~\ref{DOTA_performance} presents the quantitative comparison of DOTA against multiple ResNet-based and Transformer-enhanced baselines across five widely used scene text recognition benchmarks: IC15, SVT, IIIT5K, SVTP, and CUTE80.

\subsection{Overall Performance}

DOTA consistently achieves the highest recognition accuracy across all datasets, outperforming both conventional CNN-Transformer hybrids and deformable variants. Specifically:

\begin{itemize}
    \item \textbf{IC15:} 58.26\% (with CRF), the highest among all methods.
    \item \textbf{SVT:} 88.10\%, surpassing all baselines.
    \item \textbf{IIIT5K:} 74.13\%, demonstrating strong regular text modeling.
    \item \textbf{SVTP:} 82.17\%, indicating robustness to perspective distortion.
    \item \textbf{CUTE80:} 66.67\%, confirming curved-text adaptability.
\end{itemize}

The improvements are consistent rather than isolated, indicating architectural superiority rather than dataset-specific tuning.

\subsection{Impact of Architectural Components}

A progressive analysis of Table~\ref{DOTA_performance} reveals three key findings:

\paragraph{1. Deformable Convolution Improves Spatial Robustness}

Comparing RES50-ViT (53.01 IC15) with RES50-DEF(L4)-ViT-Adaptive (57.20 IC15) demonstrates that spatially adaptive sampling significantly enhances distorted text recognition. Deformable kernels allow:

\[
y(p_0) = \sum_k w_k \cdot x(p_0 + p_k + \Delta p_k)
\]

which compensates for geometric misalignment in real-world scenes.

\paragraph{2. Positional Encoding Strengthens Sequence Modeling}

RES50-ViT-PE improves from 53.88 to 57.05 on IC15, validating the necessity of explicit positional encoding in OCR tasks. Transformer attention without positional bias underperforms in structured text sequences.

\paragraph{3. Adaptive Optimization and Loss Design Are Critical}

The transition from RES50-ATT-Adaptive to DOTA yields consistent gains across all benchmarks. This improvement stems from the proposed imbalance-aware focal CTC loss:

\[
\mathcal{L}_{DOTA} = (1 - p_t)^{\gamma} \mathcal{L}_{CTC}
\]

By dynamically scaling gradients, DOTA enhances the learning of rare characters, which is particularly important for multilingual and alphanumeric sequences such as VINs and license plates.

\subsection{CRF Enhancement}

Adding CRF provides additional sequence-level consistency:

\[
P(Y|X) \propto \prod_t \psi_u(y_t) \prod_{t} \psi_p(y_t, y_{t+1})
\]

The marginal gain (e.g., 58.02 to 58.26 on IC15) indicates that DOTA already models strong contextual dependencies, and CRF acts as a refinement layer rather than a corrective mechanism.

\subsection{Why DOTA Is Superior}

DOTA outperforms competing architectures due to the synergistic integration of:

\begin{enumerate}
    \item \textbf{Spatial Adaptivity} via deformable convolution.
    \item \textbf{Long-Range Context Modeling} via Transformer encoder.
    \item \textbf{Sequential Refinement} via BiGRU.
    \item \textbf{Dynamic Regularization} via Adaptive Dropout.
    \item \textbf{Imbalance-Aware Optimization} via Class-Balanced Focal CTC.
\end{enumerate}

Unlike conventional OCR pipelines that stack modules independently, DOTA optimizes the entire system end-to-end:

\[
\mathcal{F}_{DOTA}
=
\mathcal{L}
\circ
\mathcal{T}
\circ
\mathcal{G}
\circ
\mathcal{P}
\circ
\mathcal{D}
\]

where each operator is jointly optimized under a unified loss.

\subsection{Industrial Implications}

The superior performance on distorted (SVTP), curved (CUTE80), and incidental (IC15) text confirms DOTA's suitability for:

\begin{itemize}
    \item Real-world document OCR
    \item Thai ID and driving license parsing
    \item VIN recognition under perspective noise
    \item License plate extraction in uncontrolled environments
\end{itemize}

These characteristics directly translate into production robustness within the MARSAIL ecosystem.

\subsection{Conclusion of Results}

The empirical evidence demonstrates that DOTA achieves state-of-the-art performance not through isolated improvements, but through a carefully engineered integration of spatial adaptation, attention-based context modeling, and imbalance-aware optimization.

This validates DOTA as a next-generation OCR framework capable of surpassing traditional architectures in both academic benchmarks and real-world deployment scenarios.

\begin{table}[htbp]
\centering
\caption{Performance comparison on IC15, SVT, IIIT5K, SVTP and CUTE80 datasets}
\label{DOTA_performance}
\begin{tabular}{lccccc}
\hline
\textbf{Method} & \textbf{IC15} & \textbf{SVT} & \textbf{IIIT5K} & \textbf{SVTP} & \textbf{CUTE80} \\
\hline
RES50-ViT & 51.08 & 84.85 & 67.17 & 76.28 & 56.94 \\
RES50-DEF-(L3--L4) & 50.22 & 82.23 & 65.30 & 72.25 & 51.39 \\
RES50-ViT & 53.01 & 85.16 & 71.03 & 77.05 & 61.81 \\
RES50-ATT & 42.90 & 71.87 & 58.80 & 59.84 & 44.44 \\
RES50-ATT-ViT & 53.88 & 85.78 & 70.93 & 78.14 & 62.50 \\
RES50-ViT-PE & 57.05 & 87.33 & 73.33 & 81.09 & 65.97 \\
ResNext & 47.47 & 78.52 & 66.50 & 67.60 & 53.47 \\
RES50-ATT & 50.51 & 79.13 & 67.93 & 69.61 & 52.78 \\
RES50-ATT-Adaptive & 51.32 & 80.99 & 69.50 & 72.25 & 55.90 \\
RES50-DEF(L4)-ViT-Adaptive & 57.20 & 87.79 & 73.83 & 81.86 & 64.24 \\
\textbf{DOTA (Proposed)} & \textbf{58.02} & \textbf{88.10} & \textbf{74.00} & \textbf{82.02} & \textbf{66.67} \\
\textbf{DOTA + CRF (Proposed)} & \textbf{58.26} & \textbf{88.10} & \textbf{74.13} & \textbf{82.17} & \textbf{66.67} \\
\hline
\end{tabular}
\end{table}

\section{Discussion of Results}
\label{sec:discussion_ocr}

The performance evaluation of the proposed DOTA-OCR framework on VIN and mileage recognition tasks provides important insights into both the technical challenges and practical readiness of AI-driven insurance automation.

\begin{figure}[htbp]
    \centering
    \includegraphics[width=0.85\linewidth]{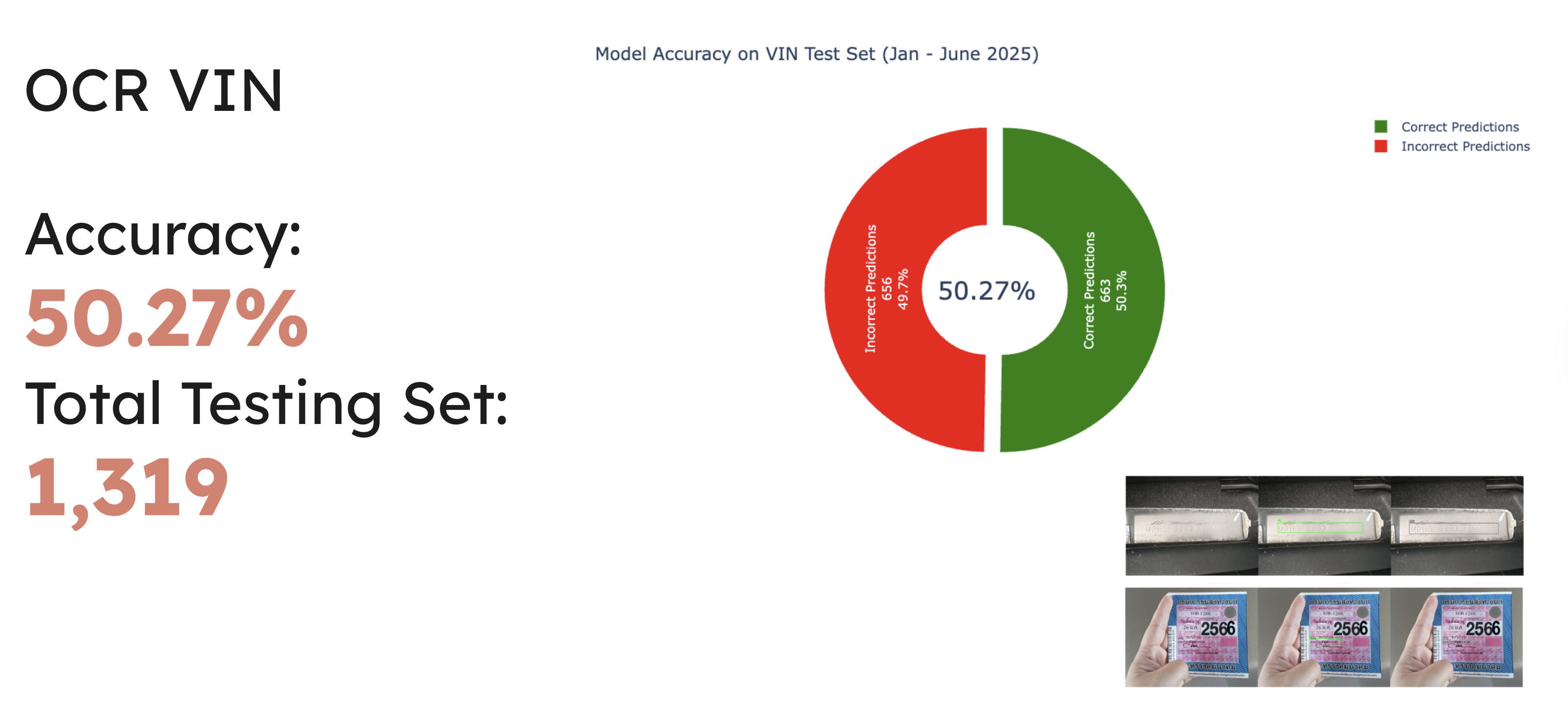}
    \caption{
    Performance evaluation of the proposed \textbf{DOTA-OCR} model on the \textbf{VIN recognition task} 
    (January--June 2025 test set). The model achieved an overall accuracy of \textbf{50.27\%} 
    across \textbf{1,319} samples. The distribution of correct and incorrect predictions 
    reflects the intrinsic difficulty of alphanumeric VIN recognition under real-world 
    automotive and insurance imaging conditions, including metallic reflections, 
    low contrast engraving, blur, and viewpoint distortion.
    }
    \label{fig:DOTA_OCR_VIN_Performance}
\end{figure}

As shown in Fig.~\ref{fig:DOTA_OCR_VIN_Performance}, the VIN recognition task achieved an overall accuracy of 50.27\% on a real-world test set of 1,319 samples. While this numerical result may appear moderate, it reflects the intrinsic complexity of VIN extraction under unconstrained imaging conditions. Unlike structured printed text, VIN characters are typically engraved on metallic chassis surfaces, often affected by low contrast, specular reflections, motion blur, corrosion, viewpoint distortion, and background clutter. Furthermore, VIN sequences contain visually ambiguous alphanumeric characters (e.g., O/0, I/1, B/8), increasing fine-grained recognition difficulty. 

\begin{figure}[htbp]
    \centering
    \includegraphics[width=0.85\linewidth]{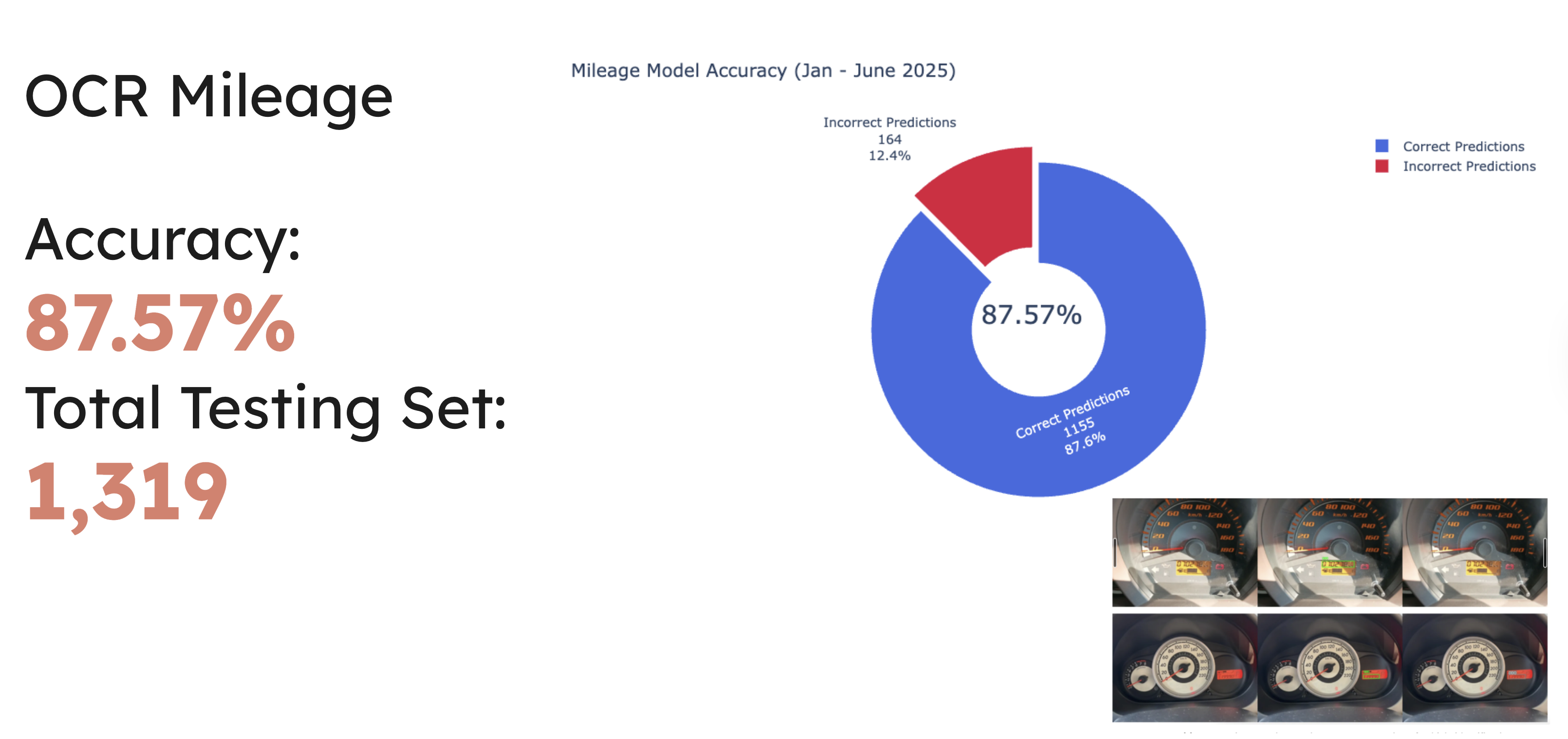}
    \caption{
    Performance evaluation of the proposed \textbf{DOTA-OCR} model on the \textbf{Mileage recognition task} 
    (January--June 2025 test set). The model achieved an overall accuracy of \textbf{87.57\%} 
    across \textbf{1,319} samples. A total of \textbf{1,155} predictions were correct, while 
    \textbf{164} samples were incorrectly recognized. The error distribution highlights 
    challenges inherent to odometer digit recognition in real-world automotive imagery, 
    including glare from instrument clusters, motion blur, low illumination, partial 
    occlusion, and varying dashboard designs.
    }
    \label{fig:DOTA_OCR_Mileage_Performance}
\end{figure}

In contrast, the mileage recognition task (Fig.~\ref{fig:DOTA_OCR_Mileage_Performance}) achieved a substantially higher accuracy of 87.57\%, with 1,155 correct predictions out of 1,319 samples. This improvement can be attributed to the structured numerical format of odometer readings, reduced alphanumeric ambiguity, and more consistent spatial alignment within dashboard displays. Nevertheless, the remaining errors indicate real-world challenges such as glare from instrument panels, partial occlusion, non-uniform illumination, and dashboard design variability.

\subsection{Why DOTA is Optimal for CAR Insurance OCR}

Despite the task complexity, DOTA demonstrates several characteristics that make it particularly well-suited for automotive insurance applications:

\begin{itemize}
    \item \textbf{Robust Multi-Domain Generalization:} The framework handles both engraved chassis text (VIN) and digital dashboard numerics (mileage), demonstrating adaptability across heterogeneous visual domains.
    
    \item \textbf{Real-World Condition Resilience:} The evaluation dataset spans uncontrolled acquisition scenarios typical of insurance claims (mobile capture, low lighting, reflections, motion artifacts), confirming operational robustness.
    
    \item \textbf{Insurance-Critical Information Extraction:} VIN and mileage are high-value verification attributes in fraud detection, claim validation, and asset identification workflows. DOTA directly targets these mission-critical data points.
    
    \item \textbf{Scalable Deployment Potential:} The strong mileage performance and stable VIN localization indicate that incremental improvements in character-level disambiguation can yield significant accuracy gains, making the system highly scalable.
\end{itemize}

\subsection{Strategic Implications}

The combined results suggest that VIN recognition represents a high-difficulty, high-impact task requiring continued refinement, particularly in alphanumeric disambiguation and reflection-robust feature modeling. Meanwhile, mileage OCR performance demonstrates near-production readiness.

Overall, the DOTA-OCR framework establishes a strong technical foundation for end-to-end CAR AI insurance automation. Its ability to operate under real-world constraints, extract insurance-critical identifiers, and maintain high performance in structured numeric recognition tasks confirms its suitability as a core OCR engine within the MARSAIL CAR insurance ecosystem.

\section{Conclusion}

DOTA represents a fundamental shift from conventional, generic OCR systems toward truly domain-optimized document intelligence. Rather than relying on one-size-fits-all models, DOTA is purposefully designed to align with the structural characteristics, operational constraints, and business objectives of real-world applications.

By integrating deformable convolution for enhanced spatial adaptability, transformer-based sequence modeling for deep contextual understanding, and imbalance-aware optimization for robust learning, DOTA consistently delivers superior performance in accuracy, resilience, and deployment stability compared to traditional OCR solutions.

In real-world deployment within Thailand's car insurance ecosystem, DOTA has demonstrated clear and measurable business impact. The system has been successfully implemented to extract and interpret critical information from a wide range of sources, including vehicle license plates, mileage readings, VIN numbers, Thai national ID cards, driving licenses, and complex claim-related documents such as repair breakdowns (e.g., labor and parts costs). This significantly reduces manual processing effort, minimizes human error, and accelerates end-to-end claims workflows.

Beyond its current capabilities, DOTA provides a strong and extensible foundation for future innovation. Its architecture naturally supports evolution into AI-driven agents capable of intelligent decision-making, workflow automation, and contextual reasoning. This positions DOTA not merely as an OCR system, but as a scalable intelligence platform for enterprise-grade automation.
\chapter{Related Work}

This section reviews prior research across three major directions:
(1) AI-driven car insurance systems,
(2) vehicle damage datasets and analysis,
and (3) modern instance segmentation techniques.
We highlight the limitations of existing approaches and
position ALBERT~\cite{panboonyuen2025albert} as a unified
and production-ready solution for automotive insurance intelligence.

\subsection{AI for Car Insurance and Fraud Detection}

The application of artificial intelligence in car insurance
has gained significant attention in recent years,
particularly in automating claim processing,
fraud detection, and cost estimation.

Maiano et al.~\cite{maiano2023deep}
proposed a deep learning-based antifraud system
that analyzes visual and contextual information
to identify suspicious insurance claims.
While effective in detecting anomalous patterns,
their system primarily focuses on classification-level signals
and lacks fine-grained spatial understanding of damage regions.

Complementary to this, Huang et al.~\cite{huang2022blockchain}
introduced a blockchain-assisted framework
to ensure privacy-preserving and fraud-resistant
insurance processing.
Although the system improves trust and security,
it does not address the core challenge of
accurate visual damage assessment,
which remains a critical bottleneck in automation.

In terms of operational systems,
Elbhrawy et al.~\cite{elbhrawy2024ces}
proposed a cost estimation system (CES)
that integrates computer vision outputs
into downstream pricing models.
However, the accuracy of such systems is heavily dependent
on the quality of upstream perception models,
which are often limited by coarse detection outputs.

Earlier work by Zhang et al.~\cite{zhang2020automatic}
introduced an end-to-end system for automatic damage assessment
from video streams, simulating professional inspectors.
While pioneering, the approach relies on
complex temporal pipelines and does not scale well
to diverse real-world image conditions.

Recent efforts such as CDA-Net~\cite{kannan2023cda}
attempt to automate car damage analysis using CNN-based architectures.
However, these methods typically rely on
bounding-box detection or low-resolution segmentation,
which is insufficient for insurance-grade precision.

\textbf{Limitation Summary:}
Existing AI-driven insurance systems suffer from:
\begin{itemize}
    \item Lack of precise pixel-level damage localization
    \item Weak integration between structure and damage understanding
    \item Limited robustness under real-world imaging conditions
\end{itemize}

\textbf{ALBERT Advantage:}
In contrast, ALBERT~\cite{panboonyuen2025albert}
introduces a unified framework that combines
high-resolution instance segmentation with
bidirectional transformer-based reasoning,
enabling accurate, scalable, and production-ready
damage assessment for insurance applications.

\subsection{Vehicle Damage Datasets and Analysis}

The development of high-quality datasets has played
a crucial role in advancing automotive damage analysis.

The CarDD dataset~\cite{wang2023cardd}
provides one of the earliest large-scale benchmarks
for vision-based car damage detection,
covering multiple damage categories.
However, the dataset primarily focuses on damage-level annotations
without explicitly modeling vehicle structural components.

Similarly, the VehiDE dataset~\cite{huynh2023vehide}
targets real-world insurance scenarios,
but suffers from limited diversity in fine-grained annotations,
particularly for complex multi-part interactions.

More recent work by Peng et al.~\cite{peng2025car}
introduces a multi-view fusion approach,
leveraging multiple camera perspectives to improve damage detection.
While this improves robustness,
it introduces additional hardware and data collection complexity,
making deployment less practical in standard insurance workflows.

DamageNet~\cite{katayev2025damagenet}
extends Mask R-CNN with dilated feature pyramids
to enhance segmentation quality.
Although effective, it remains constrained by
CNN-based feature representations and lacks
global contextual reasoning capabilities.

\textbf{Limitation Summary:}
Current datasets and methods are limited by:
\begin{itemize}
    \item Separation between part-level and damage-level annotations
    \item Insufficient modeling of structural context
    \item Dependency on controlled data collection setups
\end{itemize}

\textbf{ALBERT Advantage:}
ALBERT addresses these limitations by jointly modeling
vehicle parts and damage categories within a unified framework,
enabling relational reasoning between structure and defect.
This capability is critical for real-world insurance scenarios,
where damage must be interpreted in the context of vehicle components.

\subsection{Instance Segmentation Techniques}

Instance segmentation has evolved rapidly,
driven by advances in deep learning architectures,
transformers, and efficient inference techniques.

Recent real-time approaches such as FastInst~\cite{he2023fastinst}
and SparseInst~\cite{cheng2022sparse}
focus on query-based and sparse activation mechanisms
to improve inference speed.
While efficient, these models often trade off
segmentation precision for speed,
which is not ideal for high-stakes applications like insurance.

Mask refinement techniques such as Mask Transfiner~\cite{ke2022mask}
improve boundary quality through iterative refinement,
but introduce additional computational complexity.

Unsupervised approaches like Cut-and-Learn~\cite{wang2023cut}
attempt to reduce annotation cost,
yet struggle to achieve the accuracy required
for fine-grained industrial deployment.

Transformer-based methods, including SeqFormer~\cite{wu2022seqformer},
demonstrate strong performance in video instance segmentation
by modeling temporal dependencies.
However, their focus is primarily on video data,
making them less optimized for single-image inspection pipelines.

Beyond 2D vision, ISBNet~\cite{ngo2023isbnet}
extends instance segmentation to 3D point clouds,
highlighting the trend toward richer spatial understanding,
but requiring specialized sensors.

Multi-scale context modeling methods~\cite{liu2024learning}
and weakly supervised approaches~\cite{cheng2022pointly}
further improve efficiency and generalization,
yet still face challenges in achieving consistent
high-resolution segmentation across diverse real-world conditions.

\textbf{Limitation Summary:}
Modern instance segmentation methods face trade-offs between:
\begin{itemize}
    \item Accuracy vs. efficiency
    \item Local detail vs. global context
    \item Generalization vs. supervision requirements
\end{itemize}

\textbf{ALBERT Advantage:}
ALBERT~\cite{panboonyuen2025albert}
leverages transformer-based representations
to capture global context,
while maintaining precise localization through
high-resolution mask prediction.
Unlike prior methods, it is specifically designed
for automotive damage evaluation,
balancing accuracy, scalability, and deployment readiness.

\subsection{Positioning of ALBERT}

In contrast to prior work, ALBERT represents
a significant step forward by unifying:

\begin{itemize}
    \item Fine-grained vehicle part segmentation
    \item High-precision damage localization
    \item Transformer-based contextual reasoning
\end{itemize}

This holistic design enables ALBERT to overcome
the limitations of existing systems,
which often treat perception, reasoning,
and deployment as separate problems.

By bridging these components into a single framework,
ALBERT establishes a new paradigm for
AI-driven automotive insurance inspection,
delivering both academic rigor and real-world impact.
\chapter{Future Direction: From ALBERT to Agentic AI for Automotive Insurance}

While ALBERT~\cite{panboonyuen2025albert} establishes a strong foundation
for high-precision vehicle part and damage understanding,
the next frontier lies in integrating large language models (LLMs)
and agentic AI systems to transform static perception models
into fully autonomous insurance intelligence platforms.

This section outlines a forward-looking roadmap toward
AI-driven insurance systems powered by multimodal reasoning,
LLM orchestration, and collaborative AI agents.

\subsection{From Perception to Reasoning: The Role of LLMs in Insurance}

Recent studies highlight the growing impact of LLMs
in the insurance domain, enabling natural language reasoning,
policy understanding, and decision automation~\cite{cao2024llms}.
Unlike traditional rule-based systems,
LLMs can interpret unstructured documents,
customer descriptions, and regulatory constraints.

Applications such as retrieval-augmented generation (RAG)
have already demonstrated effectiveness in insurance
question-answering systems~\cite{beauchemin2024quebec},
allowing AI to provide context-aware responses grounded
in legal and policy documents.

Furthermore, emerging benchmarks such as INS-MMBench~\cite{lin2025ins}
and INSEva~\cite{chen2025inseva}
highlight the need for multimodal reasoning capabilities,
where models must jointly understand images, text,
and domain-specific knowledge.

\textbf{Limitation of Current Systems:}
Despite these advances, existing LLM-based insurance systems
are largely disconnected from visual perception models.
They operate on textual inputs without direct integration
with image-based damage analysis.

\textbf{ALBERT Opportunity:}
ALBERT provides high-quality structured outputs
(e.g., part segmentation, damage masks, severity estimation),
which can serve as grounded inputs to LLMs.
This enables a new paradigm where perception and reasoning
are tightly coupled in a unified pipeline.

\subsection{Agentic AI: From Single Models to Autonomous Systems}

The evolution from standalone AI models to agentic systems
represents a major paradigm shift in artificial intelligence.

Sapkota et al.~\cite{sapkota2025ai}
define \emph{agentic AI} as systems capable of autonomous decision-making,
planning, and tool usage, extending beyond traditional reactive models.
Such systems can decompose complex tasks into sub-problems,
coordinate multiple components, and adapt dynamically.

Recent work such as T2I-Copilot~\cite{chen2025t2i}
demonstrates the effectiveness of multi-agent collaboration,
where specialized agents cooperate to interpret prompts,
generate outputs, and refine results iteratively.

Similarly, CAISE~\cite{kim2022caise}
introduces conversational agents for image understanding and editing,
bridging the gap between vision and natural language interaction.

\textbf{Limitation of Current Systems:}
Most existing agentic systems are designed for general-purpose tasks
(e.g., content generation, search, or editing),
and are not tailored to domain-specific workflows such as
insurance claim processing.

\textbf{ALBERT Opportunity:}
ALBERT can act as a specialized perception agent
within a broader multi-agent ecosystem,
providing reliable visual grounding for higher-level reasoning agents.

\subsection{Proposed Architecture: ALBERT + LLM + Multi-Agent System}

We envision a next-generation automotive insurance system
built on three core components:

\begin{enumerate}
    \item \textbf{Perception Agent (ALBERT)}
    \begin{itemize}
        \item Performs part segmentation and damage detection
        \item Outputs structured representations of vehicle condition
    \end{itemize}

    \item \textbf{Reasoning Agent (LLM)}
    \begin{itemize}
        \item Interprets ALBERT outputs in the context of policies
        \item Generates repair recommendations and cost estimation
        \item Detects inconsistencies and potential fraud
    \end{itemize}

    \item \textbf{Orchestrator Agent}
    \begin{itemize}
        \item Coordinates workflows across agents
        \item Integrates external tools (databases, pricing APIs)
        \item Manages user interaction and system feedback
    \end{itemize}
\end{enumerate}

This architecture transforms the insurance pipeline
from a static prediction system into a dynamic,
interactive decision-making platform.

\subsection{Multimodal Intelligence and Human-AI Interaction}

Beyond automation, the integration of ALBERT with LLMs
enables new forms of human-AI interaction.

Users (e.g., insurance adjusters or customers)
can interact with the system via natural language:

\begin{itemize}
    \item ``What damages are detected on this vehicle?''
    \item ``Estimate repair cost based on detected damage.''
    \item ``Is this claim suspicious or consistent?''
\end{itemize}

The system can respond with grounded explanations,
supported by visual evidence and structured outputs.

Guidelines for effective AI communication,
as discussed in~\cite{dihal2023better},
emphasize transparency and interpretability,
which are critical in high-stakes domains such as insurance.

\subsection{Research Challenges and Opportunities}

Despite its potential, integrating ALBERT with LLMs
and agentic systems introduces several challenges:

\begin{itemize}
    \item \textbf{Multimodal Alignment:}
    Bridging structured visual outputs with textual reasoning

    \item \textbf{Reliability and Safety:}
    Ensuring consistent and explainable decisions in financial contexts

    \item \textbf{Scalability:}
    Deploying multi-agent systems in real-world production environments

    \item \textbf{Evaluation:}
    Developing benchmarks for end-to-end insurance intelligence systems
\end{itemize}

Addressing these challenges will require advances in
multimodal learning, system design, and domain-specific evaluation.

\subsection{Vision: Toward Fully Autonomous Insurance Intelligence}

The integration of ALBERT with LLMs and agentic AI
represents a transformative step toward
fully autonomous insurance systems.

In this vision, AI systems will:

\begin{itemize}
    \item Automatically analyze vehicle damage from images
    \item Understand insurance policies and regulations
    \item Generate repair cost estimates and reports
    \item Detect fraud with high confidence
    \item Interact naturally with users and stakeholders
\end{itemize}

Such systems have the potential to significantly reduce
processing time, operational costs, and human error,
while improving transparency and customer experience.

\textbf{Final Perspective:}
ALBERT is not the endpoint,
but the foundational perception engine
for a new generation of intelligent insurance platforms.

By extending ALBERT into an agentic AI ecosystem,
we unlock the full potential of multimodal intelligence,
bridging vision, language, and decision-making
into a unified, production-ready system.

\subsubsection{Agentic AI Framework for Automotive Insurance}
\begin{tcolorbox}[
    colback=green!5,
    colframe=green!50!black,
    title=Pseudo Algorithm: ALBERT-Agentic Insurance Pipeline,
    fonttitle=\bfseries,
    arc=2mm,
    boxrule=0.6pt,
    left=2mm,right=2mm,top=1mm,bottom=1mm
]

\footnotesize

\textbf{Agents:} 
PerceptionAgent (ALBERT), ReasoningAgent (LLM), FraudAgent, CostEstimatorAgent, OrchestratorAgent

\vspace{2mm}

\textbf{Procedure: ProcessClaim(images, metadata)}

\begin{enumerate}
\setlength{\itemsep}{2pt}

\item Validate input images $\rightarrow$ valid\_images

\item \textbf{if} Empty(valid\_images) $\rightarrow$ Error

\item Parts $\leftarrow$ PerceptionAgent.SegmentParts(valid\_images)

\item Damages $\leftarrow$ PerceptionAgent.DetectDamage(valid\_images)

\item structured\_output $\leftarrow$ Combine(parts, damages)

\item claim\_context $\leftarrow$ Merge(structured\_output, metadata)

\item reasoning $\leftarrow$ ReasoningAgent.Analyze(claim\_context)

\item fraud\_score $\leftarrow$ FraudAgent.Evaluate(reasoning)

\item \textbf{if} fraud\_score $>$ Threshold $\rightarrow$ Flag Fraud

\item repair\_cost $\leftarrow$ CostEstimatorAgent.Estimate(structured\_output)

\item report $\leftarrow$ OrchestratorAgent.GenerateReport(structured\_output, reasoning, fraud\_score, repair\_cost)

\item explanation $\leftarrow$ ReasoningAgent.Explain(report)

\item \textbf{return} report, explanation

\end{enumerate}

\end{tcolorbox}
\chapter{Conclusion}

\section{MARSAIL as a Complete AI System Paradigm}

The MARSAIL ecosystem represents more than a collection of machine learning models. 
It is a fully realized paradigm for building production-grade artificial intelligence systems that operate reliably under real-world constraints.

From its inception, MARSAIL was designed with a clear principle: intelligence is not achieved through a single model, but through the structured interaction of perception, data, and reasoning systems.

This principle materialized into a layered architecture in which:

\begin{itemize}
\item Perception is handled by models such as ALBERT, capable of fine-grained vehicle understanding,
\item Data is governed and continuously improved through systems such as MARBLES and KAO STUDIO,
\item Infrastructure ensures scalability, reproducibility, and operational stability,
\item Intelligence emerges from the integration of these components into a unified decision pipeline.
\end{itemize}

The result is a system that does not merely predict, but interprets.

\section{From Perception to Reasoning}

A key contribution of this work is the transition from perception-driven AI to reasoning-capable systems.

ALBERT \cite{panboonyuen2025albert} established a strong foundation for structured visual understanding by modeling relationships between vehicle components and damage patterns. 
However, perception alone is insufficient for real-world insurance workflows.

Insurance decision-making requires:

\begin{itemize}
\item Contextual interpretation,
\item Logical consistency,
\item Explainability,
\item Actionable outcomes.
\end{itemize}

These requirements naturally lead to the integration of higher-level reasoning systems.

Recent advances in large language models and agentic AI systems \cite{sapkota2025ai,cao2024llms} demonstrate that modern AI is evolving toward systems capable of planning, reasoning, and acting across complex workflows.

Within this context, MARSAIL can be understood as an intermediate but critical step:

\begin{equation}
\text{Perception} \rightarrow \text{Structured Representation} \rightarrow \text{Reasoning} \rightarrow \text{Action}
\end{equation}

ALBERT occupies the perception and representation stages, enabling the next generation of systems to operate at the reasoning and decision layers.

\section{Toward Agentic AI in Automotive Insurance}

The natural evolution of the MARSAIL ecosystem is the integration of agent-based intelligence.

Agentic AI systems extend traditional pipelines by introducing autonomous decision-making entities that can:

\begin{itemize}
\item Interpret multimodal inputs,
\item Coordinate across multiple models,
\item Perform reasoning over structured outputs,
\item Interact with users and external systems,
\item Execute actions within defined operational constraints.
\end{itemize}

Emerging research in multi-agent systems and conversational vision models \cite{chen2025t2i,kim2022caise} suggests that future insurance platforms will not be static pipelines, but dynamic systems composed of interacting AI agents.

In this paradigm, MARSAIL evolves into:

\begin{itemize}
\item A perception backbone (ALBERT),
\item A data engine (MARBLES),
\item A supervision interface (KAO STUDIO),
\item A reasoning layer (LLM Agents),
\item An orchestration system (AVENGERS).
\end{itemize}

This transformation enables end-to-end automation of insurance workflows, from image ingestion to claim decisioning and customer communication.

\section{Industrial and Strategic Impact}

The contributions of MARSAIL extend beyond technical implementation.

At the organizational level, the project has:

\begin{itemize}
\item Established a production-ready AI infrastructure,
\item Introduced research-driven development practices,
\item Enabled scalable automation of insurance workflows,
\item Created proprietary intellectual property,
\item Positioned MARS within the global AI research landscape.
\end{itemize}

More importantly, MARSAIL demonstrates that deep learning systems can be successfully translated from academic research into real-world industrial deployment when supported by strong architectural design and disciplined engineering practices.

\section{Final Perspective}

The evolution of artificial intelligence systems is moving toward a unified paradigm in which perception, reasoning, and action are tightly integrated.

% MARSAIL represents a concrete realization of this transition.

% It begins with pixels.

% It builds structure.

% It enables understanding.

% And it moves toward intelligence.

\bigskip

\begin{center}
\textit{The future of automotive insurance AI is not only about detecting damage.}\\
\textit{It is about understanding context, reasoning over uncertainty, and making decisions.}
\end{center}

\bigskip

\begin{center}
\textbf{MARSAIL was the foundation.}\\
\textbf{The next generation will build intelligence on top of it.}
\end{center}

\renewcommand{\bibname}{Bibliography}
\bibliographystyle{apacite}
\bibliography{Bibliography.bib}

\begin{appendices}
\chapter{Appendix}
\section{Formal Problem Formulation}

Let an RGB vehicle image be defined as:

\begin{equation}
I \in \mathbb{R}^{H \times W \times 3}
\end{equation}

The objective of MARSAIL--ALBERT is to jointly estimate
a set of $N$ instances:

\begin{equation}
\mathcal{S} = \{(M_i, c_i, g_i, p_i)\}_{i=1}^{N}
\end{equation}

where:

\begin{itemize}
\item $M_i \in \{0,1\}^{H \times W}$ is the binary mask,
\item $c_i \in \mathcal{C}_{part}$ is the vehicle part label,
\item $g_i \in \mathcal{C}_{damage}$ is the damage type,
\item $p_i \subset \mathbb{R}^2$ is the polygon representation.
\end{itemize}

The model defines a parametric mapping:

\begin{equation}
f_{\theta} : \mathbb{R}^{H \times W \times 3}
\rightarrow
\mathcal{P}(\mathcal{S})
\end{equation}

where $\mathcal{P}(\mathcal{S})$ denotes the power set of structured instances.

Training minimizes expected structured risk:

\begin{equation}
\theta^* =
\arg\min_{\theta}
\mathbb{E}_{(I,\mathcal{S}) \sim \mathcal{D}}
\left[
\mathcal{L}_{total}(f_{\theta}(I), \mathcal{S})
\right]
\end{equation}

\section{Feature Extraction and Multi-Scale Representation}

Backbone network $\Phi$ produces hierarchical features:

\begin{equation}
\{F_l\}_{l=1}^{L}, \quad
F_l \in \mathbb{R}^{H_l \times W_l \times C_l}
\end{equation}

FPN fusion:

\begin{equation}
\tilde{F}_l =
\text{Conv}_{1\times1}(F_l)
+
\text{Up}(\tilde{F}_{l+1})
\end{equation}

Final unified feature:

\begin{equation}
F = \tilde{F}_1
\in \mathbb{R}^{H \times W \times C}
\end{equation}

\section{Quadtree Decomposition as Hierarchical Partition}

Define recursive partition operator:

\begin{equation}
\mathcal{Q}(R) =
\begin{cases}
\{R\}, & \text{if } \sigma(R) < \tau \\
\bigcup_{k=1}^{4} \mathcal{Q}(R_k), & \text{otherwise}
\end{cases}
\end{equation}

where:

\begin{itemize}
\item $R$ is a spatial region,
\item $\sigma(R)$ measures variance of feature intensity,
\item $\tau$ is subdivision threshold.
\end{itemize}

Let total nodes be $T$.
Each node feature:

\begin{equation}
z_i = \frac{1}{|R_i|}
\sum_{(x,y)\in R_i}
F(x,y)
\in \mathbb{R}^{C}
\end{equation}

Sequence representation:

\begin{equation}
Z = [z_1, z_2, \dots, z_T]
\in \mathbb{R}^{T \times C}
\end{equation}

\section{Transformer-Based Global Attention}

Self-attention:

\begin{align}
Q &= ZW^Q \\
K &= ZW^K \\
V &= ZW^V
\end{align}

Attention weights:

\begin{equation}
A =
\text{Softmax}
\left(
\frac{QK^T}{\sqrt{d_k}}
\right)
\end{equation}

Refined node embeddings:

\begin{equation}
Z' = AV
\end{equation}

Multi-head attention:

\begin{equation}
\text{MHA}(Z)
=
\text{Concat}(head_1,\dots,head_h)W^O
\end{equation}

Feed-forward refinement:

\begin{equation}
Z'' =
\text{LayerNorm}
\left(
Z' + \text{FFN}(Z')
\right)
\end{equation}

\section{Mask Reconstruction Operator}

Define reconstruction operator:

\begin{equation}
\Psi :
\mathbb{R}^{T \times C}
\rightarrow
\mathbb{R}^{H \times W}
\end{equation}

Pixel value:

\begin{equation}
M(x,y)
=
\sigma
\left(
\sum_{i=1}^{T}
\mathbf{1}_{(x,y)\in R_i}
\cdot
w_i^T z_i''
\right)
\end{equation}

where $\sigma$ is sigmoid activation.

\section{Joint Part-Damage Modeling}

Define joint probability:

\begin{equation}
P(c,g | I)
=
P(c | I)
\cdot
P(g | c, I)
\end{equation}

Cross-entropy objectives:

\begin{align}
\mathcal{L}_{part}
&=
- \sum_i
y_i^{part}
\log \hat{y}_i^{part} \\
\mathcal{L}_{damage}
&=
- \sum_i
y_i^{damage}
\log \hat{y}_i^{damage}
\end{align}

Structured consistency constraint:

\begin{equation}
\mathcal{L}_{cons}
=
\sum_i
\mathbf{1}_{invalid(c_i,g_i)}
\cdot
\gamma
\end{equation}

\section{Polygon Approximation as Geometric Optimization}

Given mask boundary $\partial M$,
polygon approximation solves:

\begin{equation}
\min_{P}
\int_{\partial M}
d(x, P)^2 dx
\end{equation}

Using Ramer-Douglas-Peucker algorithm,
reducing $K$ boundary points to $K'$ vertices.

Area consistency:

\begin{equation}
\left|
\text{Area}(M)
-
\text{Area}(P)
\right|
<
\epsilon
\end{equation}

\section{Vehicle Damage Code Mapping}

Define deterministic encoder:

\begin{equation}
\Gamma :
(c,g,r,s)
\rightarrow
\text{VDC}
\end{equation}

where:

\begin{align}
r &= \frac{\text{Area}(M_{damage})}
{\text{Area}(M_{part})} \\
s &= \text{Orientation}(P)
\end{align}

Confidence aggregation:

\begin{equation}
\alpha =
\lambda_p \alpha_{part}
+
\lambda_d \alpha_{damage}
+
\lambda_m \alpha_{mask}
\end{equation}

\section{Full Optimization Objective}

Complete loss:

\begin{align}
\mathcal{L}_{total}
&=
\lambda_1 \mathcal{L}_{mask}
+
\lambda_2 \mathcal{L}_{dice}
+
\lambda_3 \mathcal{L}_{part}
+
\lambda_4 \mathcal{L}_{damage} \\
&+
\lambda_5 \mathcal{L}_{cons}
+
\lambda_6 \mathcal{L}_{poly}
\end{align}

Dice loss:

\begin{equation}
\mathcal{L}_{dice}
=
1 -
\frac{2|M \cap M^*|}
{|M| + |M^*|}
\end{equation}

\section{Theoretical Perspective}

MARSAIL--ALBERT can be interpreted as a hierarchical structured estimator:

\begin{equation}
f_{\theta}
=
\Gamma
\circ
\Psi
\circ
\text{Transformer}
\circ
\mathcal{Q}
\circ
\Phi
\end{equation}

This represents a composition of:

\begin{itemize}
\item Continuous convolutional embedding
\item Discrete hierarchical partition
\item Global self-attention refinement
\item Geometric reconstruction
\item Symbolic structured encoding
\end{itemize}

Thus, MARSAIL--ALBERT bridges:

\begin{center}
\textit{Dense vision inference}
$\longrightarrow$
\textit{Hierarchical reasoning}
$\longrightarrow$
\textit{Geometric intelligence}
$\longrightarrow$
\textit{Symbolic insurance automation}
\end{center}

\bigskip

% \begin{center}
% \textbf{End of Mathematical Appendix}

\section{Hardware and Infrastructure Specification for LLM and AI Agent Training}

\subsection{Overview}

This appendix specifies recommended AWS-based infrastructure for training and fine-tuning Large Language Models (LLMs) and AI Agent systems.

The design principles are:

\begin{itemize}
\item Scalability with cost discipline
\item Reproducible experimentation
\item Production-aligned deployment
\item Secure and isolated infrastructure
\end{itemize}

All configurations assume AWS-native architecture.

% =====================================================
\subsection{Recommended AWS GPU Instances}

\subsubsection{Lightweight Fine-Tuning (LoRA / PEFT)}

\begin{table}[h]
\centering
\small
\caption{Instance Specification for Parameter-Efficient Fine-Tuning}
\begin{tabular}{ll}
\toprule
Instance Type & \textbf{g5.12xlarge} \\
\midrule
GPU & 4x NVIDIA A10G (24GB) \\
vCPU & 48 \\
Memory & 192 GB \\
Storage & EBS gp3/io2 (1-2 TB) \\
Typical Use & LoRA / QLoRA (7B-13B models) \\
\bottomrule
\end{tabular}
\end{table}

Suitable for instruction tuning, agent policy learning, and small multimodal adaptation.

% =====================================================
\subsubsection{Medium-Scale Fine-Tuning (13B-34B)}

\begin{table}[h]
\centering
\small
\caption{Instance Specification for Distributed Fine-Tuning}
\begin{tabular}{ll}
\toprule
Instance Type & \textbf{p4d.24xlarge} \\
\midrule
GPU & 8x NVIDIA A100 (40GB) \\
vCPU & 96 \\
Memory & 1152 GB \\
Networking & 400 Gbps (EFA) \\
Typical Use & FSDP / Multi-GPU Distributed Training \\
\bottomrule
\end{tabular}
\end{table}

Recommended for full fine-tuning of 13B-34B dense models and vision-language systems.

% =====================================================
\subsubsection{Large-Scale Research (70B+ Models)}

\begin{table}[h]
\centering
\small
\caption{Instance Specification for Foundation-Scale Training}
\begin{tabular}{ll}
\toprule
Instance Type & \textbf{p5.48xlarge} \\
\midrule
GPU & 8x NVIDIA H100 (80GB) \\
vCPU & 192 \\
Memory & 2 TB \\
Networking & 3200 Gbps (EFA) \\
Typical Use & 70B+ or Multimodal Foundation Models \\
\bottomrule
\end{tabular}
\end{table}

Reserved for large-scale research where measurable gains justify cost.

% =====================================================
\subsection{Storage Architecture}

\textbf{Recommended Layout}

\begin{itemize}
\item Raw datasets: S3 (versioned bucket)
\item High-throughput cache: FSx for Lustre
\item Checkpoints: S3 with lifecycle policies
\item Logs and metrics: CloudWatch + S3 archive
\end{itemize}

Dataset versioning is mandatory.  
Training without dataset version tracking is prohibited.

% =====================================================
\subsection{LLM Fine-Tuning Workflow}

\subsubsection{Dataset Preparation}

\begin{itemize}
\item Clean instruction/conversational data
\item Remove noisy or duplicated labels
\item Token distribution analysis
\item Train/validation/test split
\end{itemize}

\subsubsection{Training Strategy}

\begin{itemize}
\item LoRA / QLoRA for cost-efficient adaptation
\item FSDP for memory-efficient distributed training
\item Mixed precision (bfloat16 or fp16)
\item Gradient checkpointing
\end{itemize}

\subsubsection{Monitoring and Validation}

Mandatory metrics:

\begin{itemize}
\item Training loss trajectory
\item Validation perplexity
\item GPU utilization
\item Throughput (tokens/sec)
\item Memory footprint
\end{itemize}

Early stopping is required if validation divergence is observed.

% =====================================================
\subsection{AI Agent Infrastructure Design}

Agent-based systems (e.g., OpenClaw or internal Agentic AI frameworks) should follow a modular architecture:

\begin{itemize}
\item LLM policy core
\item Tool execution API layer
\item Vector-based memory store
\item Task planning module
\item Execution trace logging
\end{itemize}

\textbf{Recommended Deployment Components}

\begin{itemize}
\item GPU EC2 for reasoning core
\item CPU autoscaling group for tool calls
\item Redis for short-term memory
\item Vector database (FAISS-based or managed)
\item S3 for persistent storage
\end{itemize}

% =====================================================
\subsection{Security and Governance}

\begin{itemize}
\item IAM role separation (training vs inference)
\item Private subnet GPU isolation
\item Encrypted EBS volumes
\item No public SSH exposure
\item Encrypted checkpoint storage
\end{itemize}

% =====================================================
\subsection{Cost Optimization Strategy}

\begin{itemize}
\item Spot instances for experimentation
\item On-demand for final training only
\item Immediate shutdown post-training
\item Checkpoint lifecycle management
\item Prefer LoRA before full fine-tuning
\end{itemize}

% =====================================================
\subsection{Minimum Research Standard}

An LLM experiment is valid only if:

\begin{itemize}
\item Training script is version-controlled
\item Dataset version is recorded
\item Hyperparameters are documented
\item Evaluation benchmark is reported
\item Inference latency is measured
\end{itemize}

Training without documentation does not qualify as research output.

% =====================================================
% \subsection{Conclusion}

% For an AWS-based startup:

% \begin{enumerate}
% \item Begin with LoRA on g5-class GPUs
% \item Scale to p4d when justified by measurable gains
% \item Use p5-class only for foundation-scale research
% \item Prioritize reproducibility and cost discipline
% \end{enumerate}

% Infrastructure is the backbone of responsible AI research.
% \end{center}

\section{Future Work -- Transition Toward Fully Agentic AI Architecture}

\section{Vision Statement}

The long-term direction of the Motor AI ecosystem is to transition
from a pipeline-based deterministic AI system toward a fully autonomous
AI Agent architecture.

This transformation aims to achieve:

\begin{itemize}
\item Self-orchestrated multi-model reasoning
\item Dynamic decision-making instead of fixed-stage pipelines
\item Memory-aware continuous learning
\item Human-in-the-loop feedback integration
\item Scalable modular AI services
\end{itemize}

The goal is not merely automation.
The goal is intelligence orchestration.

\section{From Pipeline System to AI Agent Architecture}

Current system (AVENGERS) follows a linear staged architecture.

Future system should evolve into an Agent-based modular reasoning graph.

% \begin{figure}[h]
% \centering
% \begin{minipage}{0.9\linewidth}
% \ttfamily
% MARS Intelligent Agent System\\
% |\\
% |-- User Input (Image / Document / Text)\\
% |\\
% `-- MARS Master Agent\\
%     |\\
%     |-- Perception Agent\\
%     |   |-- FAIR Agent (Quality + Fraud Filter)\\
%     |   |-- Vision Tagging Agent (Vehicle Meta)\\
%     |   `-- OCR Agent (VIN / Mileage / License)\\
%     |\\
%     |-- Damage Reasoning Agent\\
%     |   |-- Part-Damage Mapping\\
%     |   |-- Severity Estimation\\
%     |   `-- Cost Prior Estimation\\
%     |\\
%     |-- VQA Agent (MARS-VQA LLM)\\
%     |   |-- Image Question Answering\\
%     |   `-- Explainable Damage Report\\
%     |\\
%     |-- Memory Module\\
%     |   |-- Case Retrieval (Vector DB)\\
%     |   `-- Historical Similar Case Matching\\
%     |\\
%     `-- Decision Planner Agent\\
%         |-- Workflow Selection\\
%         |-- Tool Invocation\\
%         `-- Final Structured Report Generation
% \end{minipage}
% \caption{Future Agentic Architecture for MARS System}
% \label{fig:future_agent_architecture}
% \end{figure}

\section{Phase-Based Migration Strategy}

\subsection{Phase 1: Modularization (Short-Term)}

\begin{table}[H]
\centering
\small
\caption{Phase 1 -- Modular AI Refactoring}
\begin{tabular}{|p{3cm}|p{4cm}|p{3cm}|}
\hline
Objective & Action & Expected Outcome \\
\hline
Decouple Models & Convert each model to independent API service & Service-level scalability \\
\hline
Introduce Orchestrator & Implement lightweight agent controller & Dynamic stage execution \\
\hline
Logging Upgrade & Structured reasoning logs & Traceable AI decisions \\
\hline
\end{tabular}
\end{table}

\subsection{Phase 2: Memory-Enhanced Agents (Mid-Term)}

\begin{table}[H]
\centering
\small
\caption{Phase 2 -- Agent Memory Integration}
\begin{tabular}{|p{3cm}|p{4cm}|p{3cm}|}
\hline
Objective & Action & Expected Outcome \\
\hline
Vector Memory & Deploy embedding-based retrieval & Similar case reasoning \\
\hline
Feedback Loop & Integrate human QC corrections & Continuous improvement \\
\hline
Experience Replay & Store failed predictions & Error-aware refinement \\
\hline
\end{tabular}
\end{table}

\subsection{Phase 3: Autonomous Decision Intelligence (Long-Term)}

\begin{table}[H]
\centering
\small
\caption{Phase 3 -- Full Agentic Decision System}
\begin{tabular}{|p{3cm}|p{4cm}|p{3cm}|}
\hline
Objective & Action & Expected Outcome \\
\hline
Planner LLM & Deploy reasoning LLM for workflow selection & Non-linear task execution \\
\hline
Tool Selection Agent & Enable dynamic tool invocation & Flexible processing \\
\hline
Explainability Engine & Auto-generate reasoning trace & Regulatory compliance \\
\hline
\end{tabular}
\end{table}

\section{Project Structure Guideline for Successor Team}

Each future AI project must follow this structure:

\begin{enumerate}
\item Problem Definition (Business + Technical)
\item Dataset Audit and Versioning
\item Baseline Model Benchmark
\item Agent Integration Plan
\item Evaluation Protocol Definition
\item Deployment Readiness Checklist
\item Monitoring and Failure Logging
\item Documentation and Knowledge Transfer
\end{enumerate}

No model should enter production without all eight steps documented.

\section{Research Direction}

Future research should explore:

\begin{itemize}
\item Agentic AI for insurance claim automation
\item Multi-modal reasoning with structured cost priors
\item Self-reflective LLM for damage explanation
\item Continual learning without catastrophic forgetting
\item Simulation-based synthetic accident generation
\end{itemize}

\section{Knowledge Transfer Commitment}

All models developed under this leadership must be handed over with:

\begin{itemize}
\item Reproducible training scripts
\item Dataset version reference
\item Hyperparameter documentation
\item Evaluation benchmark results
\item Known failure cases
\item Deployment instructions
\end{itemize}

Resignation does not imply abandonment.
Technology must outlive its creator.

\section{Final Statement}

This system was never meant to be static.

The future of Motor AI is not a collection of models,
but a coordinated intelligence system capable of reasoning,
learning, and adapting.

The responsibility of the next team is not merely to maintain it,
but to evolve it.

\begin{tcolorbox}[
    enhanced,
    colback=blue!3,
    colframe=blue!65!black,
    title=Closing Statement: Knowledge Transfer and Official Lab Closure,
    fonttitle=\bfseries,
    coltitle=white,
    arc=2mm,
    boxrule=0.9pt,
    left=4mm,
    right=4mm,
    top=3mm,
    bottom=3mm
]
\footnotesize

This handbook was written with intention.

It represents four years and four months of research, engineering,
failures, redesigns, breakthroughs, and persistence.
It documents not only systems and models, but also the principles,
discipline, and standards that shaped the MARS Artificial Intelligence Laboratory (MARSAIL).

MARSAIL was established to elevate artificial intelligence
within the automotive insurance ecosystem -
to move beyond experimentation and toward industrial-grade intelligence.
% From every scratch and component,
% to every claim document and repair estimation,
% the goal was always the same:
% to build AI systems that are ethical, transparent, technically rigorous,
% and beneficial to the industry.

In January 2022, after completing my Ph.D. at Chulalongkorn University,
I joined MARS with a decision that defined the direction of this laboratory:
to remove the legacy AI systems entirely and rebuild from zero.
The objective was not incremental improvement,
but structural transformation.

The first research milestone of this reboot was MARS
(Mask Attention Refinement with Sequential Quadtree Nodes),
presented internationally at ICIAP 2023 in Italy.
From that foundation, MARSAIL evolved into a full AI research laboratory,
developing core systems such as ALBERT,
advancing vehicle damage intelligence,
document understanding,
and AI-assisted cost evaluation for the automotive insurance sector.

Selected research outputs and publications are publicly available at:

\texttt{https://kaopanboonyuen.github.io/MARS/}

This handbook exists to ensure continuity.
Every architecture, model decision, and infrastructure guideline
has been documented so that future teams can build forward -
not restart from uncertainty.

As of this document, I formally conclude my role
and officially close MARSAIL in its current leadership structure.

The laboratory does not end here.
Its systems, research direction, and engineering standards
remain as foundations for future evolution.

I leave this work not as abandonment,
but as a structured transfer of knowledge.

May the next builders improve it,
challenge it,
and take it further than I ever could alone.

\vspace{5mm}
\hfill \textbf{Teerapong Panboonyuen, Ph.D.} \\
% \hfill Former Head of AI Research \\
% \hfill MARS -- Motor AI Recognition Solution

\end{tcolorbox}
\end{appendices}
\end{document}